\documentclass[journal]{IEEEtran}
\usepackage{amsmath}
\usepackage{array}
\usepackage{graphicx}
\usepackage{amssymb}
\usepackage{times}
\usepackage{float}
\usepackage{subfigure}
\usepackage{url}
\usepackage{color}
\usepackage{enumerate}
\usepackage{paralist}

\newcommand{\figdir}{figures}

\def\swthree{0.3\linewidth}
\def\swfour{0.22\linewidth}

\def\swsix{0.145\linewidth}

\usepackage{booktabs}
\usepackage{algorithm}
\usepackage{algorithmic}
\graphicspath{{figures/}}
\ifCLASSINFOpdf
\else
\fi

\begin{document}
%
\title{Deep Dynamic Scene Deblurring from Optical Flow}

\author{Jiawei~Zhang,
		~Jinshan~Pan,
		~Daoye~Wang,
		~Shangchen~Zhou,\\
        ~Xing~Wei,
        ~Furong~Zhao,
        ~Jianbo~Liu,
        ~and Jimmy~Ren
\thanks{J. Zhang, D. Wang, S. Zhou, F. Zhao, J, Liu and J. Ren are with SenseTime Research, Shenzhen, P.R. China.}%
\thanks{J. Ren is also with Qing Yuan Research Institute, Shanghai Jiao Tong University, Shanghai, P.R. China.}%
\thanks{J. Pan is with Nanjing University of Science and Technology, Nanjing, P.R. China. E-mail: \{sdluran@gmail.com\}}%
\thanks{X. Wei is with School of Software Engineering, Xi'an Jiaotong University, Xi'an, P.R. China.}%
\thanks{This work has been supported in part by the National Natural Science Foundation of
China (Nos. 61922043, 61872421), the Natural Science Foundation of Jiangsu Province (No.
BK20180471), the Fundamental Research Funds for the Central Universities  (No. 30920041109), and National Key R\&D Program of China (No. 2018AAA0102001).}
\thanks{Copyright\copyright20xx IEEE. Personal use of this material is permitted. However, permission to use this material for any other purposes must be obtained from the IEEE by sending an email to pubs-permissions@ieee.org.}
}

\maketitle
\title{Deep Dynamic Scene Deblurring from Optical Flow}

\begin{abstract}
Deblurring can not only provide visually more pleasant pictures and make photography more convenient, but also can improve the performance of objection detection as well as tracking.
However, removing dynamic scene blur from images is a non-trivial task as it is difficult to model the non-uniform blur mathematically.
Several methods first use single or multiple images to estimate optical flow (which is treated as an approximation of blur kernels)
and then adopt non-blind deblurring algorithms to reconstruct the sharp images.
However, these methods cannot be trained in an end-to-end manner and are usually computationally expensive.
In this paper, we explore optical flow to remove dynamic scene blur by using the multi-scale spatially variant recurrent neural network (RNN).
We utilize FlowNets to estimate optical flow from two consecutive images in different scales.
The estimated optical flow provides the RNN weights in different scales so that the weights can better help RNNs to remove blur in the feature spaces.
Finally, we develop a convolutional neural network (CNN) to restore the sharp images from the deblurred features.
Both quantitatively and qualitatively evaluations on the benchmark datasets demonstrate that the proposed method performs favorably against state-of-the-art algorithms
in terms of accuracy, speed, and model size.
\end{abstract}


%
\IEEEpeerreviewmaketitle

\vspace{-2mm}
\section{Introduction}
\vspace{-1mm}
Restoring a sharp image from a blurry one is a highly ill-posed problem which attracts much attention from vision and image processing communities in the last decade.
This problem becomes increasingly important as more and more photos are taken, and some of them suffer from significant blur due to the camera shake or object motions.
Since most of the captured scenes (e.g., moving cars, pedestrians) are ephemeral and difficult to reproduce, it is of great interest to develop an effective algorithm to recover the clear images.
Most existing algorithms usually assume that a blurred image is modeled as a latent image convolved with a blur kernel.
However, this assumption does not always hold for dynamic scenes as the blur in the dynamic scenes is spatially variant, which is mainly caused by abrupt depth variations and object motions.

To solve the dynamic scene deblurring problem, conventional algorithms~\cite{kim2013dynamic, kim2014segmentation, pan2016soft} usually use image segmentation methods to help estimate blur kernels and sharp latent images. However, these algorithms highly depend on whether the image segmentation is accurate or not.

Motivated by the success of the deep learning in high-level vision tasks, deep convolutional neural networks (CNNs) have also been developed in dynamic scene deblurring \cite{sun2015learning, gong2017motion,nah2017deep, tao2018scale, zhou2019spatio,CascadedVD/cvpr20}.
One of the representative algorithms is to use fully end-to-end trainable neural networks to directly estimate sharp images.
However, removing blur should require networks with large receptive fields~\cite{xu2014inverse, xu2014deep}.
Thus, deblurring networks~\cite{nah2017deep, tao2018scale} usually use a multi-scale strategy to enlarge the receptive field.
However, this will lead to relatively large models which accordingly increases the computational cost.

%
Zhang~et al.~\cite{zhang2018dynamic} develop a spatially variant recurrent neural network (RNN)~\cite{liu2016learning}, whose pixel-wise weights can be learned from a CNN, to remove dynamic scene blur. They demonstrate that using the spatially variant RNN is able to increase the receptive field.
In addition, as one of the key component for blur removal, the spatially variant RNN can be treated as a deblurring process.
However, this method is limited to the single image deblurring, which cannot be generalized to the multi-frames (video) deblurring problem.
%

%
%
As the dynamic scene blur is mainly caused by moving objects, it is natural to use the optical flow information to model the blur and guide the image restoration.
Motivated by this, we propose a dynamic scene deblurring algorithm based on optical flow which is used to model the blur and provide the information to generate the weights of RNNs to deblur images.
Specifically, we propose to use FlowNets~\cite{dosovitskiy2015flownet} to estimate the optical flow from two successive blurry images.
At the same time, FlowNets also provides RNN weights at different scales.
In \cite{zhang2018dynamic}, there are only four convolution layers in total before and after RNNs which prevents their network from utilizing deep information when removing blur.
Our deblurring network is based on a U-net structure~\cite{ronneberger2015u}, and we put RNNs in the different scales of encoder part.
In this way, RNNs can remove blur from features containing tiny structures in shallow layers and use the subsequent layers to reconstruct a sharp image with fewer artifacts.

We summarize the contributions of this work as follows:
\begin{compactitem}
	\item We propose an effective dynamic scene deblurring algorithm based on optical flow and spatially variant RNNs, where the optical flow is used to guide the spatially variant RNNs for the blur removal. Our network is trained in an end-to-end manner.
	\item We mathematically analyze the relationship between optical flow and blur kernel and show that using optical flow is able to help the dynamic scene deblurring.
	\item We both quantitatively and qualitatively evaluate our algorithm on the public available benchmark datasets and show that it performs favorably against state-of-the-art algorithms in terms of accuracy, speed as well as model size.
\end{compactitem}

\vspace{-2mm}
\section{Related Work}
\vspace{-1mm}
Image deblurring has long been a research topic in the field of computer vision.
Early researchers assume a uniform blur kernel and utilize hand-crafted statistical priors~\cite{fergus2006removing, krishnan2009fast, xu2013unnatural, pan2016blind} or the selection of salient edges~\cite{shan2008high, cho2009fast, xu2010two} to solve the ill-posed deconvolution problem.

For real-world blurry images under dynamic scenes, the above uniform assumption does not always hold because of the object motions and abrupt depth variations.
To handle the non-uniform blur, some researchers segment the images into different regions and estimate a different blur kernel in every region of the blurry images.
Kim~et al.~\cite{kim2013dynamic} compute a weight map, which represents the likelihoods that a pixel's blurriness belongs to one of the given blur models.
Pan~et al.~\cite{pan2016soft} estimate a binary mask and a segmentation confidence map for each layer and add regularization terms on both of them in the objective function of the optimization scheme.
Another popular way is to estimate a dense motion field and a pixel-wise blur kernel by its motion without segmentation \cite{kim2014segmentation}.
The pixel-wise blur kernel is usually assumed to be linear to strike a balance between computational cost and model expressiveness.

Other methods utilize multiple images to solve dynamic scene deblurring problem from video or burst images.
%
The methods \cite{wulff2014layers,ahn2016occlusion} use optical flow information to segment layers with different blur and estimate the blur layer-by-layer.
When depth information is available, pixel-wise blur kernels can be estimated by homography \cite{sellent2016stereo, pan2017simultaneous}.
According to \cite{kim2015generalized}, there is a connection between blur kernels and optical flow.
Their methods can iteratively optimize optical flow and blur kernels.
In \cite{cho2012patch}, they assume that not all frames are equally blurry and use the clean patches in nearby frames to restore blurry ones.
Even though the above methods have promising performance, they are slow because of the complex optimization process.
Delbracio~et al.~\cite{delbracio2015burst} deblur images by merging multiple registered burst images in the frequency domain.
While this algorithm is efficient, it nevertheless cannot work well if all the input images are blurry.

With the development of deep learning in recent years, it has been widely used to solve dynamic scene deblurring problems.
\cite{sun2015learning} and \cite{gong2017motion} propose to utilize neural networks to estimate blur kernels.
However, they can only predict linear blur kernels which are inaccurate under some scenarios.
Also, a non-blind deblurring algorithm, such as \cite{zoran2011learning}, is needed to restore the sharp images with estimated blur kernels and it has a high computational burden.
\cite{noroozi2017motion, nah2017deep, tao2018scale} train end-to-end neural networks to directly restore the sharp images.
As a large receptive field is needed for deblurring, a multi-scale scheme is used in their networks.
The adversarial loss is used in \cite{nah2017deep, kupyn2017deblurgan} to generate sharper and more visually pleasing results.
Su~et al.~\cite{su2017Deep} propose an end-to-end neural network to deblur dynamic scene videos.
To share information across frames, Kim~et al.~\cite{kim2017online} propose to use recurrent structure to propagate the deblurred features from the previous frames into those of the current one.
Kim~et al.~\cite{kim2018spatio} apply a spatial-temporal transformer network to align the frames and improve the performance of existing deep learning based video deblurring algorithms.
By repeatedly exchanging information between the features from different frames, Aittala~et al.~\cite{Aittala2018burstdeblur} propose an end-to-end burst deblurring network.

To enlarge the receptive field and deal with dynamic scene blur, spatially variant RNNs~\cite{liu2016learning} are used in \cite{zhang2018dynamic}.
They utilize a neural network to estimate pixel-wise RNN weights and use RNNs to remove spatially variant blur in feature space.
In addition, as information can be propagated for a long distance by RNNs, their network can have a large receptive field without a multi-scale scheme.
According to their analysis, the RNN weights, which act as blur kernels, have a connection with optical flow.
However, their RNN weights are estimated from a single blurry image which is inaccurate.
Optical flow information has been successively used to estimate non-uniform blur kernels \cite{kim2015generalized}.
But they need complicated optimization with a large computational burden for non-blind deconvolution to restore the clean images.
Also, the inaccurately estimated blur kernel will lead to artifacts.
In \cite{chen2018reblur2deblur}, they constrain the output of their neural network by reblurring it according to the optical flow and train their network in a semi-supervised manner.
%
%
Optical flow information has also been considered as the input of the network to aid deblur process in \cite{yuan2020efficient,yan2020vdflow}.
\cite{yan2020vdflow} directly concatenates the features, which further estimate the optical flow, in the deblurring network.
However, the dynamic scene deblurring is a spatially variant process, simply using concatenation operator is less effective.
In \cite{yuan2020efficient}, it utilizes optical flow information to estimate the offsets in deformable convolution which is further used to remove dynamic scene blur.
Whereas, the receptive field of deformable convolution is limited, which is important for deblurring, compared to the spatially variant RNN used in our network.
In addition, only one blurry image is used in \cite{yuan2020efficient} to extract the optical flow information which is not as accurate as from two consecutive frames.

\begin{figure*}[!t]
	\centering
	\includegraphics[width=0.92\linewidth]{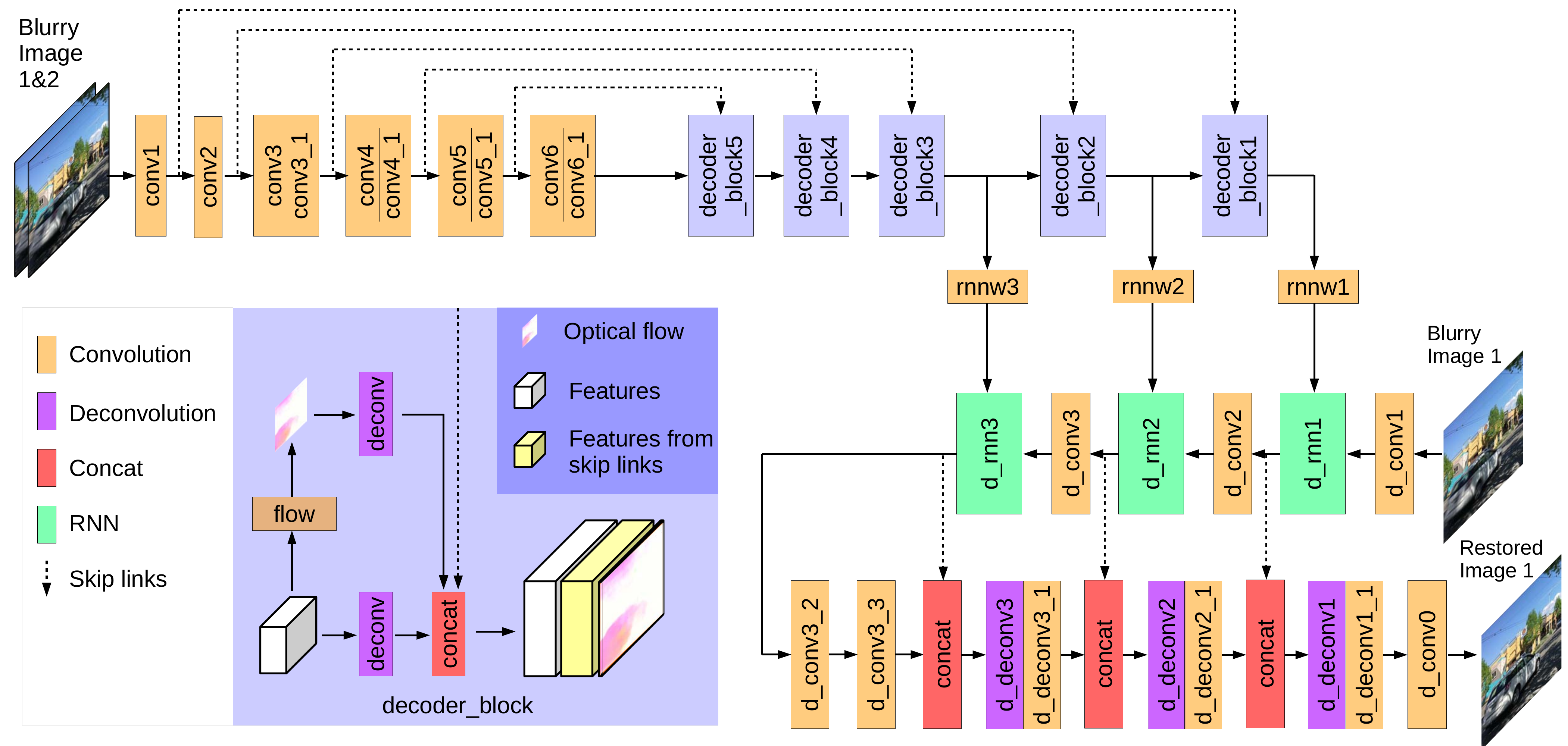}
	\vspace{-4mm}
	\caption{
		An overview of the proposed network.
		The network consists of two parts: optical flow estimation based on FlowNets~\cite{dosovitskiy2015flownet} and image deblurring based on spatially variant RNNs~\cite{liu2016learning}.
		For optical flow estimation network, it additionally predicts the RNN weights in the last three scales.
		The deblurring network extracts features from one blurry input and removes blur from these features by spatially variant RNNs.
		The prefix `d\_' denotes the deblurring network and please see Table~\ref{table:framework} for more details.
	}
	\label{fig:frame}
	\vspace{-3mm}
\end{figure*}


Motivated by our conference paper~\cite{zhang2018dynamic}, we use spatially variant RNNs for dynamic scene deblurring, where optical flow is explored and used to estimate the RNN weights.
Different from existing multiple images deblurring methods, e.g. \cite{kim2015generalized, su2017Deep, kim2017online, kim2018spatio}, which need five images or more as their network inputs, the proposed method only requires two blurry images as inputs in order to estimate the optical flow by FlowNets~\cite{dosovitskiy2015flownet}.
In addition, the proposed algorithm needs less computational cost, and the proposed network has smaller model size.

\vspace{-4mm}
\section{Proposed Method}
\vspace{-1mm}
In this section, we first analyze the relationship between the optical flow and blur kernel.
Then we demonstrate the proposed network structure, which uses FlowNets~\cite{dosovitskiy2015flownet} to estimate the optical flow as well as RNN weights in different scales and applies the spatially variant RNNs~\cite{liu2016learning} to remove the blur in feature space by utilizing the flow information.
After that, we present the loss function we used to constrain the estimated optical flow as well as the restored sharp image.
At last, we describe our training configurations.

\vspace{-1mm}
\subsection{Motivation}\label{subsec:motivation}
\vspace{-1mm}
To better motivate our work, we first analyze the relationship between optical flow and blur kernel. Then we discuss the effect of optical flow on image deblurring.
%

\vspace{-2mm}
{\flushleft \bf{Optical flow and blur kernel.}}
The blur in the dynamic scene is spatially variant due to the camera and objects random motions. Thus, it cannot be modeled by the idealized linear convolution model~\cite{pan2016blind} with the spatially invariant blur kernel.
We note that the blurred image can be modeled by the integration of the successive sharp images during the exposure time according to~\cite{levin2008motion}.
Thus, the blur process at location $(x,y)$ can be modeled as:

\begin{equation}
\label{eq:bm1}
B(x,y)=\int_0^\tau \frac{I(x-s_x(t),y-s_y(t))}{\tau}dt,
\end{equation}
where $B$ and $I$ denote the blurred image and sharp image; $\tau$ denotes the exposure time; $(s_x(t),s_y(t))$ denotes the integration curve.
As demonstrated by~\cite{kim2014segmentation,kim2015generalized,ben2003motion}, using optical flow is able to approximate the integration curve, Thus, the blur process can be rewritten as

\begin{equation}
\label{eq:bm2}
B(x,y)=\int_0^\tau \frac{I(x-u(x,y)t,y-v(x,y)t)}{\tau}dt,
\vspace{-1mm}
\end{equation}
where $(u(x,y),v(x,y))$ denotes the velocity between the successive images at location $(x,y)$ and we use $(u,v)$ for short in the following manuscript.
%
%
If we introduce an auxiliary variable $s$, $B$ becomes:

\begin{equation}
\label{eq:bm3}
\begin{split}
B(x,y)=\int_0^\tau\int_{0^-}^{0^+} \frac{I(x-ut-vs,y-vt+us)\delta(s)}{\tau}dsdt,
\end{split}
\vspace{-1mm}
\end{equation}
in which $\delta(\cdot)$ is the Dirac delta function and $\int_{0^-}^{0^+} I(x-ut-vs,y-vt+us)\delta(s)ds=I(x-ut,y-vt)$.
Let $p$ and $q$ denote $ut+vs$ and $vt-us$, we have

\begin{equation}
\label{eq:bm3-t}
\begin{split}
t = \frac{up+vq}{u^2+v^2},
\end{split}
\end{equation}
and
\begin{equation}
\label{eq:bm3-s}
\begin{split}
s=\frac{vp-uq}{u^2+v^2}.
\end{split}
\vspace{-1mm}
\end{equation}
Since $0\leqslant t\leqslant\tau$ and $0^-\leqslant s\leqslant 0^+$, we can get $0\leqslant p\leqslant u\tau$ and $(\frac{pv}{u})^-\leqslant q\leqslant (\frac{pv}{u})^+$.
We note that the determinant of Jacobian matrix $|\frac{\partial(s,t)}{\partial(p,q)}|$ and $\delta(s)$ are

\begin{equation}
\label{eq:bm3-Jacobian}
\begin{split}
\left|\frac{\partial(s,t)}{\partial(p,q)}\right|=\frac{1}{u^2+v^2},
\end{split}
\vspace{-1mm}
\end{equation}

and
\begin{equation}
\label{eq:bm3-delta}
\begin{split}
\delta(s)=\delta(\frac{vp-uq}{u^2+v^2})=(u^2+v^2)\delta(vp-uq).
\end{split}
\vspace{-1mm}
\end{equation}

Based on these two equations, $B$ can be represented by:
\begin{equation}
\label{eq:bm4}
\begin{split}
B(x,y)=\int_{0}^{u\tau}\int_{(\frac{pv}{u})^-}^{(\frac{pv}{u})^+} \frac{I(x-p,y-q)\delta(vp-uq)}{\tau}dqdp.
\end{split}
\vspace{-1mm}
\end{equation}

Since $\delta(vp-uq)\neq 0$ if and only if $q=\frac{pv}{u}$ and $0\leqslant u$, $0\leqslant v$, $0\leqslant \tau$, we can relax the range of $q$ into $[0,v\tau]$. In this way. $B$ becomes:

\begin{equation}
\label{eq:bm5}
\begin{split}
B(x,y)=\int_{0}^{u\tau}\int_{0}^{v\tau} \frac{I(x-p,y-q)\delta(vp-uq)}{\tau}dqdp.
\end{split}
\vspace{-1mm}
\end{equation}
Note that~\eqref{eq:bm5} can be represented by a 2D-convolution formulation as:

\begin{equation}
\label{eq:bm6}
\begin{split}
B(x,y)=\int_{0}^{u\tau}\int_{0}^{v\tau} I(x-p,y-q)k(p,q)dqdp,
\end{split}
\vspace{-1mm}
\end{equation}
in which $k$ is the pixel-wise blur kernel which is shown as:

\begin{equation}
\label{eq:kernel}
k=\frac{\delta(vp-uq)}{\tau}.
\vspace{-1mm}
\end{equation}
At the same time, if the time interval between the successive images is $T=\tau/r$, where $r$ is the camera duty circle, the optical flow $F(x,y)$ between these two images at position $(x,y)$ is $F(x,y)=(F_p(x,y),F_q(x,y))=(u(x,y)T,v(x,y)T)$ and we also use $(F_p,F_q)$ for short. Then we can get $\delta(vp-uq)=T\delta(F_qp-F_pq)$.
As a result, the pixel-wise blur kernel can be represented by pixel-wise optical flow at position $(x,y)$ as:

\begin{equation}
\label{eq:kernelflow}
k(x,y;p,q)=\frac{\delta(F_qp-F_pq)}{r}.
\vspace{-1mm}
\end{equation}

\vspace{-2mm}
{\flushleft \bf{Spatially variant RNN as the deconvolution.}}
In \cite{liu2016learning}, a spatially variant RNN is proposed to solve low-level vision tasks.
To apply RNNs in a two-dimensional feature, they use four one-dimensional RNNs in different directions which are down to up, up to down, right to left and left to right.
One-dimensional RNN in y-axis can be written as:
\begin{equation}
\label{eq:RNN}
g(x,y) = (1-w(x,y))f(x,y) + w(x,y)g(x,y-1),
\vspace{-1mm}
\end{equation}
where $f$ and $g$ are the input and output feature maps of an RNN, $w$ is the estimated spatially variant RNN weights.
The rest RNNs in other directions can be described in a similar way.

According to the analysis of \cite{zhang2018dynamic}, a large receptive field is needed to remove even a small blur and RNN is satisfied this requirement with a small number of parameters.
In addition, the blur is spatially variant in dynamic scenes and the network should treat different blur by different operations.
Zhang~et al.~\cite{zhang2018dynamic} use a network to estimate the spatially variant RNN weights and then utilize these RNNs~\cite{liu2016learning} to remove blur in the extracted feature space.
Based on ~\eqref{eq:kernelflow}, there is a connection between optical flow and blur kernel, it is more convenient to estimate the RNN weights $w$ by using the optical flow information and then guide the dynamic scene deblurring.
In this paper, we utilize FlowNets~\cite{dosovitskiy2015flownet} to estimate optical flow and RNN weights simultaneously in different scales.
In this way, the optical flow and RNN weights are related and the spatially variant RNNs can implicitly utilize the optical flow information to remove dynamic scene blur in the feature space.

\vspace{-2mm}
\subsection{Network Architecture}\label{subsec:network}
\vspace{-1mm}

The proposed network consists of two parts: optical flow estimation and image deblurring as shown in Figure~\ref{fig:frame}.

For the optical flow estimation part, we can use existing network architectures designed for optical flow, e.g., Flownet~\cite{dosovitskiy2015flownet}, FlowNet 2.0~\cite{ilg2017flownet} and PWC-net~\cite{sun2018pwc}.
In this paper, we use a similar network architecture to FlowNets~\cite{dosovitskiy2015flownet} for the optical flow estimation, where the network contains a U-net structure~\cite{ronneberger2015u} with skip connections.
We use this sub-network to estimate the optical flow of the input two blurry images in five different scales.
As these features contain the optical flow information and optical flow is related to the blur kernel according to the above section, we additionally add a convolution layer (`rnnw1', `rnnw2' and `rnnw3' in Figure~\ref{fig:frame}) in each of the last three scales to estimate the RNN weights.
A hyperbolic tangent (tanh) layer is added after every RNN weights estimation convolution layer to constrain the weights between -1 to 1 just as in \cite{liu2016learning}.

For the image deblurring network, we adopt the spatially variant RNN which is proposed by~\cite{zhang2018dynamic} with some modifications.
In \cite{zhang2018dynamic}, there are only two convolution layers before RNNs.
The reason is that a deep neural network will extract sparse information, e.g., edges, which is not suitable for RNN to propagate information since most of the pixels in the feature maps are zero.
In order to utilize a deeper image deblurring network to achieve better results, the proposed one is based on the U-net structure and the RNNs are at the first three scales in the encoder other than the encoded features with the smallest scale as in \cite{zhang2018dynamic}.
In addition, no non-linear activation layer, e.g., ReLU, is added in the encoder to prevent it from generating sparse features.
The detailed parameters of our network are shown in Table~\ref{table:framework} and implementation can be found in \url{https://sites.google.com/site/zhjw1988}.

\begin{table*}[]
\footnotesize
	\centering
	\caption{
		Configurations of our network which contains two parts: the optical flow estimation network and deblurring network.
		`conv', `deconv', `rnn', `ReLU' and `tanh' denote the convolution layer, deconvolution layer, spatially-variant RNN layer, leaky ReLU layer with negative slope 0.1 and tanh layer, respectively.
	}
	\vspace{-3mm}

	Optical Flow Estimation Network

	\begin{tabular}{cccccc}
		\toprule
		Name & Layer & Kernel & Str. & Ch. I/O & Input\\
		\midrule
		conv1    &  conv+ReLU  & 7$\times$7 & 2 & 6/24    & image1\&image2\\
		conv2    &  conv+ReLU  & 5$\times$5 & 2 & 24/48   & conv1\\
		conv3    &  conv+ReLU  & 5$\times$5 & 2 & 48/96   & conv2\\
		conv3\_1 &  conv+ReLU  & 3$\times$3 & 1 & 96/96   & conv3\\
		conv4    &  conv+ReLU  & 3$\times$3 & 2 & 96/192  & conv3\_1\\
		conv4\_1 &  conv+ReLU  & 3$\times$3 & 1 & 192/192 & conv4\\
		conv5    &  conv+ReLU  & 3$\times$3 & 2 & 192/192 & conv4\_1\\
		conv5\_1 &  conv+ReLU  & 3$\times$3 & 1 & 192/192 & conv5\\
		conv6    &  conv+ReLU  & 3$\times$3 & 2 & 192/384 & conv5\_1\\
		conv6\_1 &  conv+ReLU  & 3$\times$3 & 1 & 384/384 & conv6\\
		\midrule
		flow6     &  conv  & 3$\times$3 & 1 & 384/2   & conv6\_1\\
		up6to5    & deconv & 4$\times$4 & 2 & 2/2     & flow6\\
		deconv6   & deconv+ReLU & 4$\times$4 & 2 & 384/192 & conv6\_1\\
		
		flow5     &  conv  & 3$\times$3 & 1 & 386/2   & conv5\_1$+$deconv6$+$up6to5\\
		up5to4    & deconv & 4$\times$4 & 2 & 2/2     & flow5\\
		deconv5   & deconv+ReLU & 4$\times$4 & 2 & 386/96  & conv5\_1$+$deconv6$+$up6to5\\
		
		flow4     &  conv  & 3$\times$3 & 1 & 290/2   & conv4\_1$+$deconv5$+$up5to4\\
		up4to3    & deconv & 4$\times$4 & 2 & 2/2     & flow4\\
		deconv4   & deconv+ReLU & 4$\times$4 & 2 & 290/48  & conv4\_1$+$deconv5$+$up5to4\\
		
		flow3     &  conv  & 3$\times$3 & 1 & 146/2   & conv3\_1$+$deconv4$+$up4to3\\
		up3to2    & deconv & 4$\times$4 & 2 & 2/2     & flow3\\
		deconv3   & deconv+ReLU & 4$\times$4 & 2 & 146/24  & conv3\_1$+$deconv4$+$up4to3\\
		
		flow2     &  conv  & 3$\times$3 & 1 & 74/2    & conv2$+$deconv3$+$up3to2\\
		up2to1    & deconv & 4$\times$4 & 2 & 2/2     & flow2\\
		deconv2   & deconv+ReLU & 4$\times$4 & 2 & 74/12   & conv2$+$deconv3$+$up3to2\\
		\midrule
		rnnw1      &  conv+tanh  & 3$\times$3 & 1 & 38/96   & conv1$+$deconv2$+$up2to1 \\
		rnnw2      &  conv+tanh  & 3$\times$3 & 1 & 74/192  & conv2$+$deconv3$+$up3to2\\
		rnnw3      &  conv+tanh  & 3$\times$3 & 1 & 146/384 & conv3\_1$+$deconv4$+$up4to3\\
		\bottomrule
	\end{tabular}
	
\vspace{1mm}
	Deblurring Network
\vspace{1mm}

	\begin{tabular}{cccccc}
		\toprule
		Name          & Layer & Kernel     & Str. & Ch. I/O & Input\\
		\midrule
		d\_conv1      &  conv  & 5$\times$5 & 2 & 3/24    & image1\\
		d\_rnn1       &  rnn   & /          & / & 24$+$96/96 & d\_conv1+rnnw1 \\
		d\_conv2      &  conv  & 5$\times$5 & 2 & 96/48   & d\_rnn1\\
		d\_rnn2       &  rnn   & /          & / & 48$+$192/192 & d\_conv2+rnnw2\\
		d\_conv3      &  conv  & 5$\times$5 & 2 & 192/96  & d\_rnn2\\
		d\_rnn3       &  rnn   & /          & / & 96$+$384/384 & d\_conv3+rnnw3\\
		\midrule
		d\_conv3\_2    &  conv+ReLU  & 5$\times$5 & 1 & 384/96  & d\_rnn3\\
		d\_conv3\_3    &  conv+ReLU  & 5$\times$5 & 1 & 96/48   & d\_conv3\_2\\
		d\_deconv3     & deconv+ReLU & 4$\times$4 & 2 & 432/48  & d\_conv3\_3$+$d\_rnn3 \\
		d\_deconv3\_1  &  conv+ReLU  & 5$\times$5 & 1 & 48/48   & d\_deconv3\\
		d\_deconv2     & deconv+ReLU & 4$\times$4 & 2 & 240/24  & d\_deconv3\_1$+$d\_rnn2\\
		d\_deconv2\_1  &  conv+ReLU  & 5$\times$5 & 1 & 24/24   & d\_deconv2\\
		d\_deconv1     & deconv+ReLU & 4$\times$4 & 2 & 120/12  & d\_deconv2\_1$+$d\_rnn1\\
		d\_deconv1\_1  &  conv+ReLU  & 5$\times$5 & 1 & 12/12   & d\_deconv1\\
		d\_conv0       &  conv  & 5$\times$5 & 1 & 15/3    & d\_deconv1\_1$+$image1\\
		\bottomrule
	\end{tabular}
	\label{table:framework}
	\vspace{-3mm}
\end{table*}


\renewcommand{\tabcolsep}{2pt}
\begin{table*}[!t]\footnotesize
	\renewcommand{\arraystretch}{1.1}
	\centering
	\caption{Quantitative evaluation on the video deblurring dataset~\cite{su2017Deep} in terms of PSNR and SSIM.
	}
	\label{table:exp_dvd}
	\vspace{-2mm}
	\begin{tabular}{cccccccccccccc}
		\toprule
		& Whyte \cite{whyte2012non} & Sun \cite{sun2015learning} & Gong \cite{gong2017motion} & Nah \cite{nah2017deep} & Kupyn \cite{kupyn2017deblurgan} & Tao \cite{tao2018scale} & Zhang \cite{zhang2018dynamic} & Kupyn \cite{kupyn2019deblurgan} & Cai \cite{cai2020dark} & Kim \cite{kim2015generalized} & Su \cite{su2017Deep} & Kim  \cite{kim2017online} & ours \\
		\midrule
		PSNR & 25.29 & 27.24 & 28.22 & 29.51 & 26.78 & 29.95 & 30.05 & 29.47 & 29.48 & 27.01 & 30.05 & 29.95 & 30.58 \\
		SSIM & 0.832 & 0.878 & 0.894 & 0.912 & 0.848 & 0.919 & 0.922 & 0.910 & 0.914 & 0.861 & 0.920 & 0.911 & 0.928 \\
		\bottomrule
	\end{tabular}
\vspace{-3mm}
\end{table*}

\renewcommand{\tabcolsep}{1pt}
\begin{figure*}[t]\footnotesize
	\begin{center}
		\begin{tabular}{cccc}
			\includegraphics[width=\swfour]{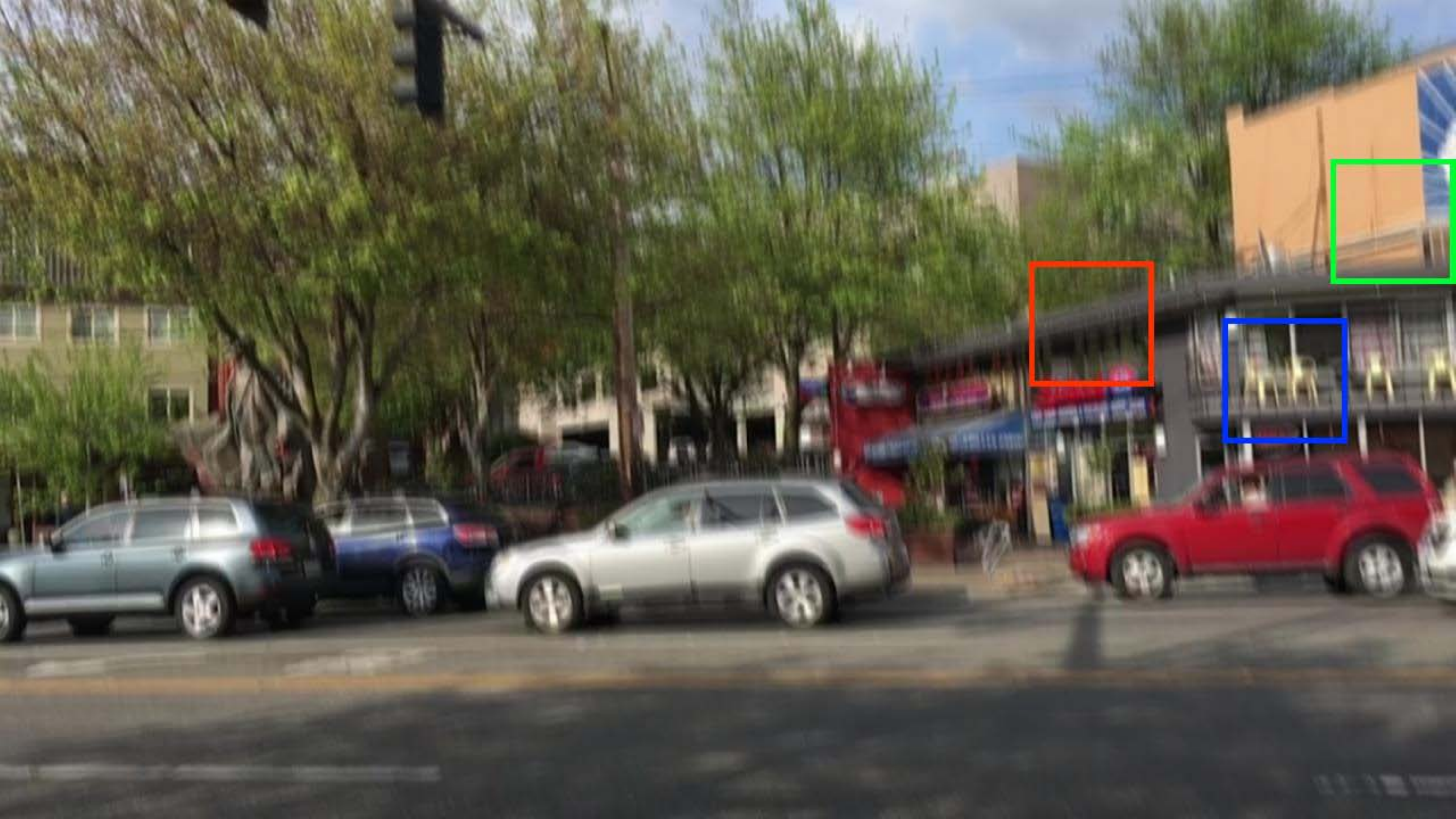} &
			\includegraphics[width=\swfour]{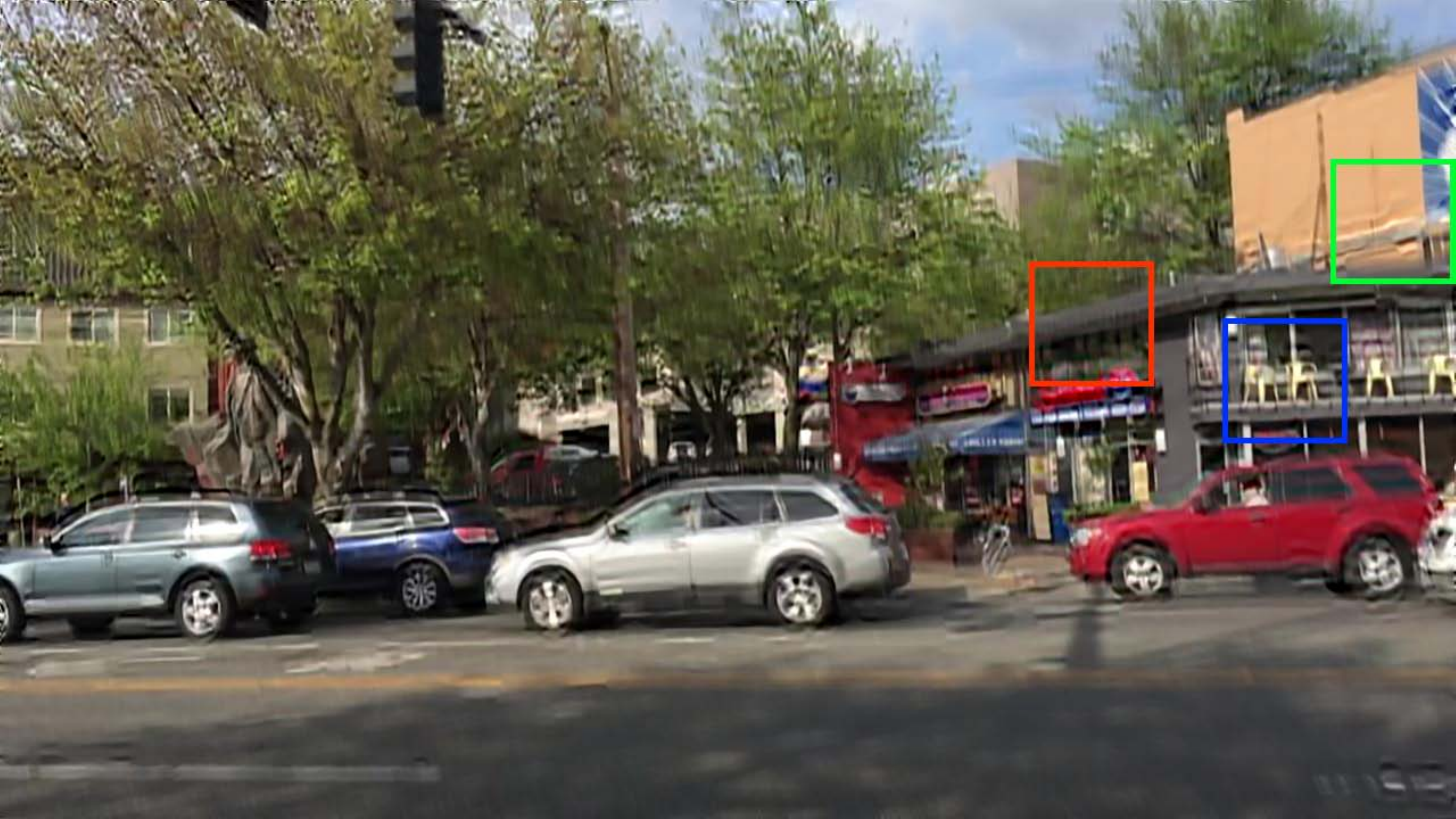} &
			\includegraphics[width=\swfour]{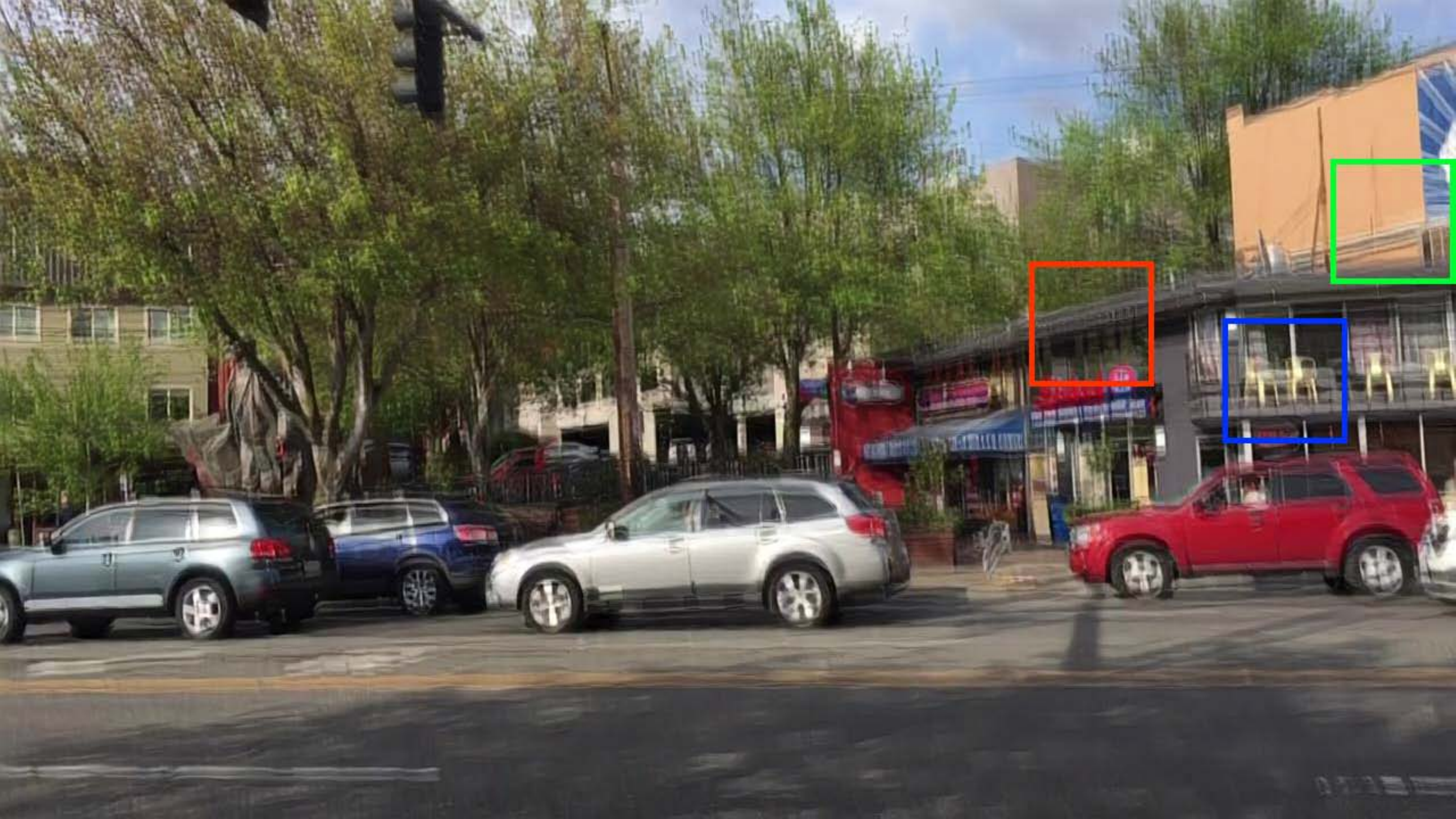} &
			\includegraphics[width=\swfour]{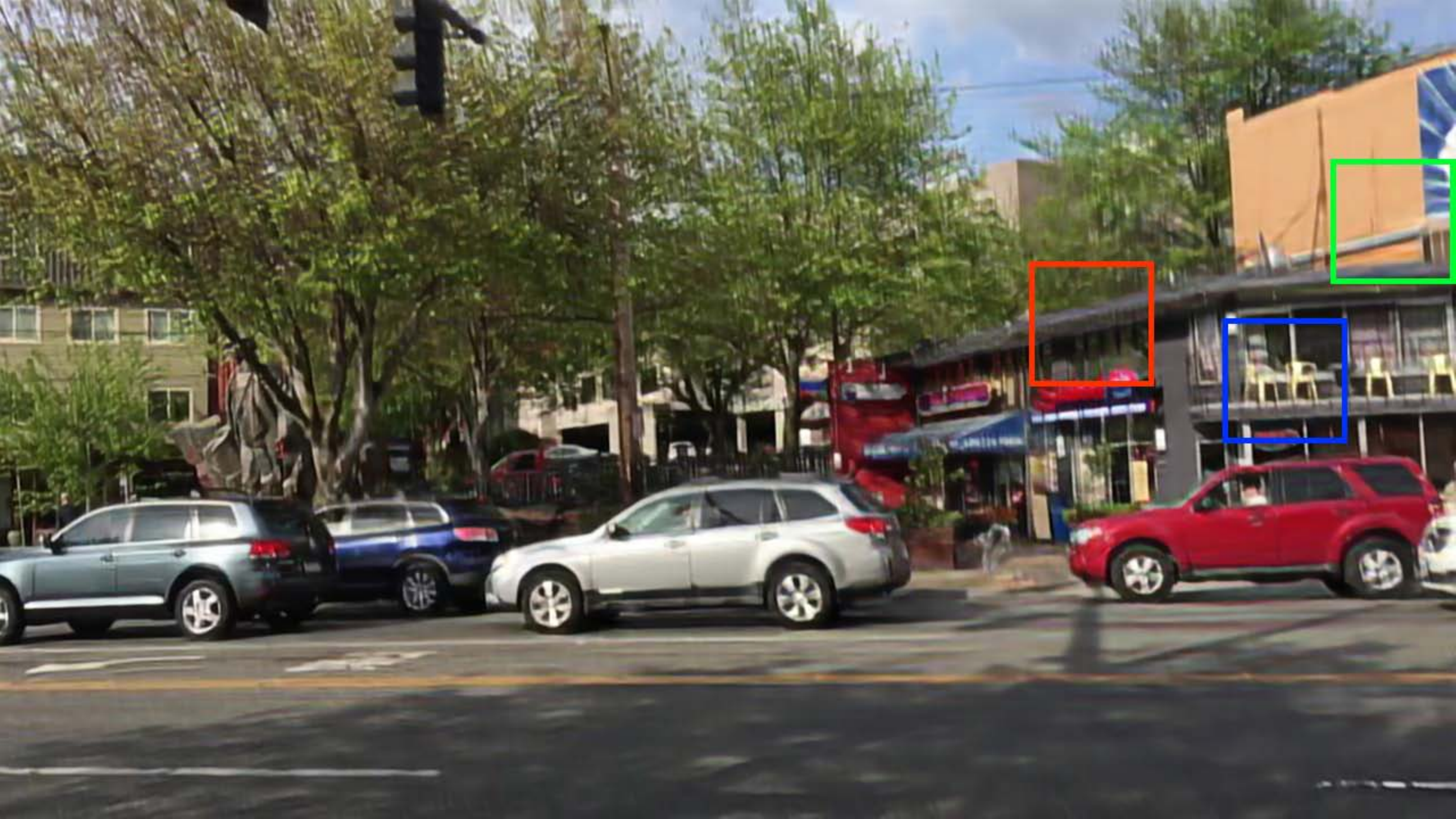} \\
			\includegraphics[width=\swfour]{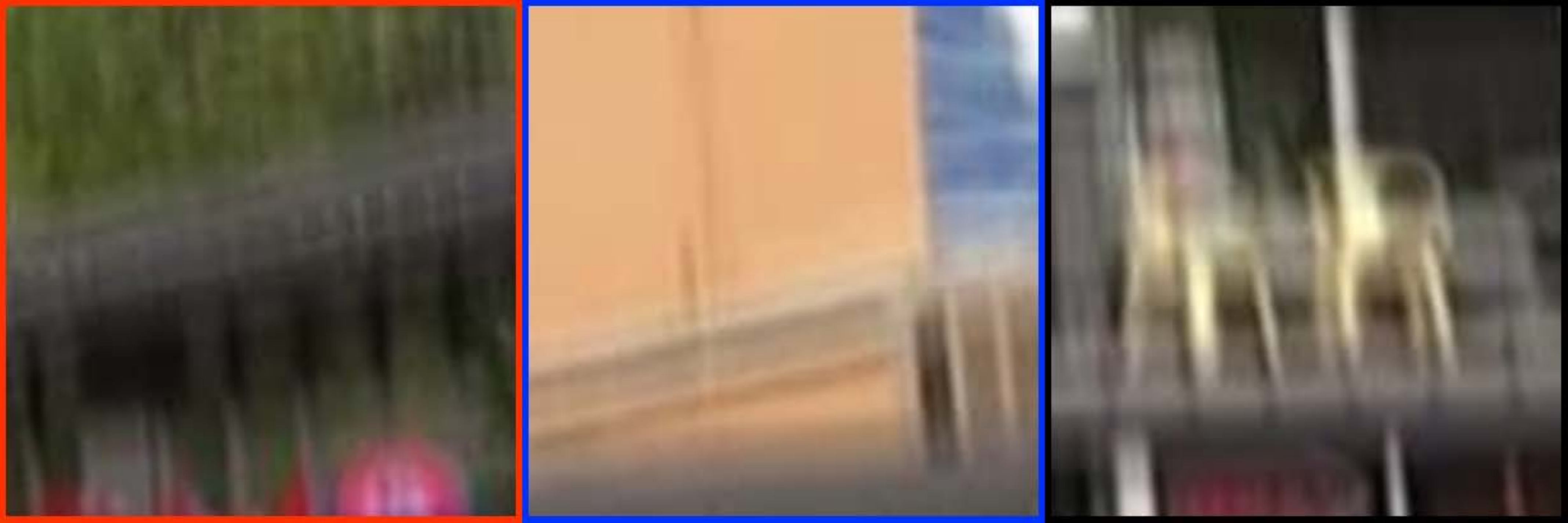} &
			\includegraphics[width=\swfour]{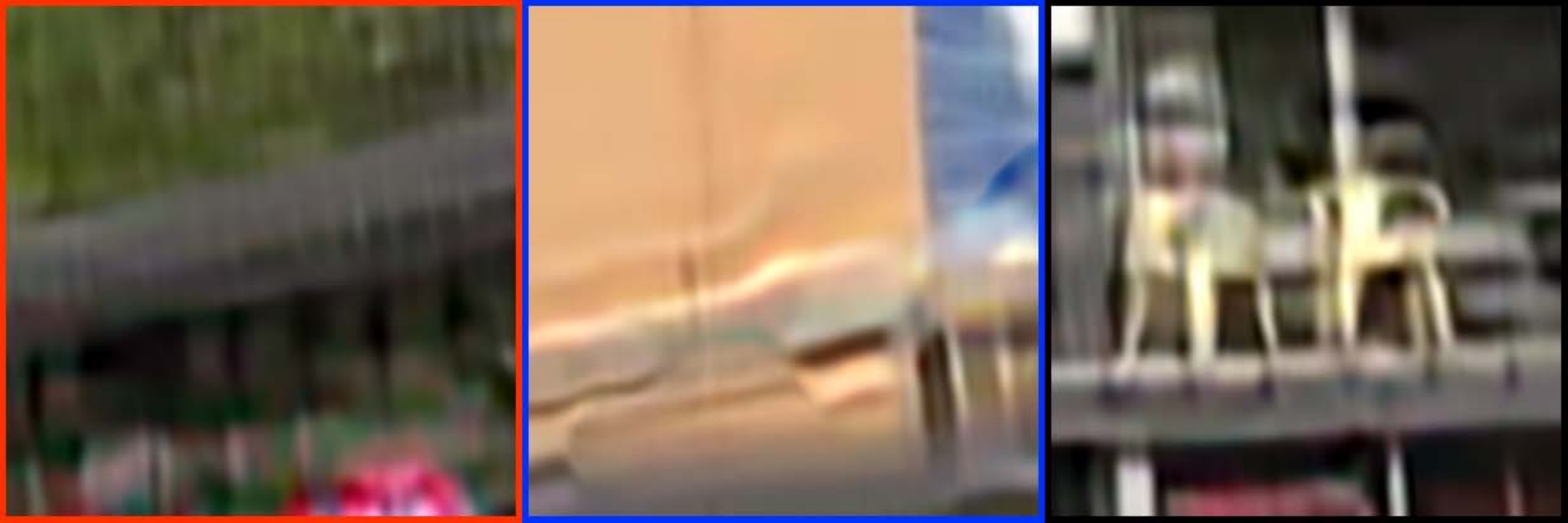} &
			\includegraphics[width=\swfour]{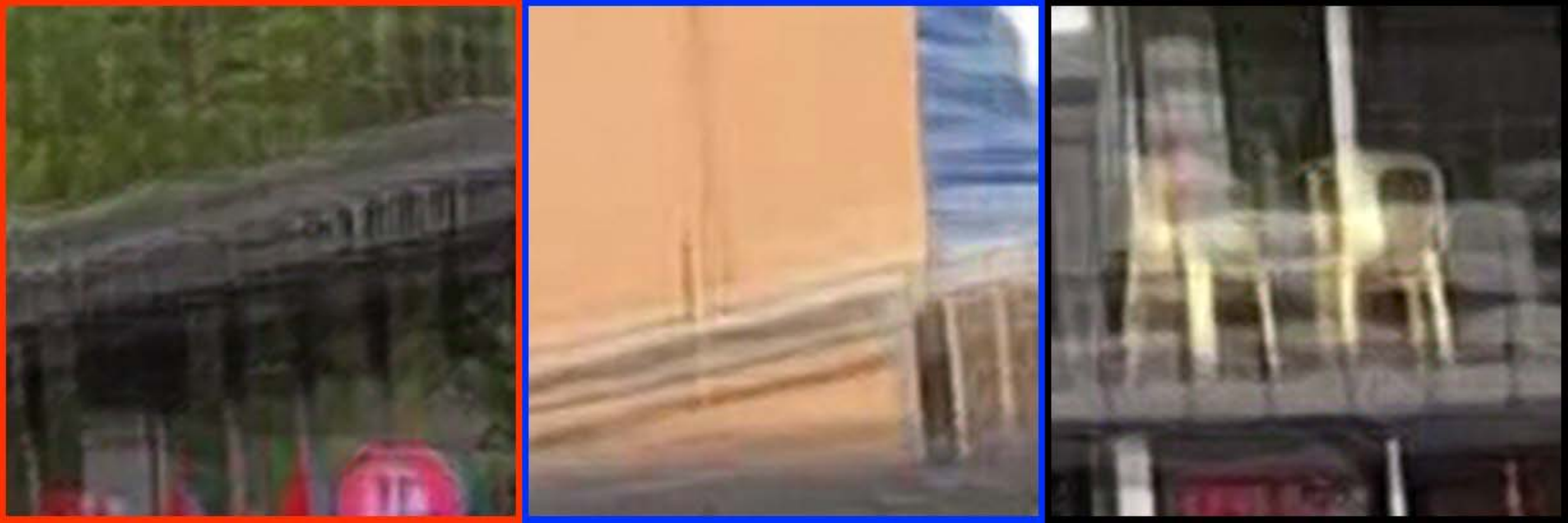} &
			\includegraphics[width=\swfour]{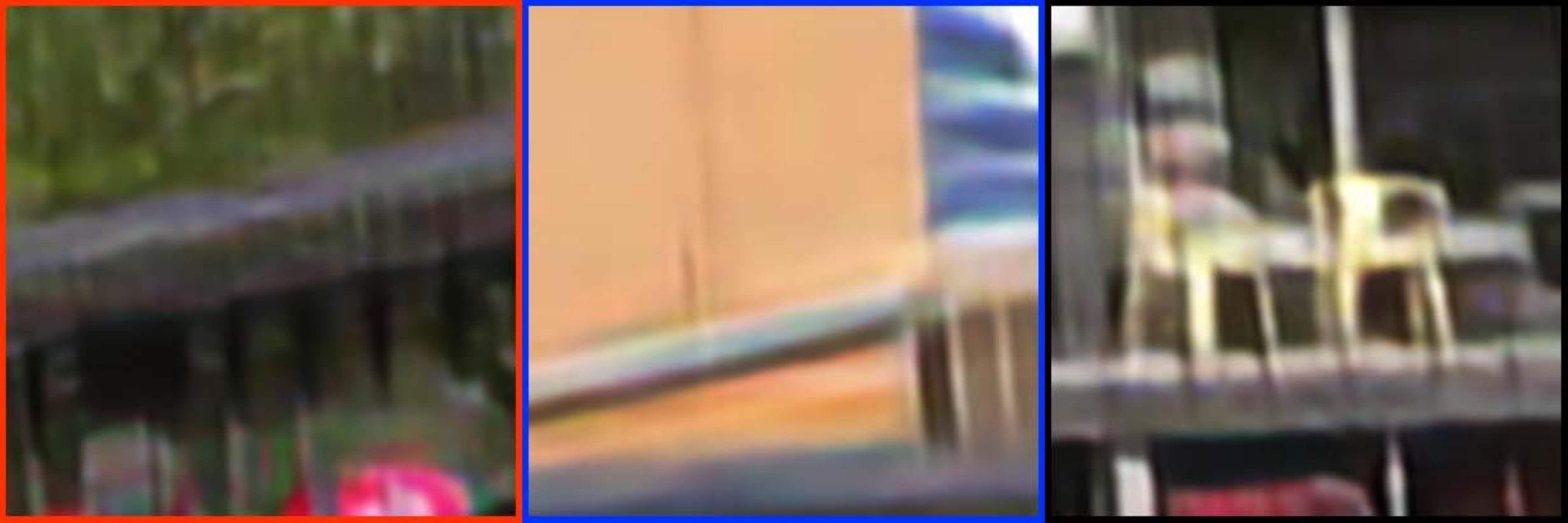} \\
			(a) blurry image & (b) Gong \cite{gong2017motion} & (c) Kupyn \cite{kupyn2017deblurgan} & (d) Tao \cite{tao2018scale} \\
			psnr/ssim & 21.86/0.7291 & (c) 20.67/0.6320 & (d) 23.73/0.7994 \\
			\includegraphics[width=\swfour]{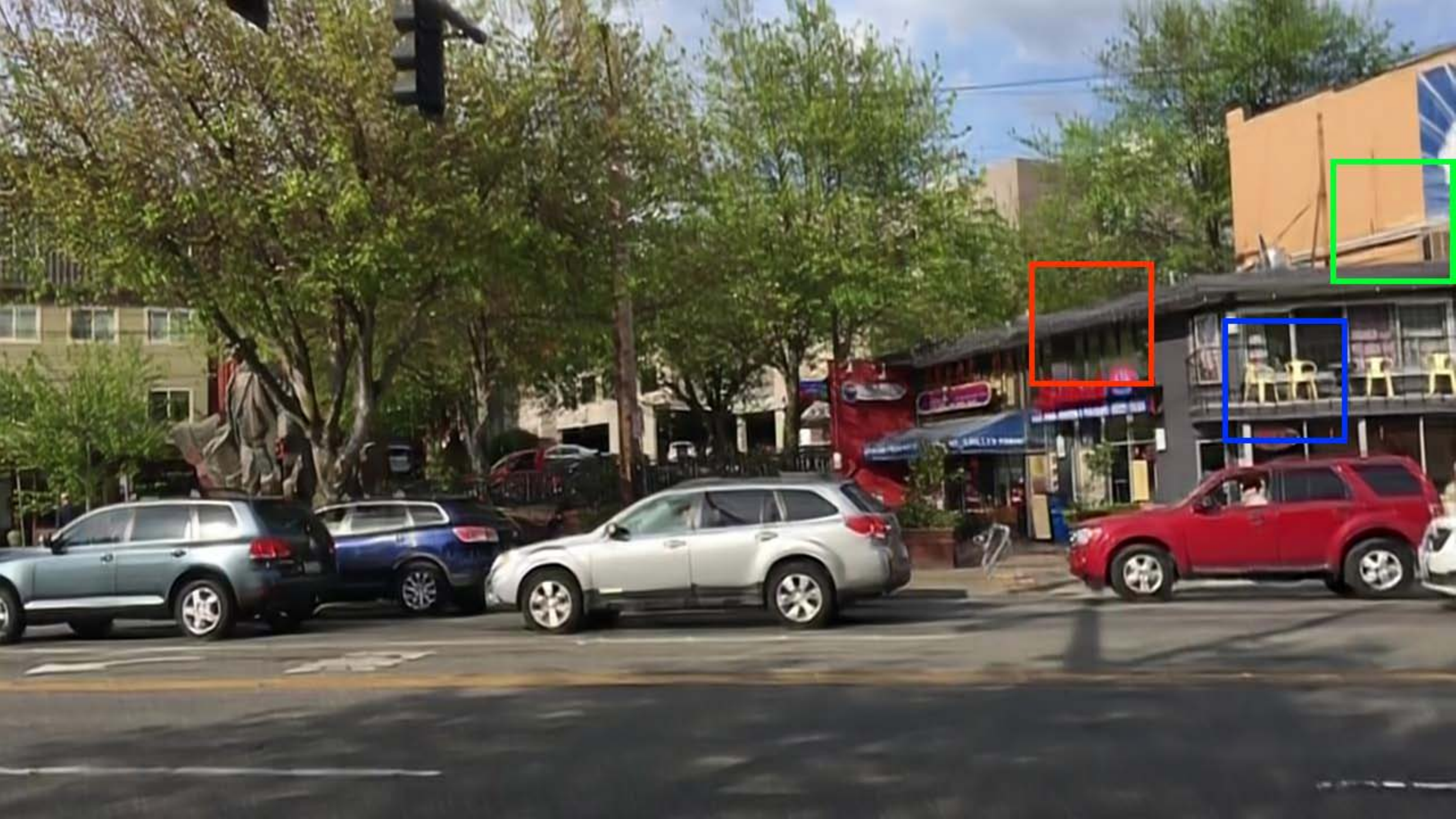} &
			\includegraphics[width=\swfour]{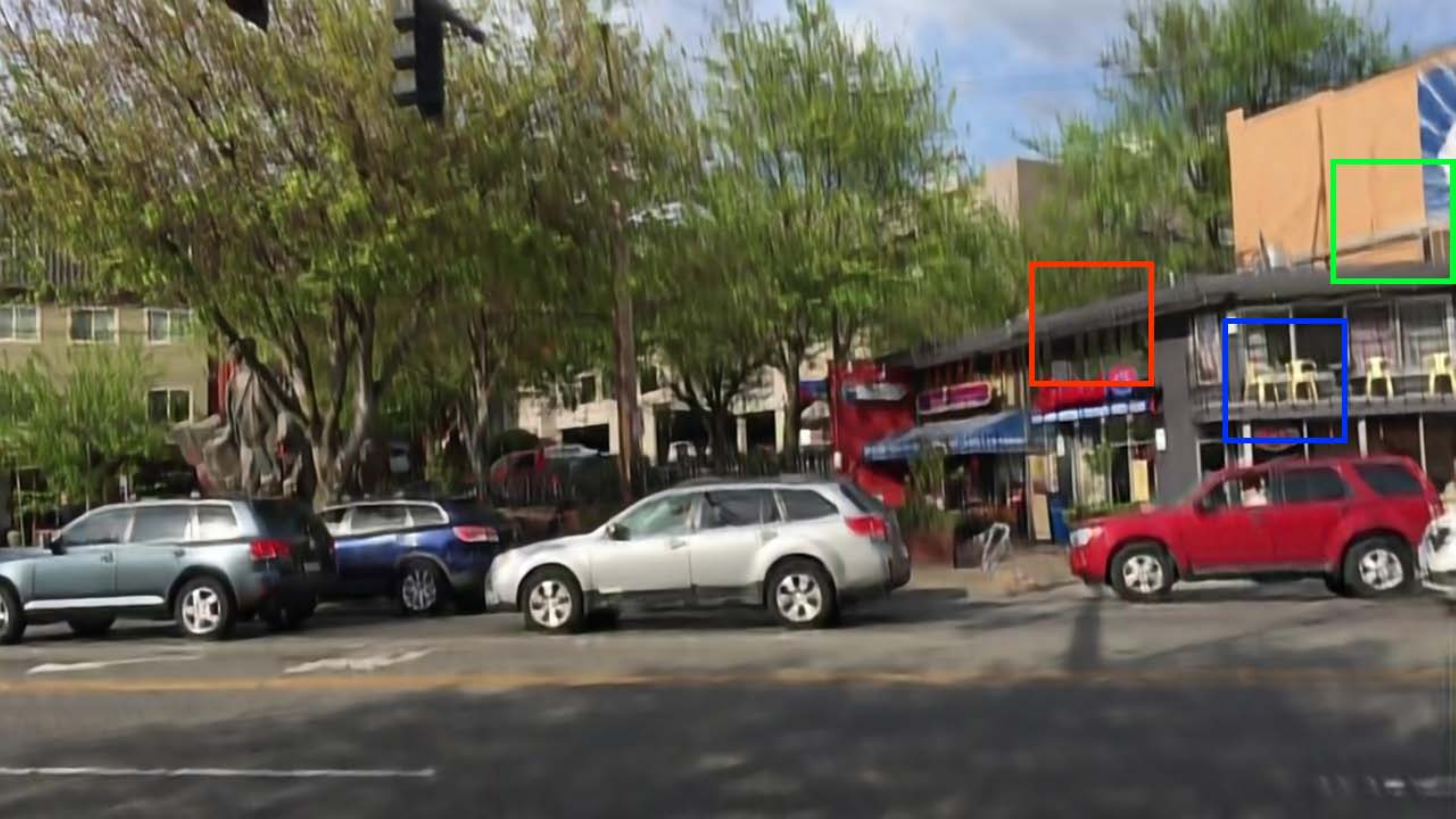} &
			\includegraphics[width=\swfour]{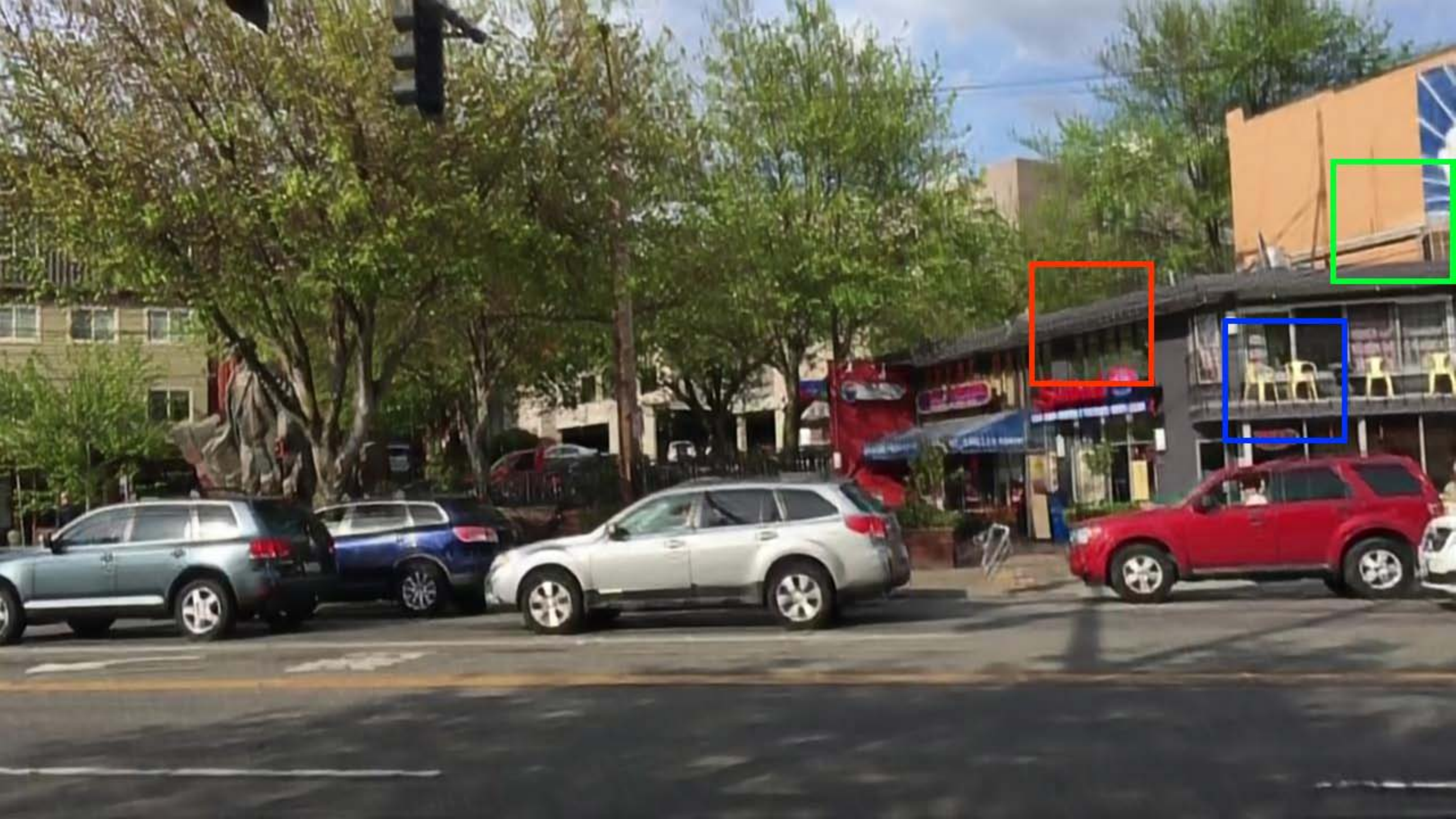} &
			\includegraphics[width=\swfour]{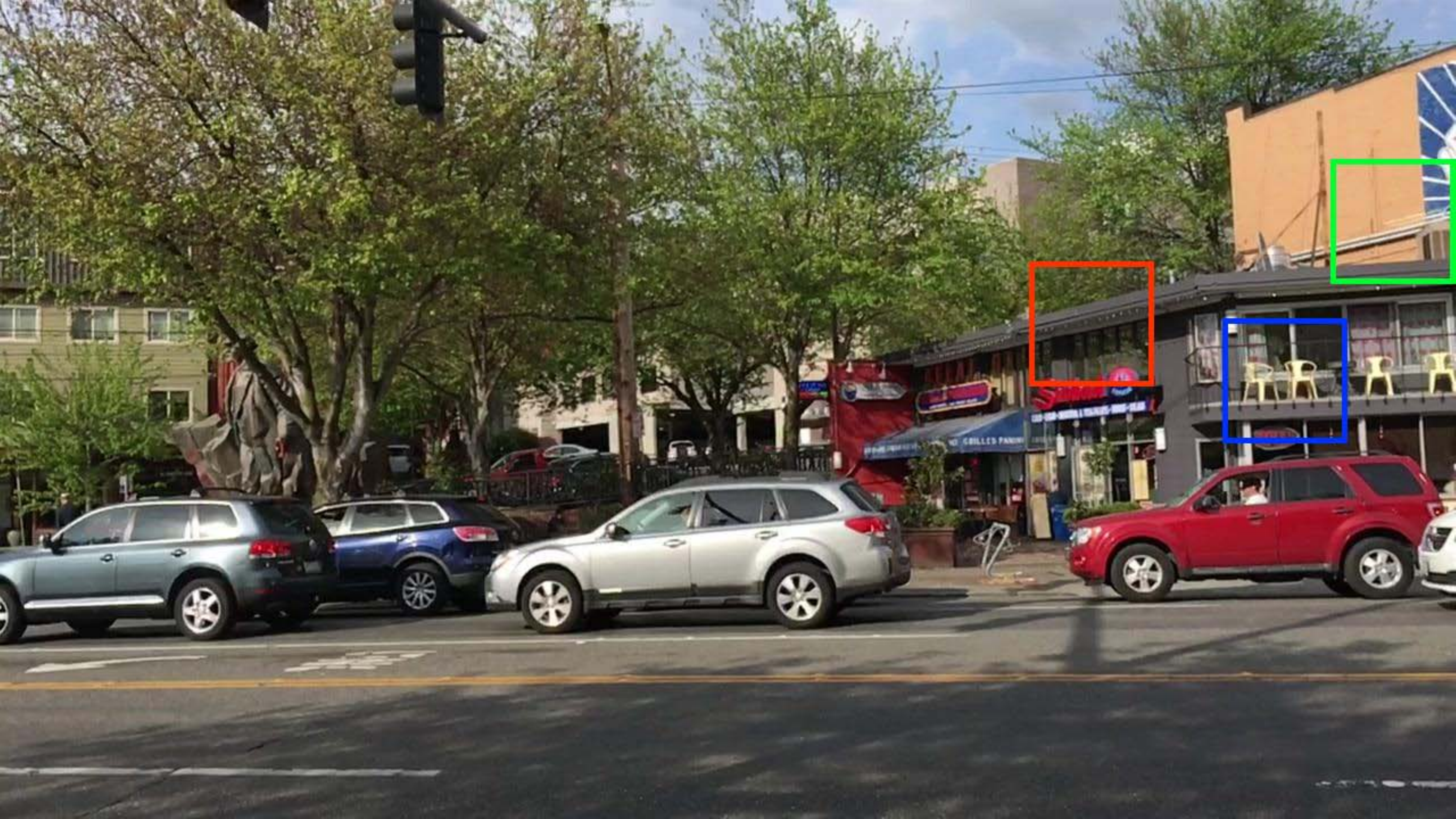} \\
			\includegraphics[width=\swfour]{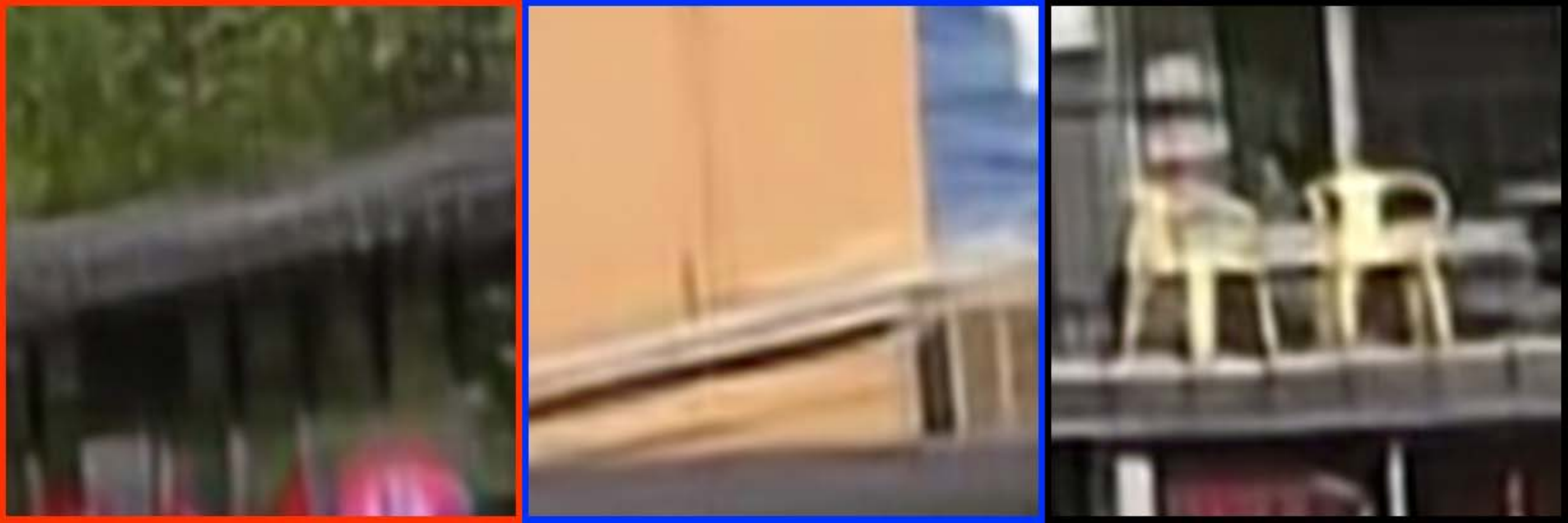} &
			\includegraphics[width=\swfour]{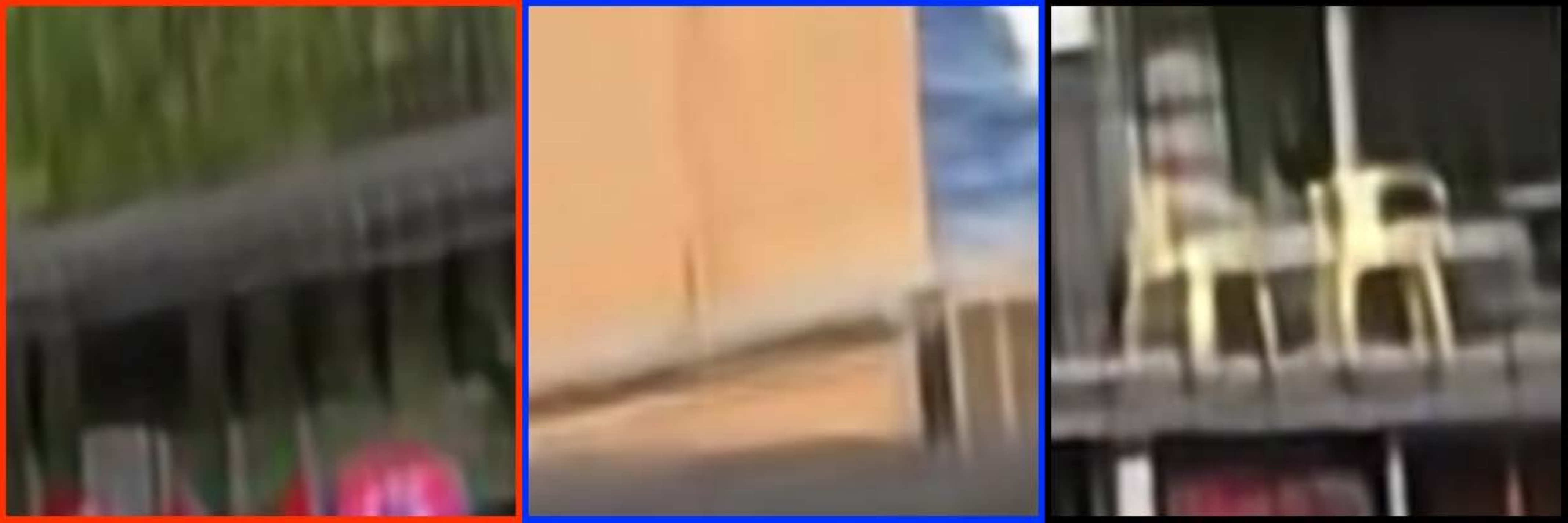} &
			\includegraphics[width=\swfour]{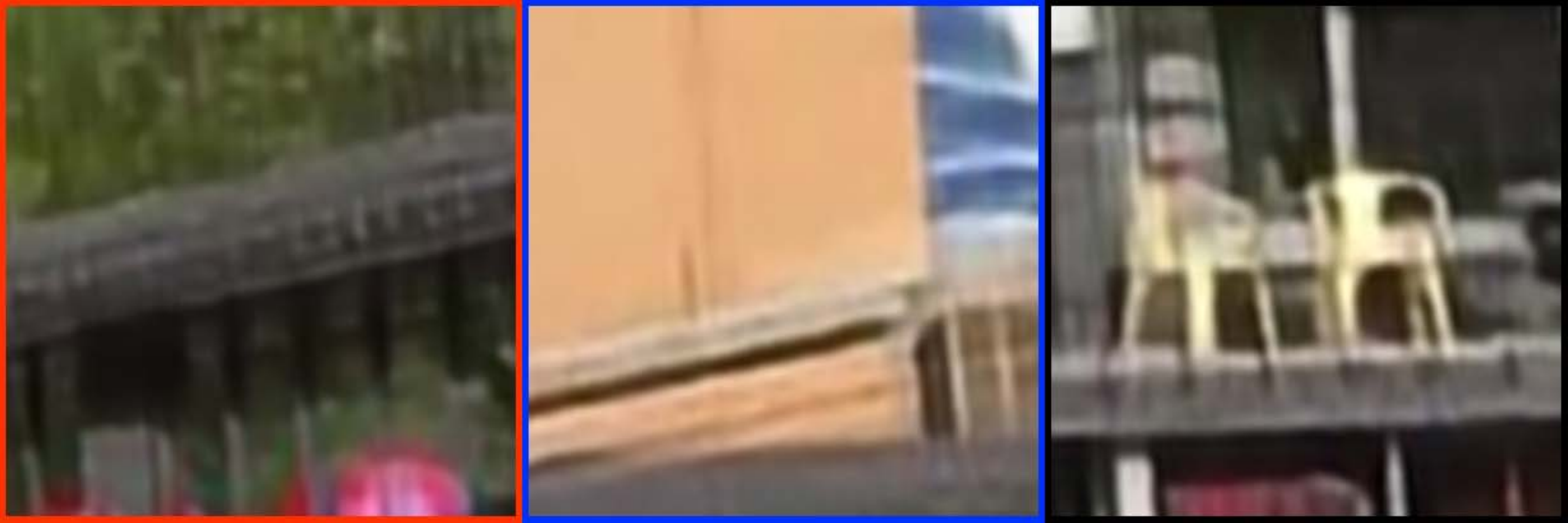} &
			\includegraphics[width=\swfour]{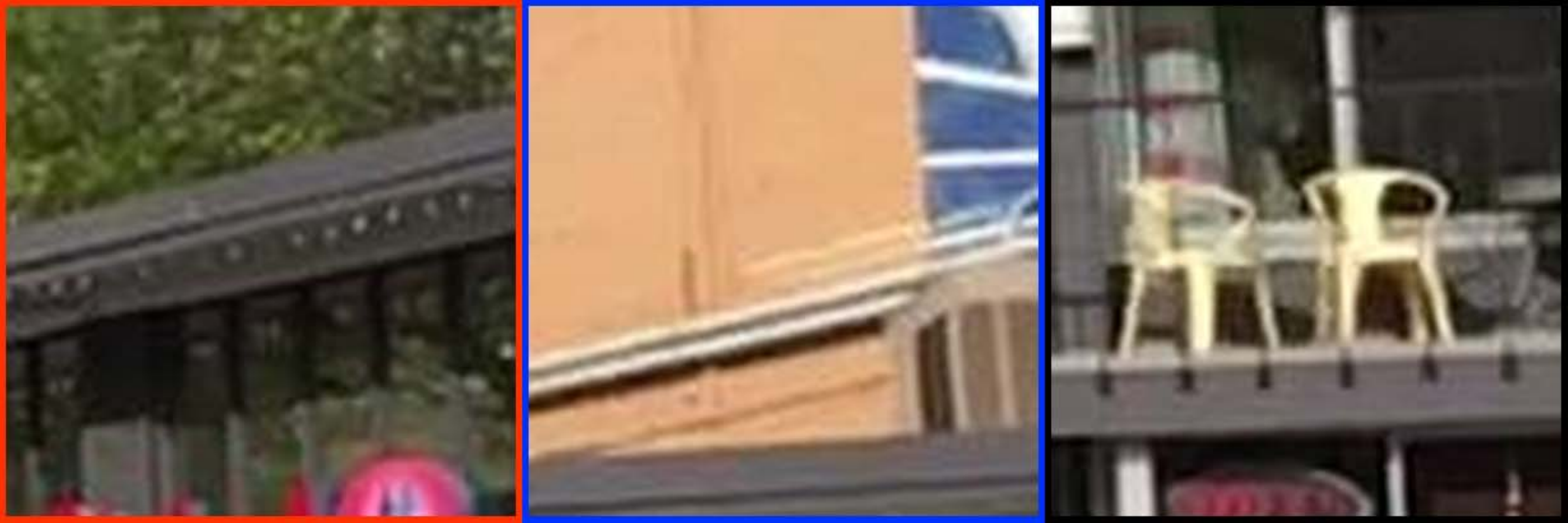} \\
			(e) Zhang \cite{zhang2018dynamic} & (f) Kim  \cite{kim2017online} & (g) ours & (h) clean \\
			24.46/0.8251 & 23.69/0.7782 & 25.67/0.8503 & $+\infty$/1 \\
		\end{tabular}
	\end{center}
	\vspace{-3mm}
	\caption{
		Qualitative evaluations on the dynamic scene deblurring dataset~\cite{su2017Deep}.
		The proposed method generates much clearer images with higher PSNR and SSIM values.
	}
	\label{fig:QuantExp}
\vspace{-3mm}
\end{figure*}

\vspace{-2mm}
\subsection{Loss Function}
\vspace{-1mm}

The loss function depends on both the estimated sharp image as well as the optical flow.
As the ground truth optical flow is not available in our training set, we constrain the optical flow in an unsupervised manner which warps the second blurry input into the first one based on the estimated optical flow according to~\cite{dosovitskiy2015flownet}.
In addition, we use the mean squared error (MSE) loss to measure the differences between the restored sharp image and the clean ground truth image.
The loss function for the proposed network can be written as follows:

\begin{equation}
\label{eq:loss}
\mathcal{L}=\sum_{n=1}^{N}(\sum_{s=1}^{S}\|\mathcal{W}(B^{(2)}_{n,s},\hat{F}_{n,s})-B^{(1)}_{n,s}\|^2)+\|\hat{I}^{(1)}_n-I^{(1)}_n\|^2,
\vspace{-1mm}
\end{equation}
where $N$ is the number of training samples in each batch; $S$ is the number of scales;
$B^{(1)}_{n,s}$ and $B^{(2)}_{n,s}$ denote the first blurry image and the second blurry image at the $s$-th scale; $\hat{F}_{n,s}$ denotes the estimated optical flow in the n-th image at the $s$-th scale;
$\mathcal{W}$ denotes the function which warps the second blurry image into the first one with given estimated optical flow;
$I^{(1)}_{n}$ and $\hat{I}^{(1)}_{n}$ are the ground truth sharp image and the restored sharp image receptively.

\vspace{-2mm}
\subsection{Network Training}
\vspace{-1mm}

We use the training set from \cite{su2017Deep} to train the proposed network.
In order to augment the training data, we randomly crop, resize, rotate, scale and color-permute images.
Every batch has 20 paired patches and the size of every patch is 256.
The proposed network is implemented by MatCaffe~\cite{jia2014caffe}.
We initialize the flow estimation network by the pretrained weights of FlowNets~\cite{dosovitskiy2015flownet} and the remaining layers use Xavier initialization.
The scale $S$ is set to be 5 in the optical flow estimation.
We use Adam method to optimize the network. The learning rate, momentum, momentum2, and weight decay are
$10^{-3}$, 0.9, 0.999 and $10^{-6}$, respectively.
The proposed network converges after 800,000 iterations.
%

\renewcommand{\tabcolsep}{1pt}
\begin{figure*}[!t]\footnotesize
	\centering
	\begin{tabular}{cccc}
		\includegraphics[width=\swfour]{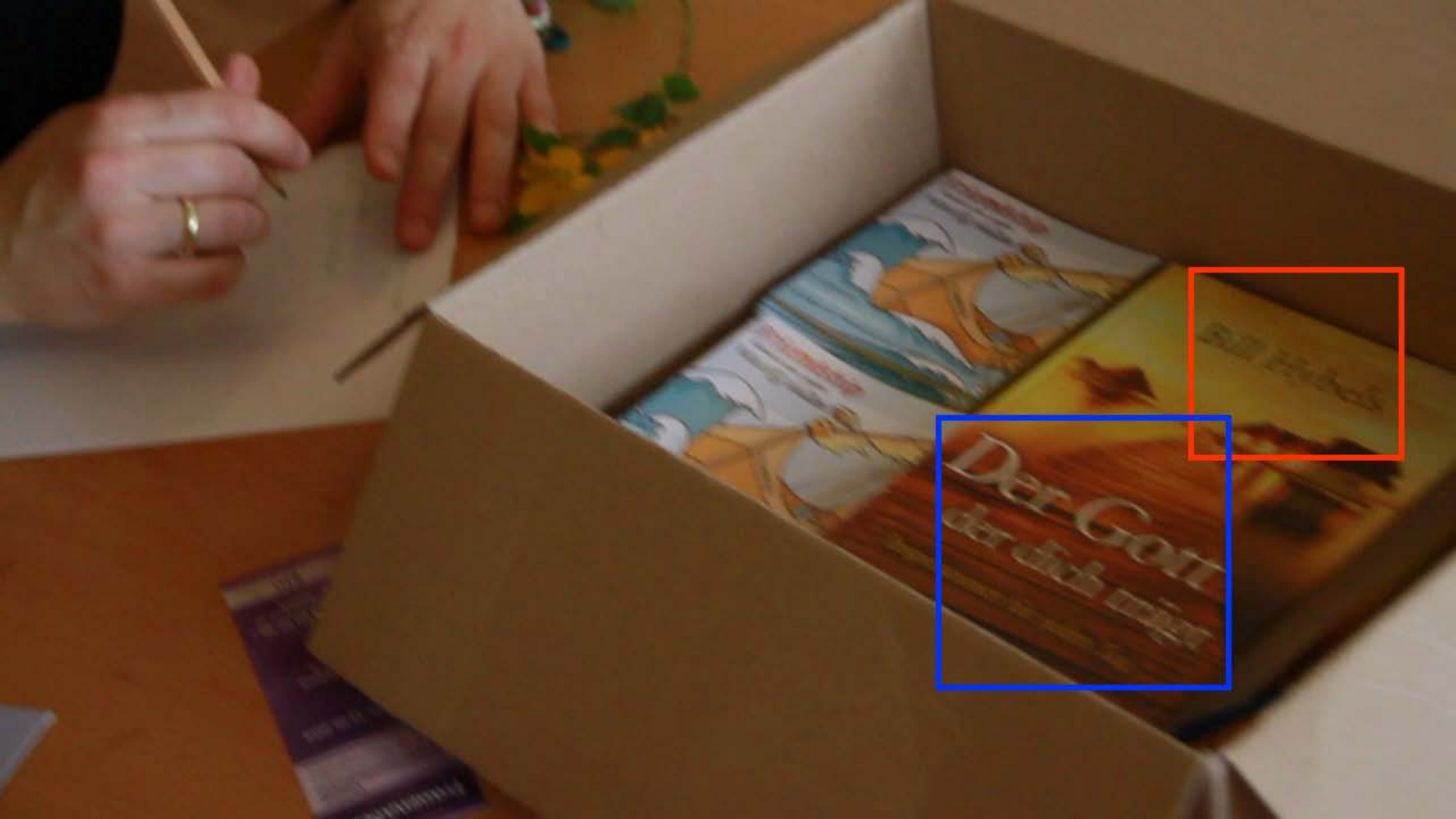} &
		\includegraphics[width=\swfour]{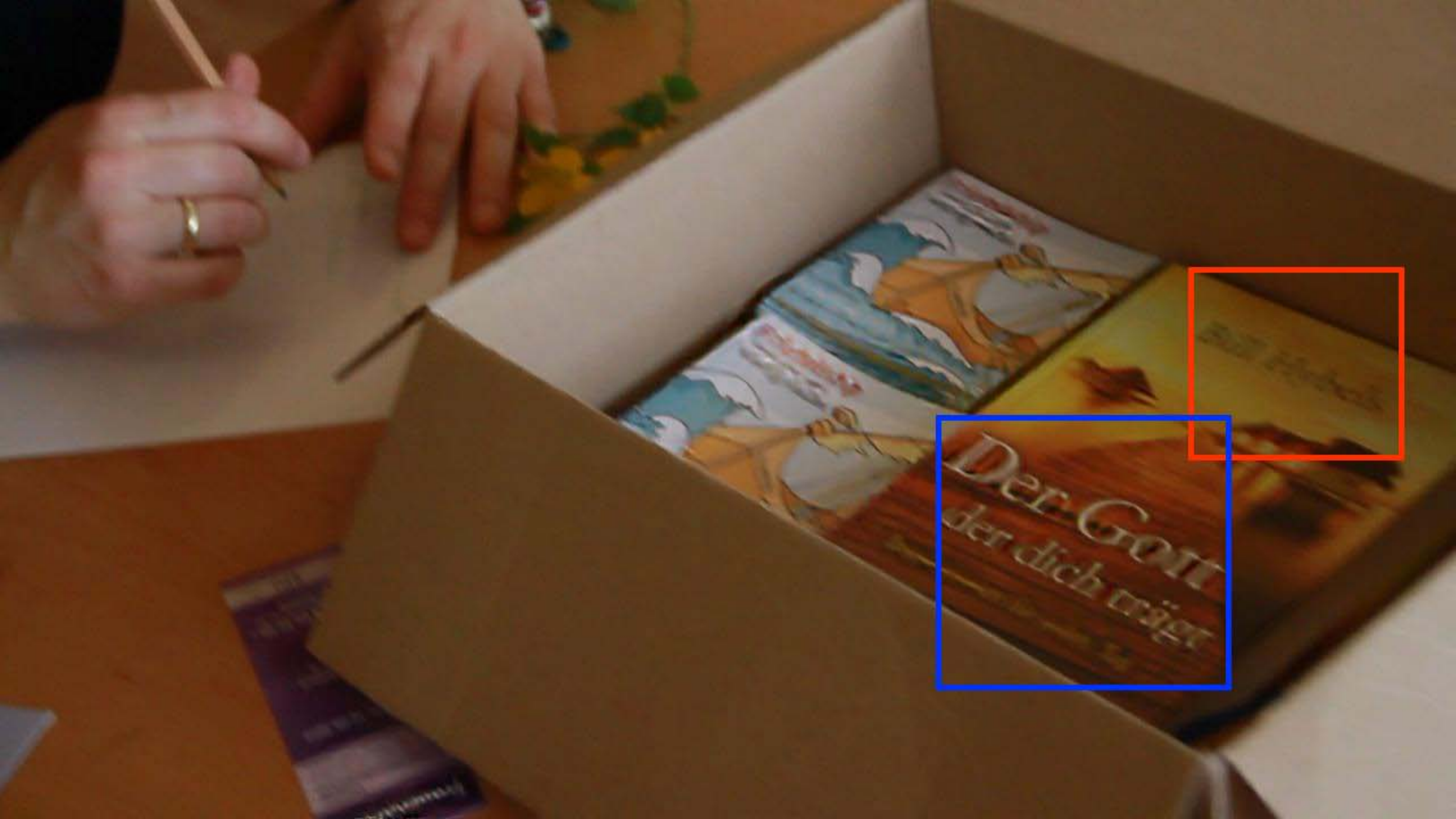} &
		\includegraphics[width=\swfour]{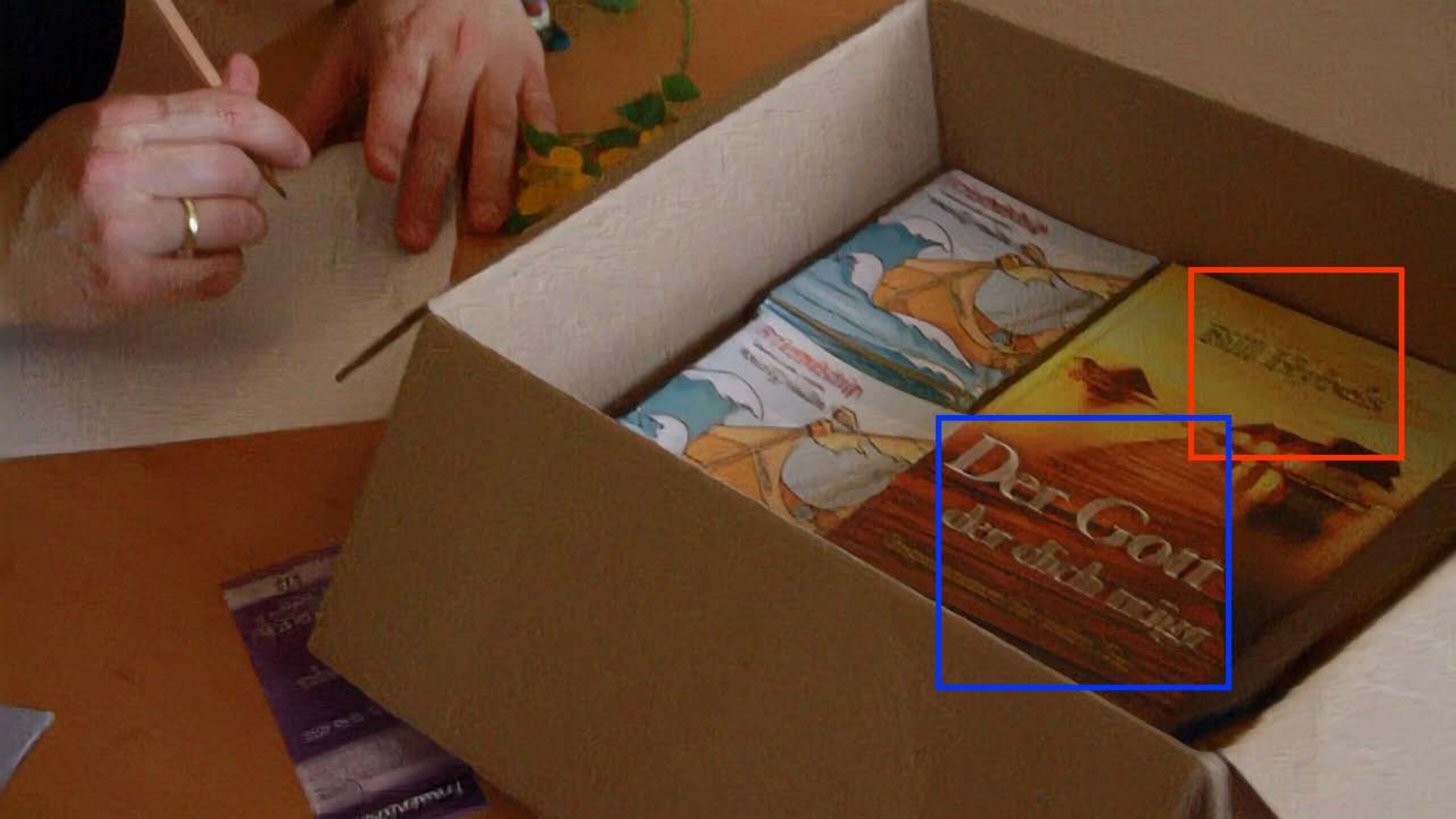} &
		\includegraphics[width=\swfour]{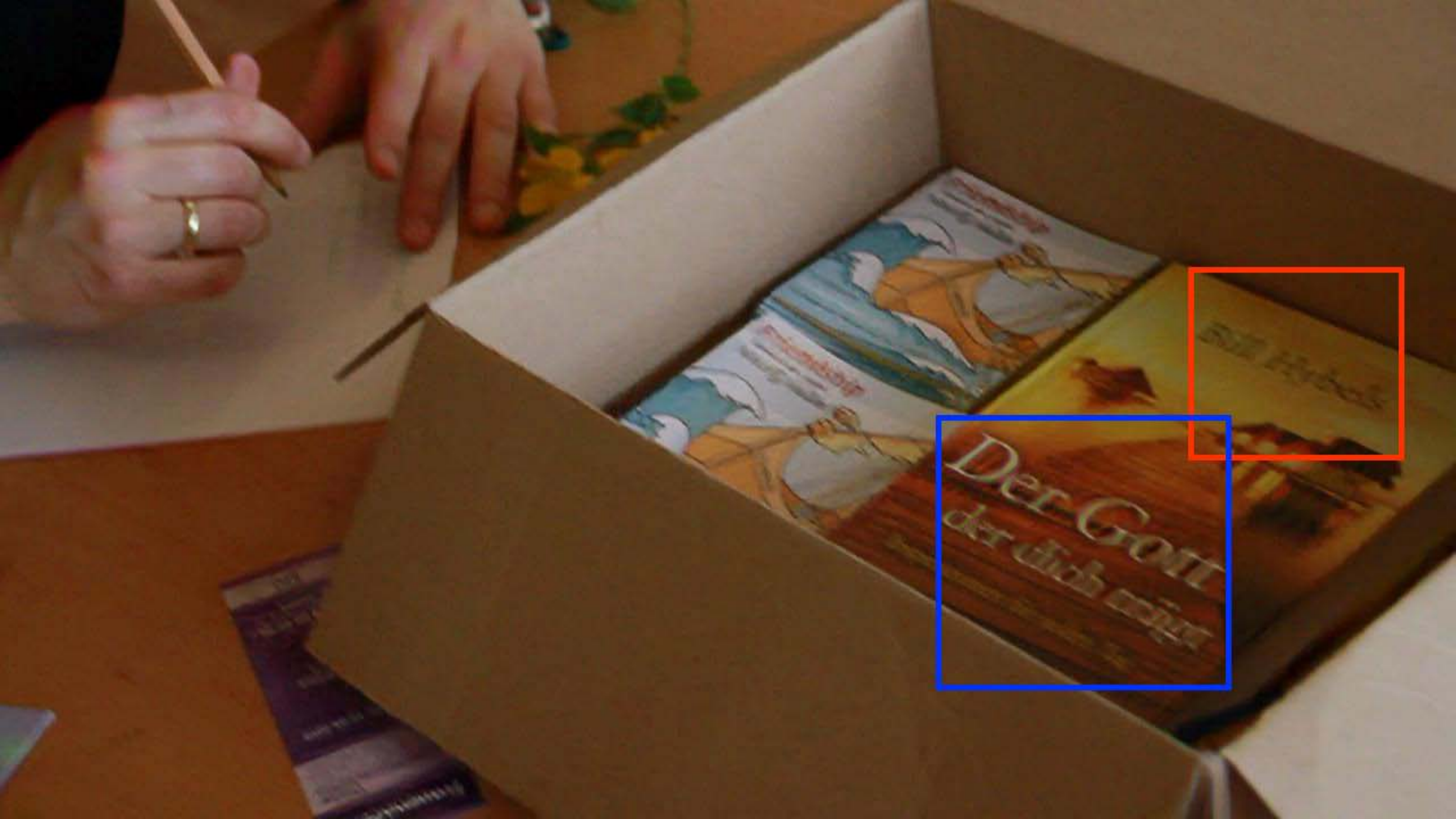} \\
		\includegraphics[width=\swfour]{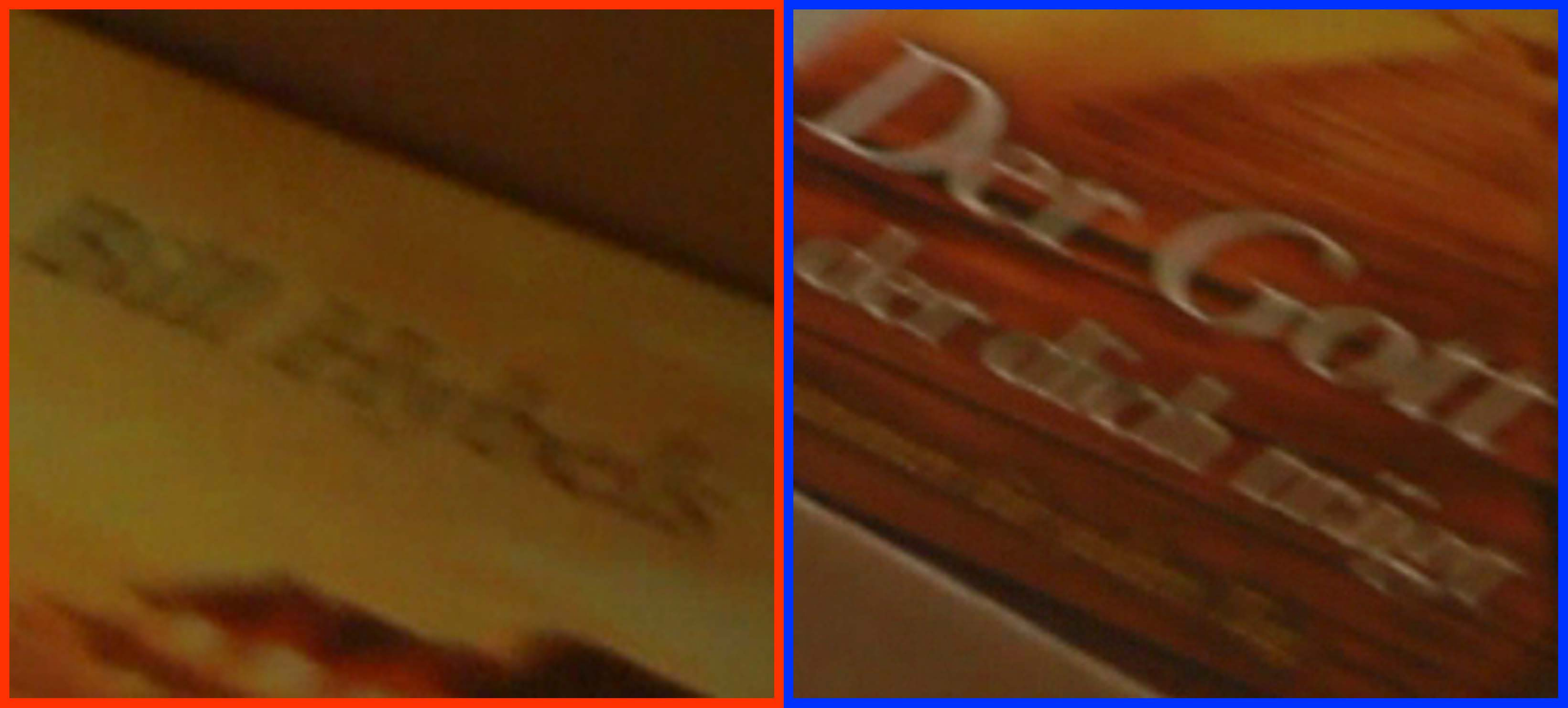} &
		\includegraphics[width=\swfour]{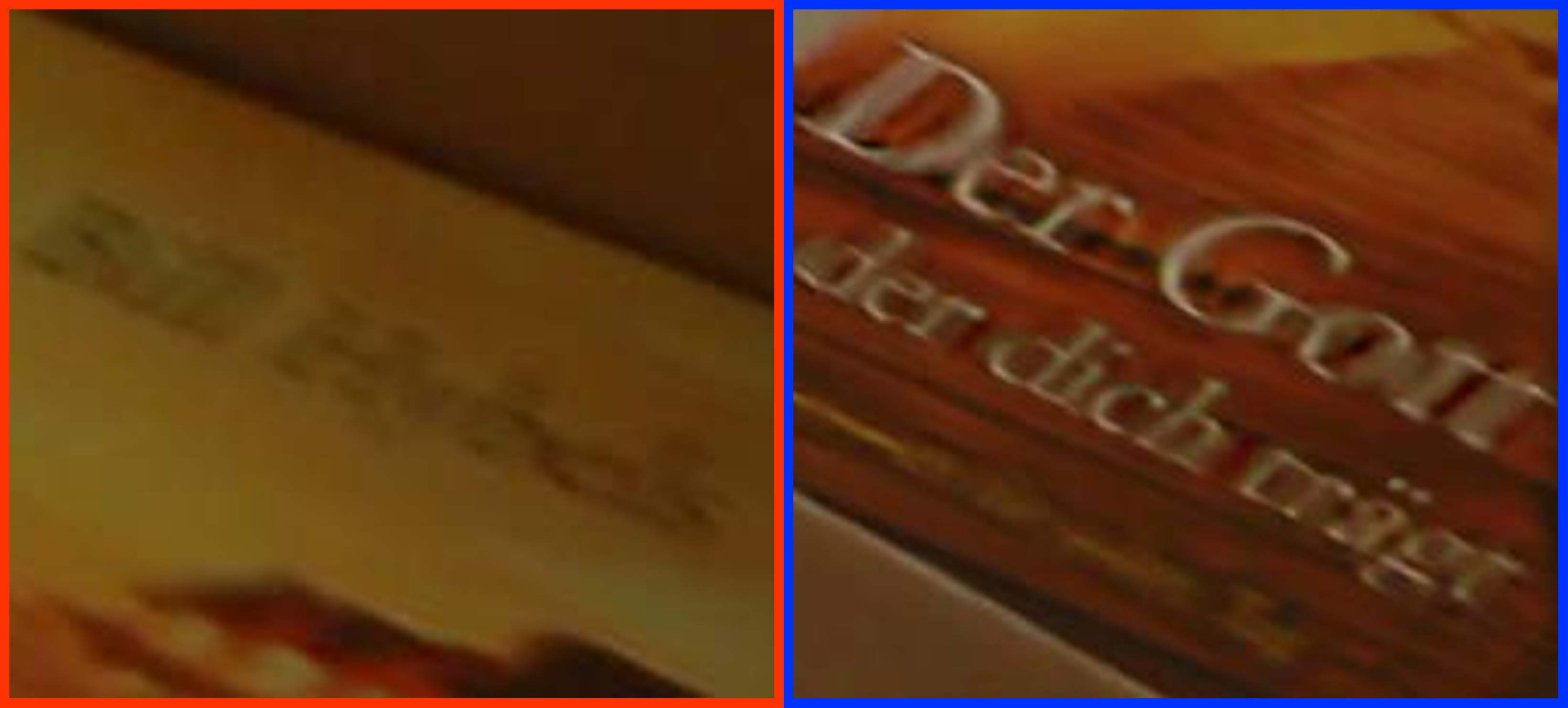} &
		\includegraphics[width=\swfour]{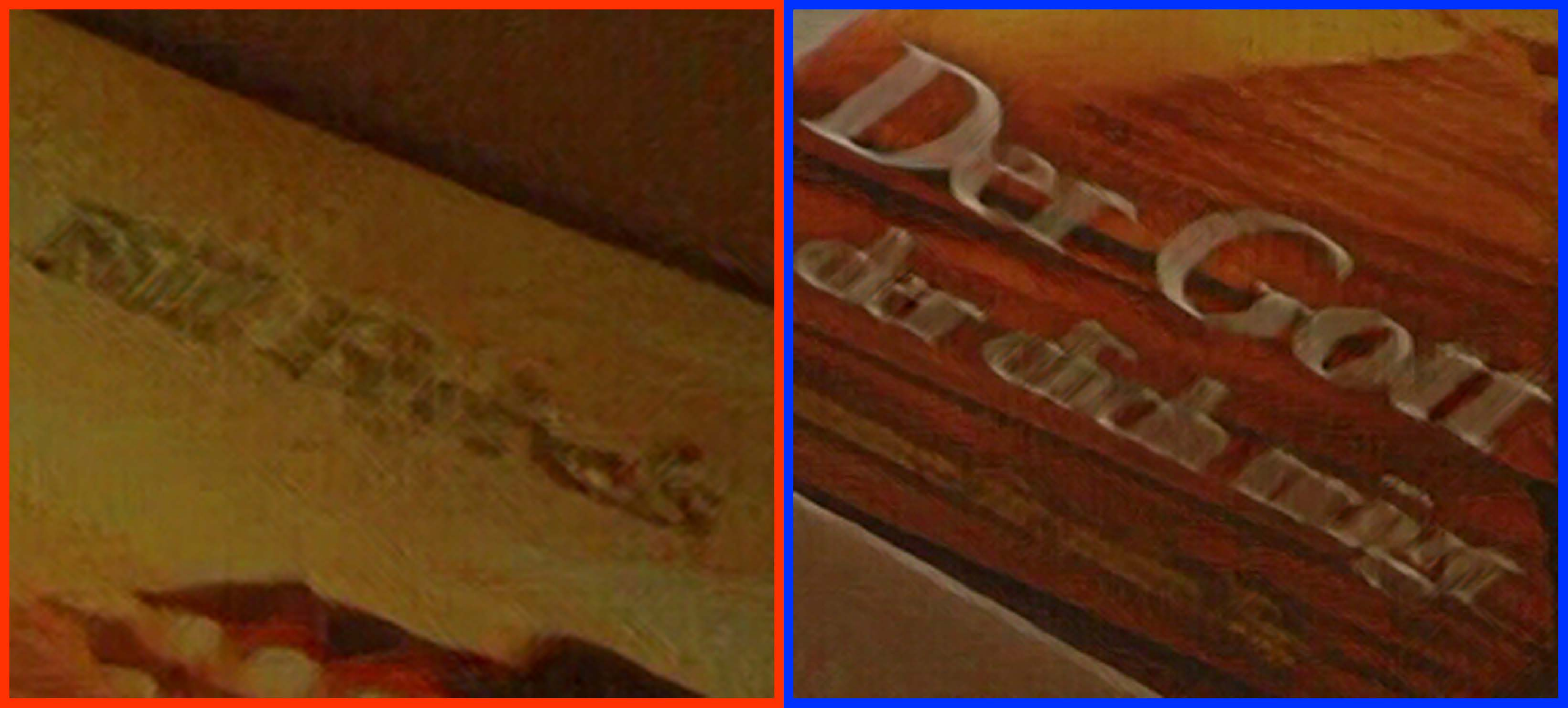} &
		\includegraphics[width=\swfour]{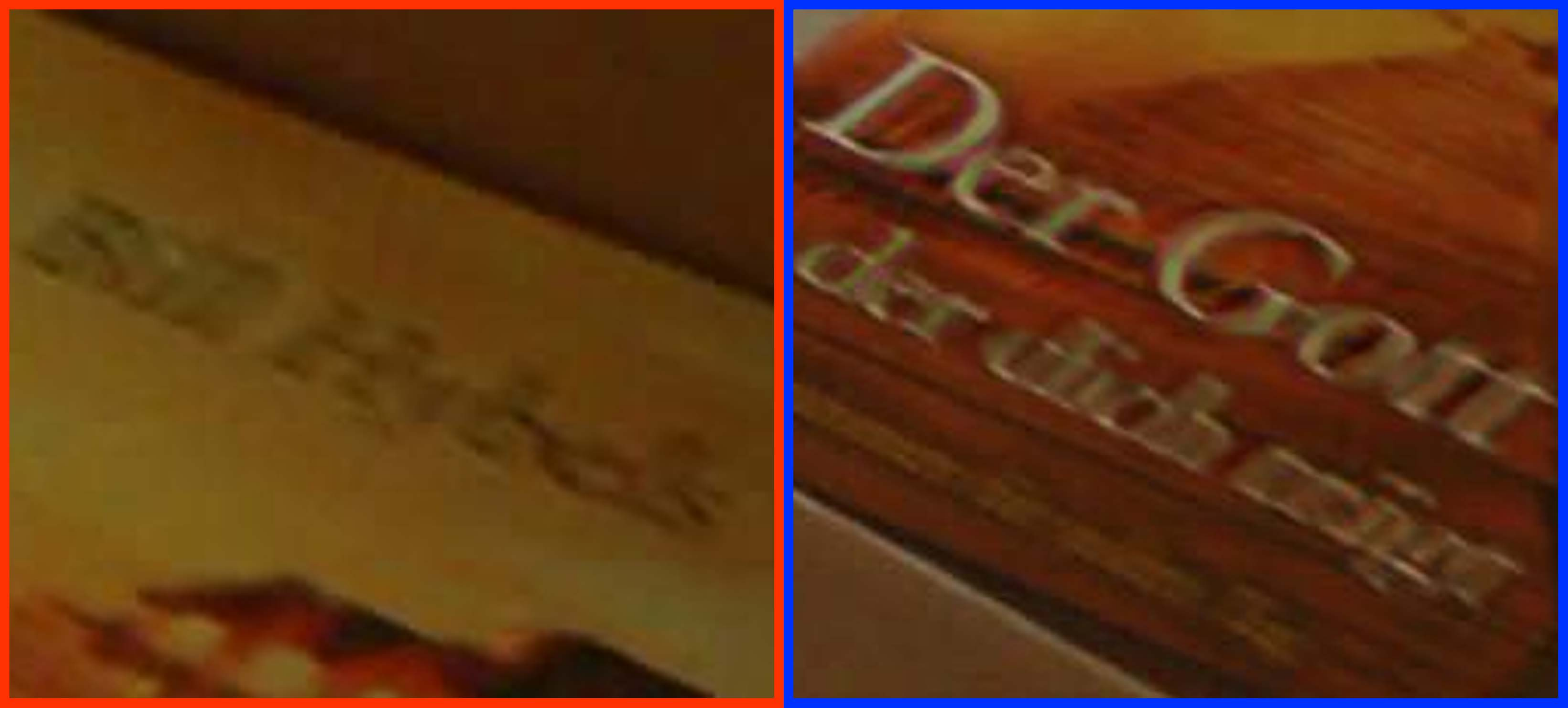} \\
		(a) blurry image & (b) Gong \cite{gong2017motion} & (c) Kupyn \cite{kupyn2017deblurgan} & (d) Tao \cite{tao2018scale} \\
		\includegraphics[width=\swfour]{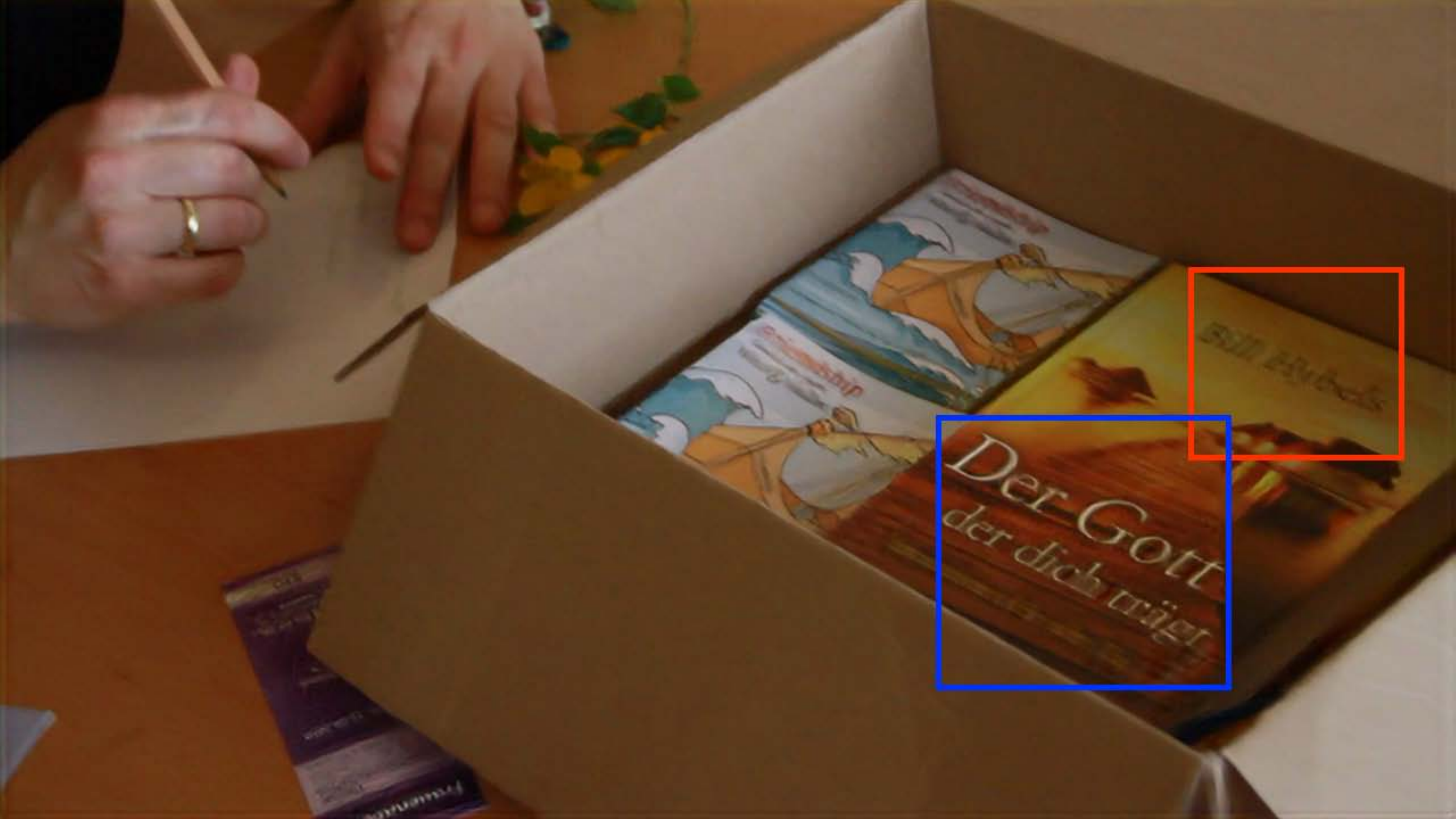} &
		\includegraphics[width=\swfour]{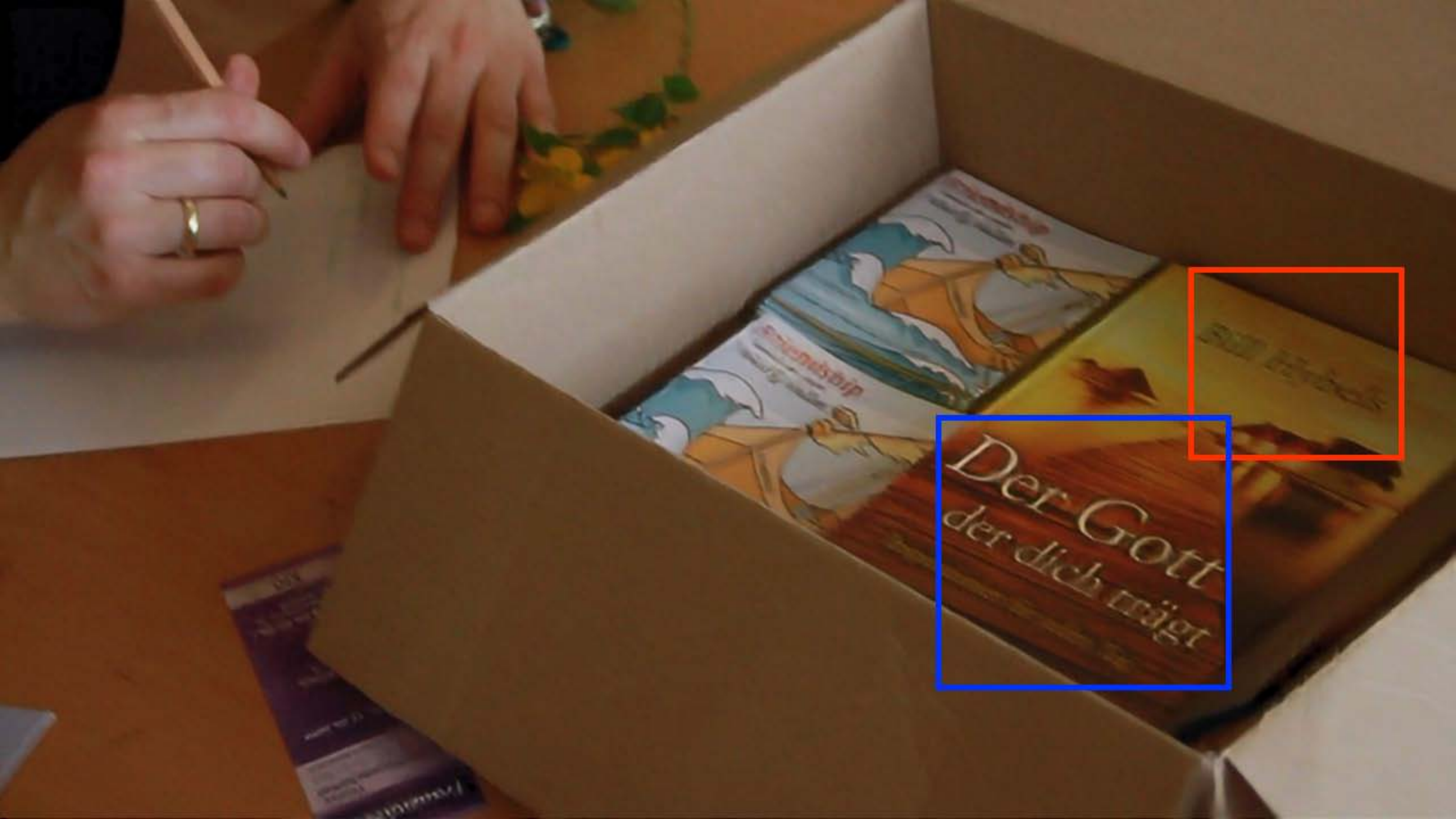} &
		\includegraphics[width=\swfour]{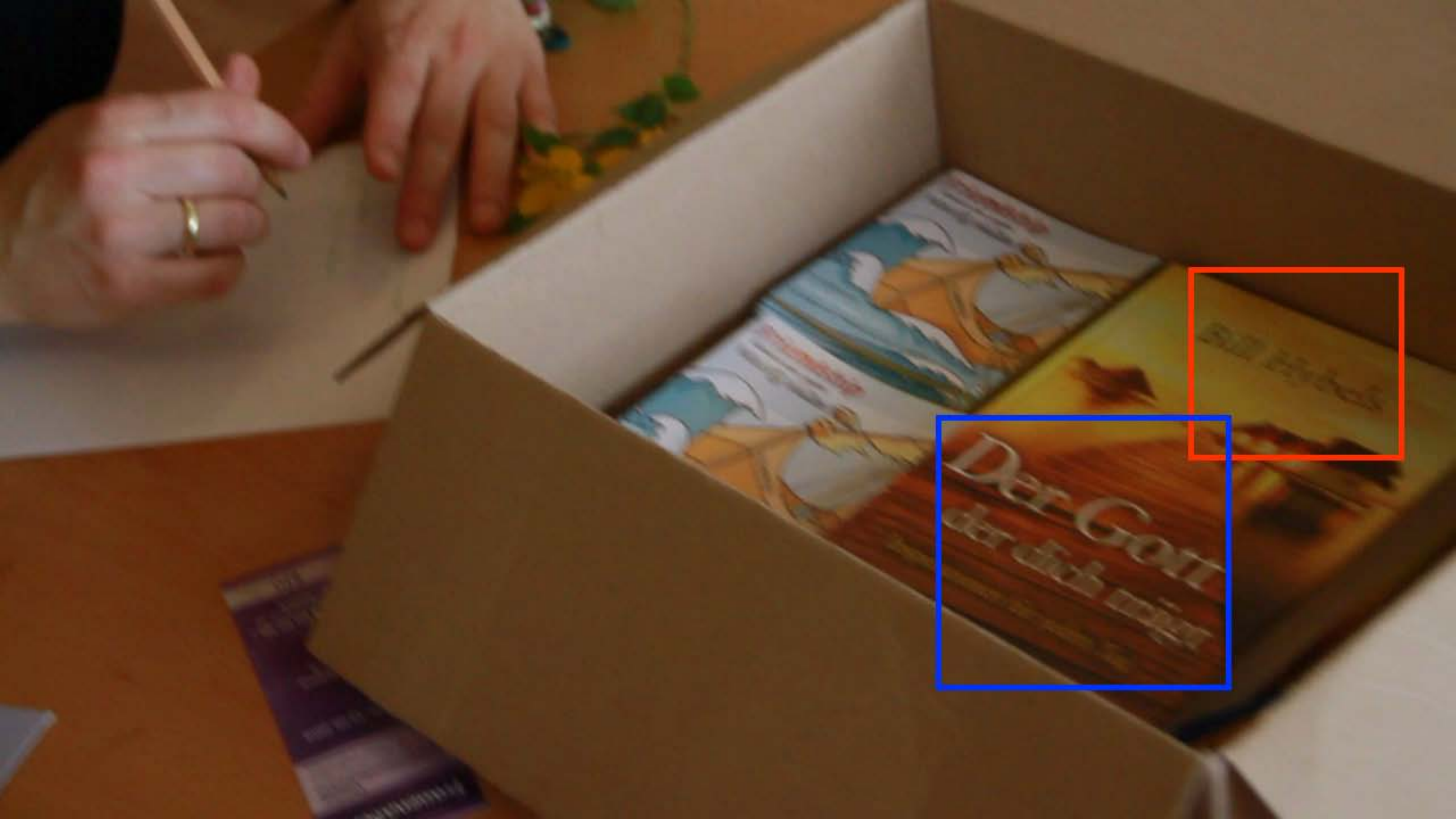} &
		\includegraphics[width=\swfour]{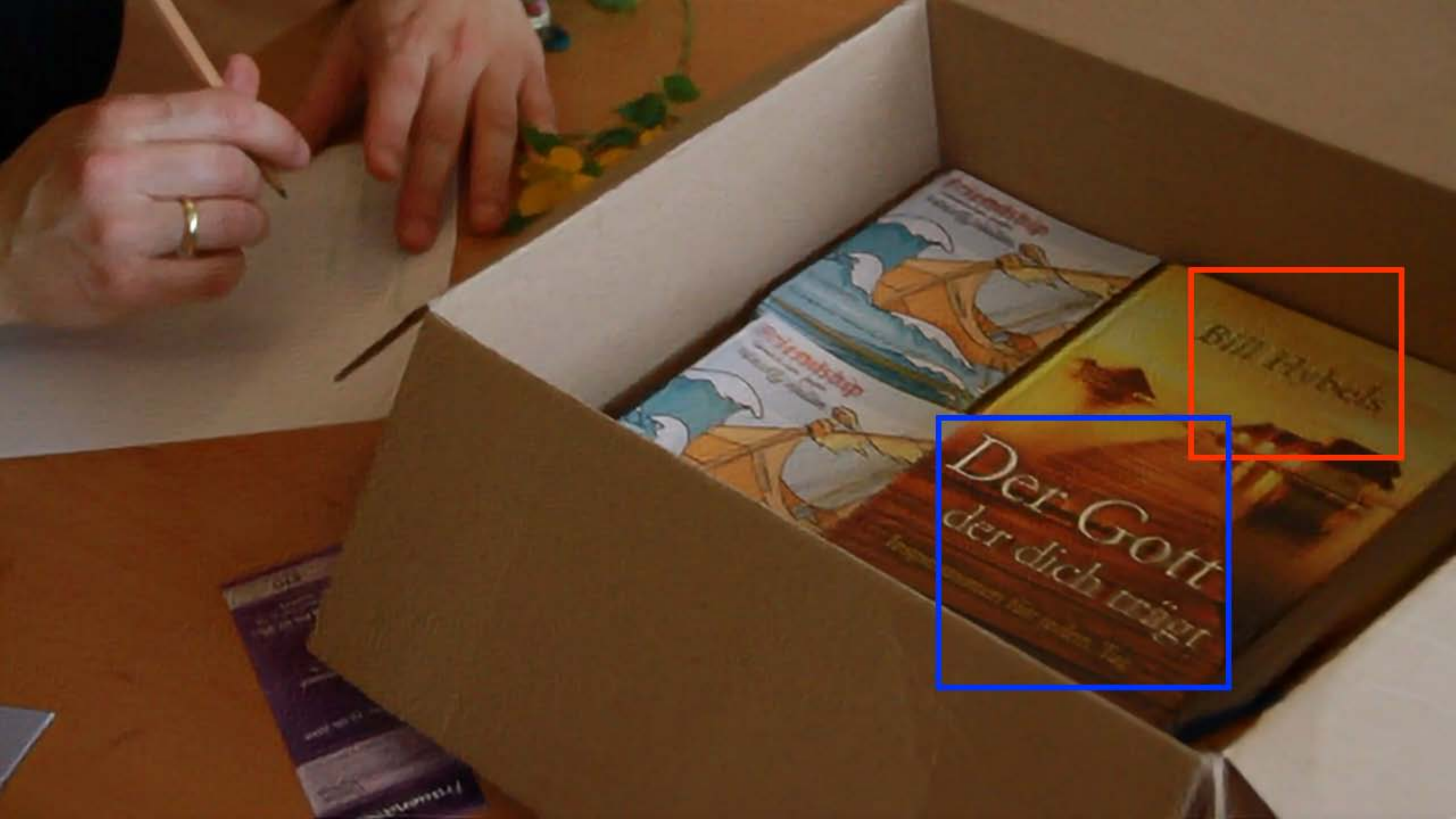} \\
		\includegraphics[width=\swfour]{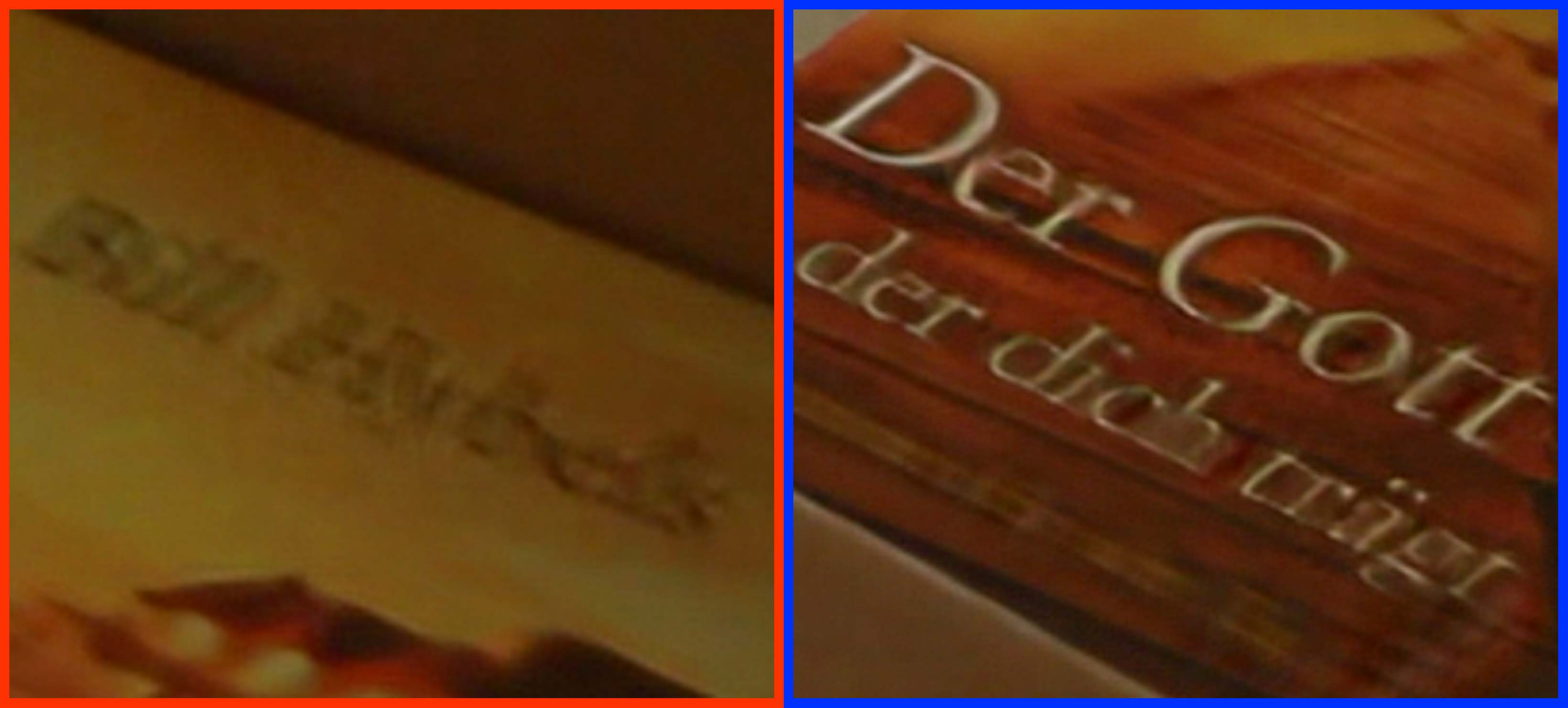} &
		\includegraphics[width=\swfour]{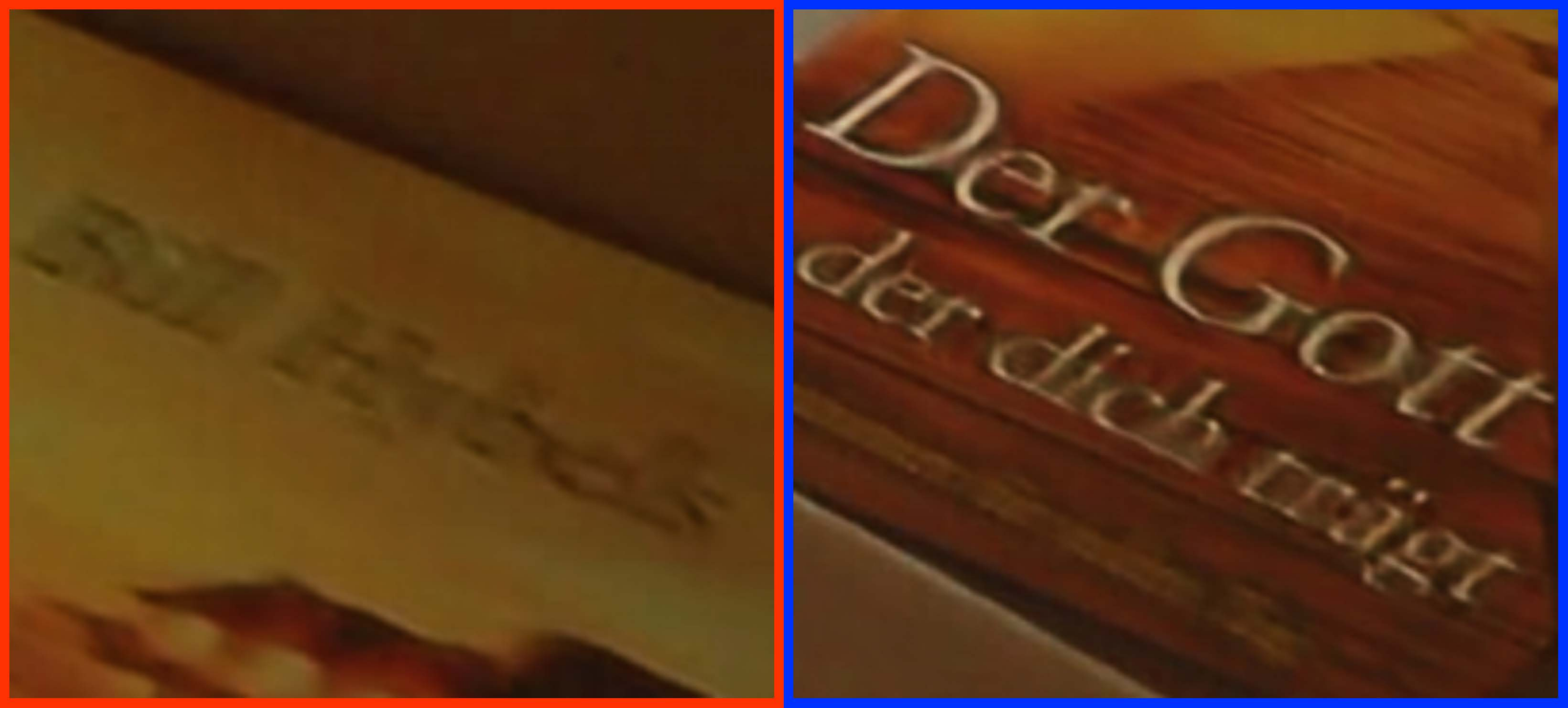} &
		\includegraphics[width=\swfour]{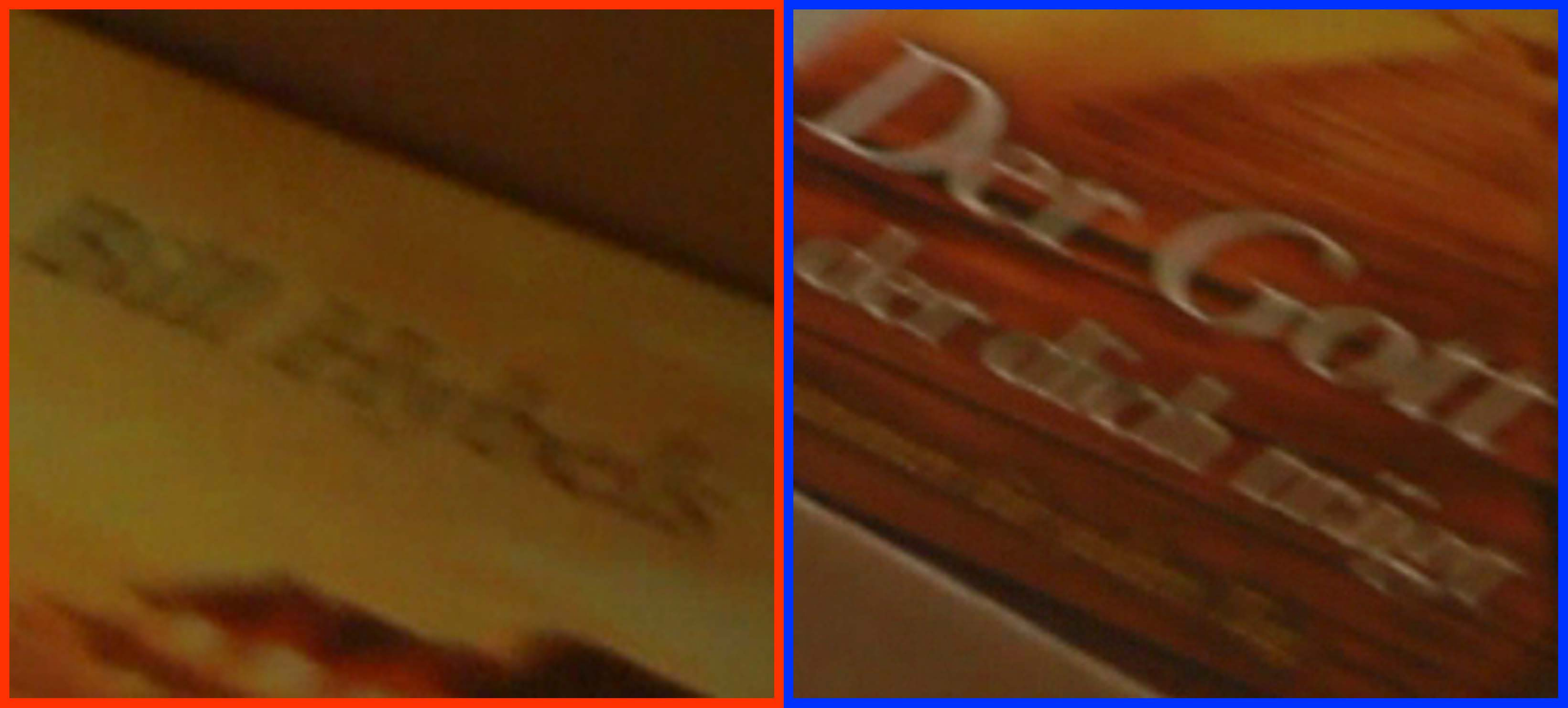} &
		\includegraphics[width=\swfour]{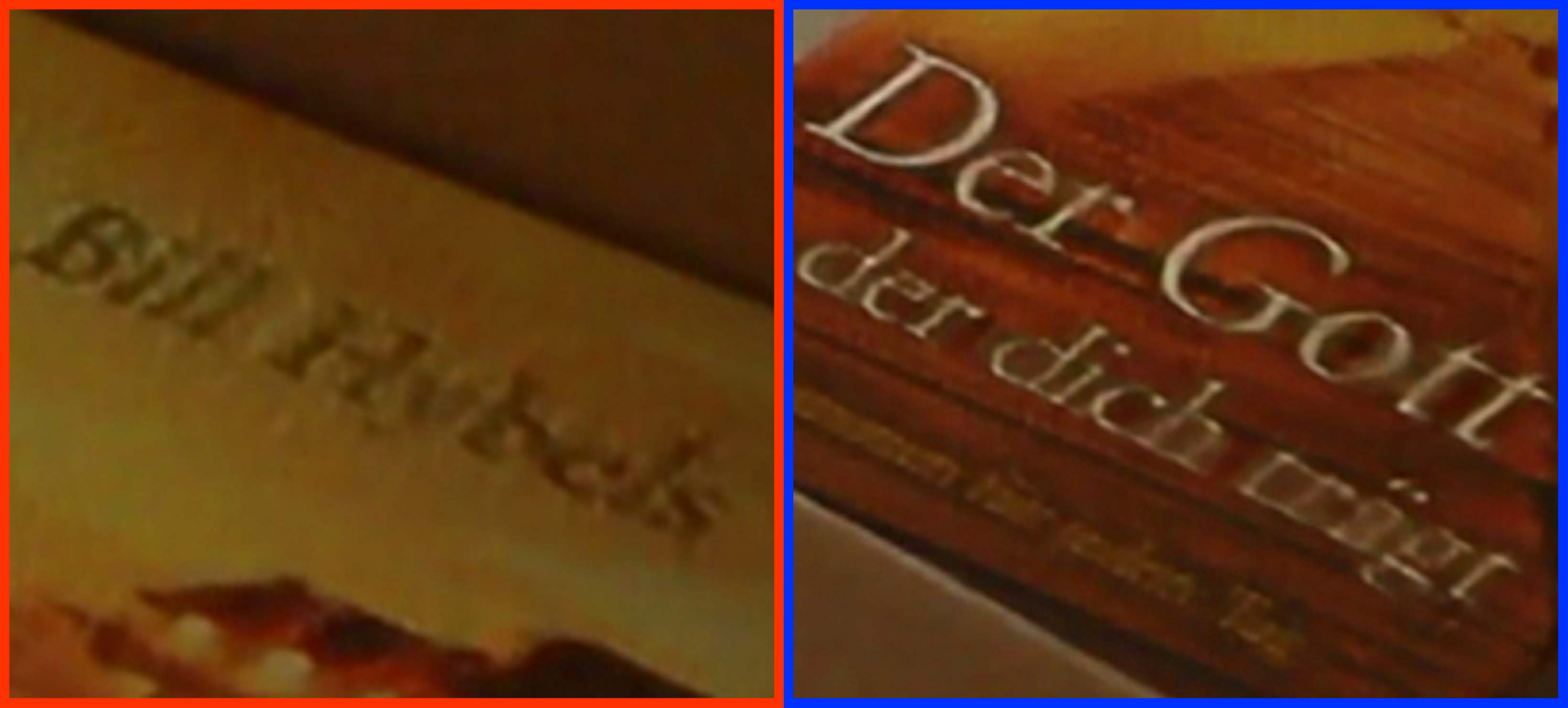} \\
		(e) Zhang \cite{zhang2018dynamic} & (f) Su \cite{su2017Deep} & (g) Kim  \cite{kim2017online} & (h) ours \\
	\end{tabular}
	\vspace{-2mm}
	\caption{
		Qualitative evaluations on the real images~\cite{su2017Deep}.
		The proposed method generates much clearer images.
	}
	\label{fig:QualiExp}
\vspace{-3mm}
\end{figure*}

\renewcommand{\tabcolsep}{3pt}
\begin{table*}[!t]\footnotesize
	\renewcommand{\arraystretch}{1.1}
	\centering
	\caption{Running time and number of parameters for an image with the size of 720$\times$1280 pixels.
		All existing methods use their publicly available scripts. A ``-" indicates that the result is not available.
	}
	\label{table:runtime}
	\vspace{-1mm}
	\begin{tabular}{cccccccccccc}
		\toprule
		& Whyte \cite{whyte2012non} & Sun \cite{sun2015learning} & Gong \cite{gong2017motion} & Nah \cite{nah2017deep} & Kupyn \cite{kupyn2017deblurgan} & Tao \cite{tao2018scale} & Zhang \cite{zhang2018dynamic} & Kim \cite{kim2015generalized} & Su \cite{su2017Deep} & Kim  \cite{kim2017online} & ours \\
		\midrule
		Running time (sec) & 700 & 1500 & 1500 & 15 & 0.68 & 2.7 & 1.4 & 880 & 6.8 & 0.13 & 0.20 \\
		params & \verb|-| & 7.3M & 10.3M & 11.7M & 11.4M & 8.1M & 9.2M & \verb|-| & 16.8M & 0.92M & 8.7M \\
		\bottomrule
	\end{tabular}
\vspace{-2mm}
\end{table*}

\section{Experimental Results}
%
In this section, we both quantitatively and qualitatively evaluate the proposed method on the publicly available benchmark datasets and real-world images and compare it against state-of-the-art dynamic scene deblurring algorithms including non-uniform image deblurring~\cite{whyte2012non} and video deblurring~\cite{kim2015generalized}, deep learning-based spatially variant blur kernel estimation~\cite{sun2015learning, gong2017motion} as well as deep learning based dynamic scene image deblurring~\cite{nah2017deep, kupyn2017deblurgan, tao2018scale, zhang2018dynamic, kupyn2019deblurgan, cai2020dark} and video deblurring~\cite{su2017Deep, kim2017online} in terms of both PSNR and SSIM.
We use the publicly available implementations of the methods mentioned above to restore the sharp images.
Except for \cite{sun2015learning, gong2017motion,kim2017online, kupyn2019deblurgan, cai2020dark} which do not provide the training scripts, we train the rest deep learning based methods on the training set provided by \cite{su2017Deep}.
Our source code and the datasets used in this paper will be made available to the public.

\vspace{-2mm}
\subsection{Quantitative Evaluations}
\vspace{-1mm}
We quantitatively evaluate our algorithm on the video deblurring dataset~\cite{su2017Deep} which is a typical dynamic scene blur dataset.
Table~\ref{table:exp_dvd} shows that the proposed algorithm
performs favorably against the start-of-the-art methods in terms of PSNR and SSIM.

Figure~\ref{fig:QuantExp} shows some examples from the test set.
The visual comparisons demonstrate that the existing methods do not completely remove blur.
In contrast, the proposed algorithm generates a much clearer image due to the use of optical flow.

\vspace{-2mm}
\subsection{Qualitative Evaluations}
\vspace{-1mm}

We also qualitatively compare our method with other algorithms on the real blurry images from \cite{su2017Deep}.
Figure~\ref{fig:QualiExp} shows that the proposed network generates the image with clear characters while other algorithms generate blurry images.

\renewcommand{\tabcolsep}{1pt}
\begin{figure*}[t]\footnotesize
	\begin{center}
		\begin{tabular}{cccc}
			\includegraphics[width=\swfour]{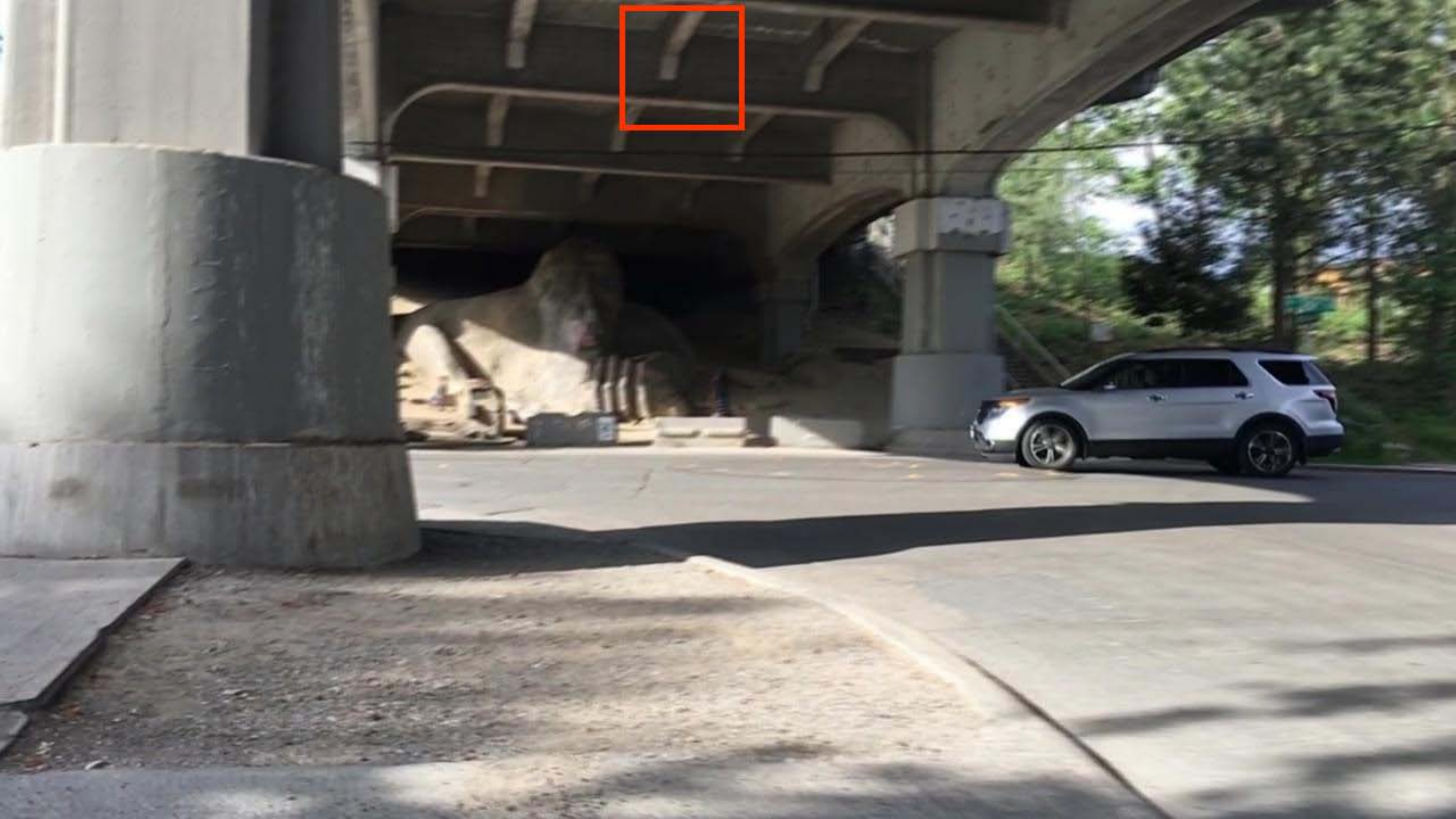} &
			\includegraphics[width=\swfour]{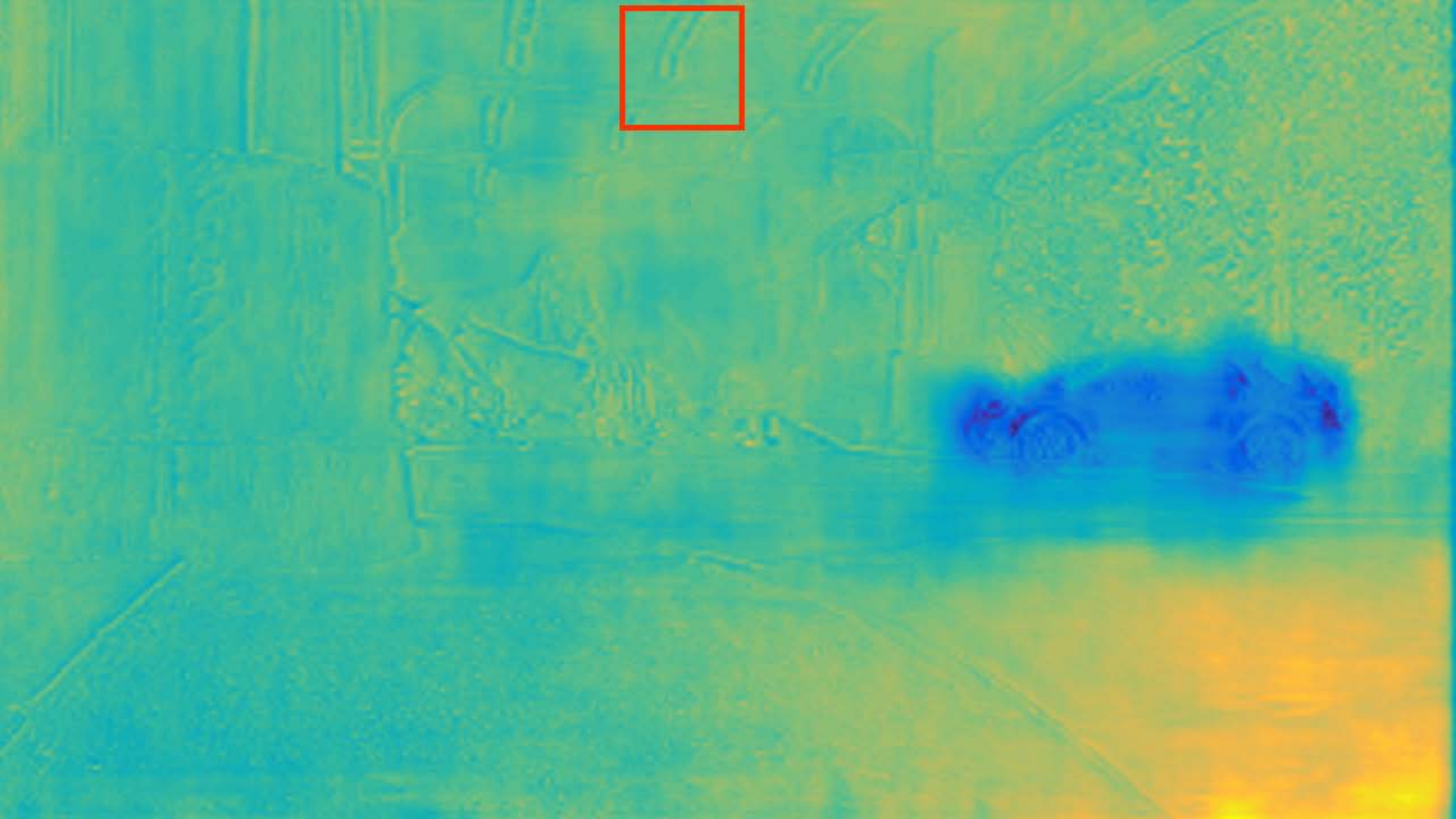} &
			\includegraphics[width=\swfour]{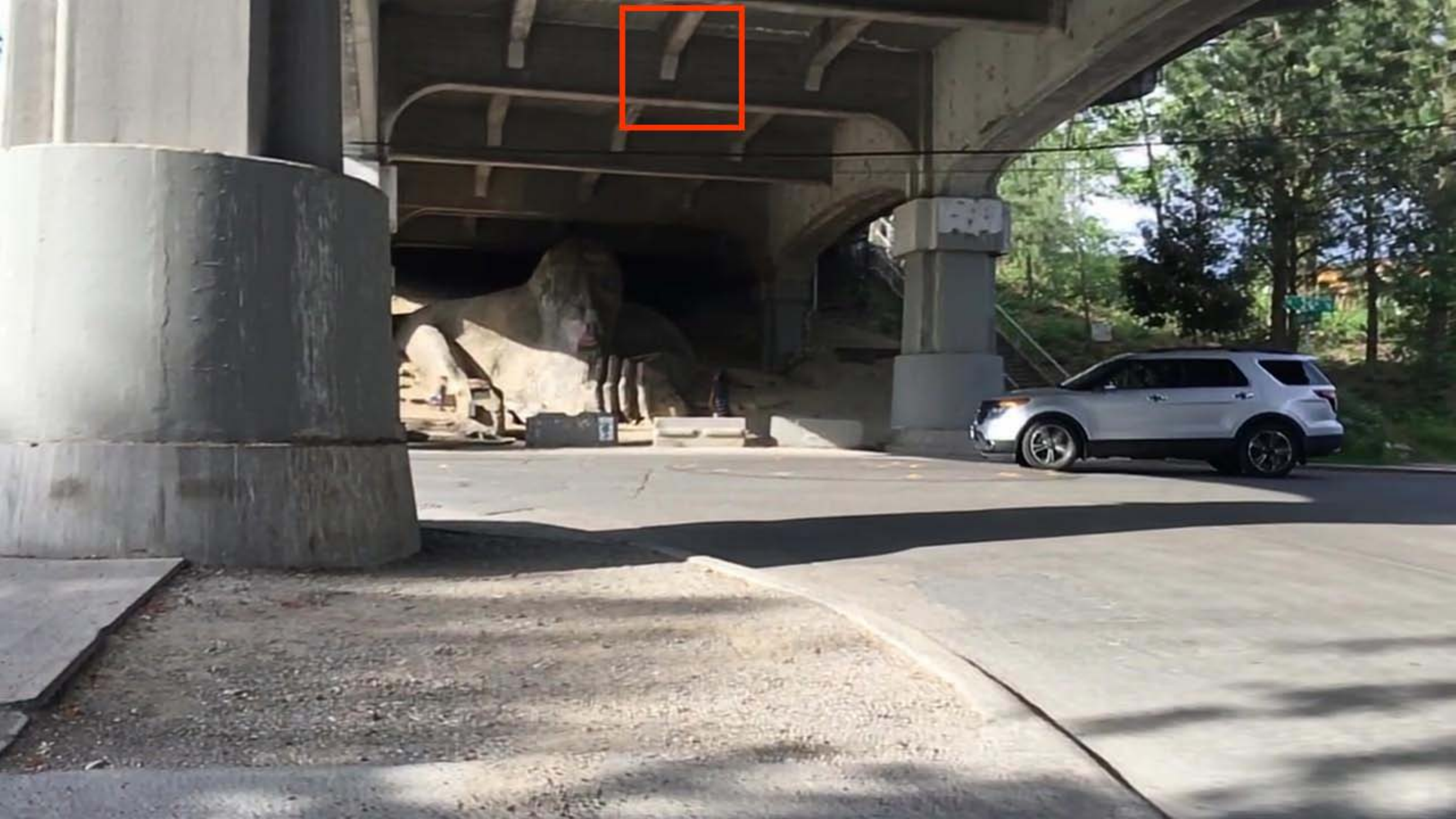} &
			\includegraphics[width=\swfour]{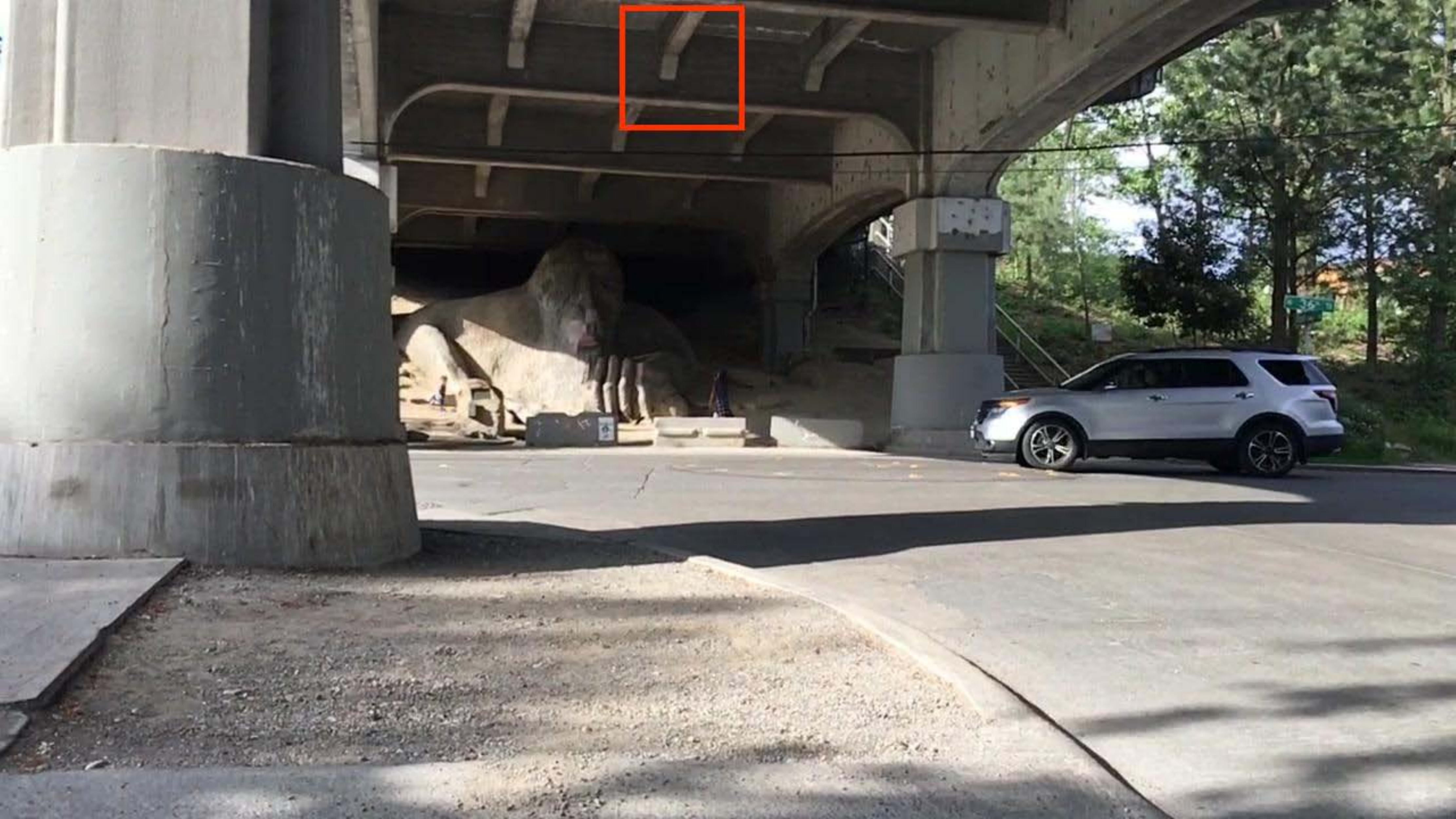} \\
			(a) blurry image & (b) horizontal flow & (c) ours & (d) clean image \\
		\end{tabular}
		\begin{tabular}{cccccc}
			\includegraphics[width=\swsix]{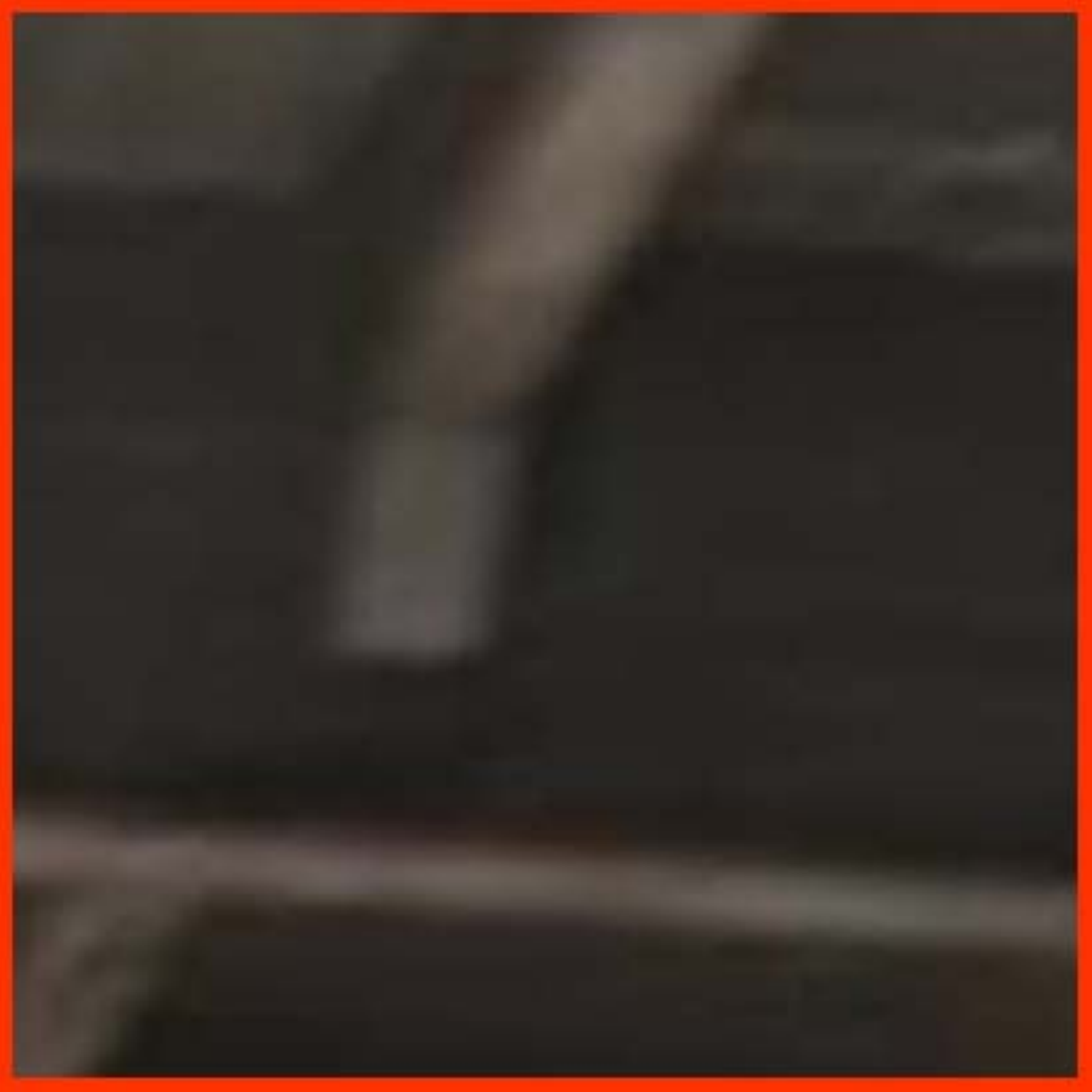} &
			\includegraphics[width=\swsix]{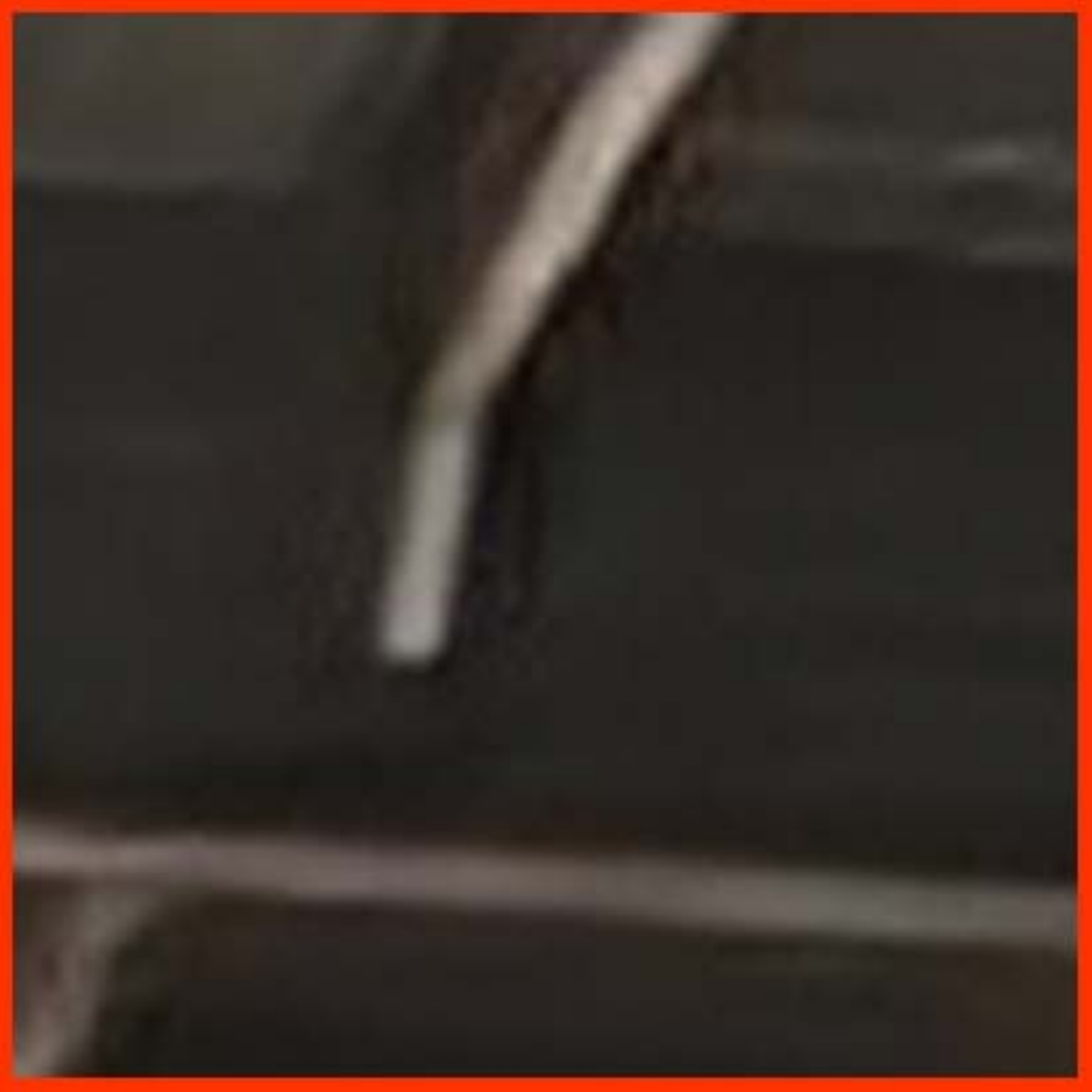} &
			\includegraphics[width=\swsix]{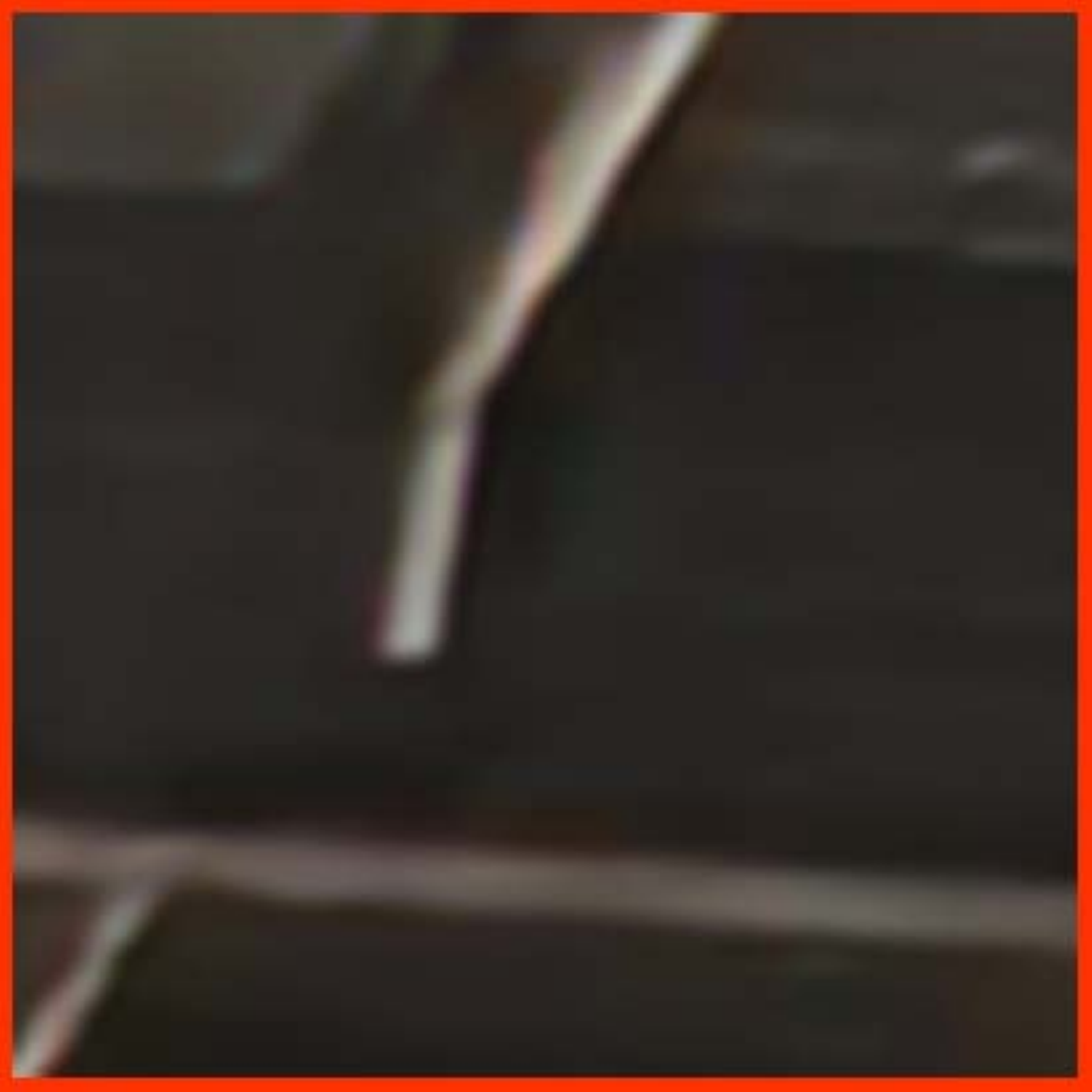} &
			\includegraphics[width=\swsix]{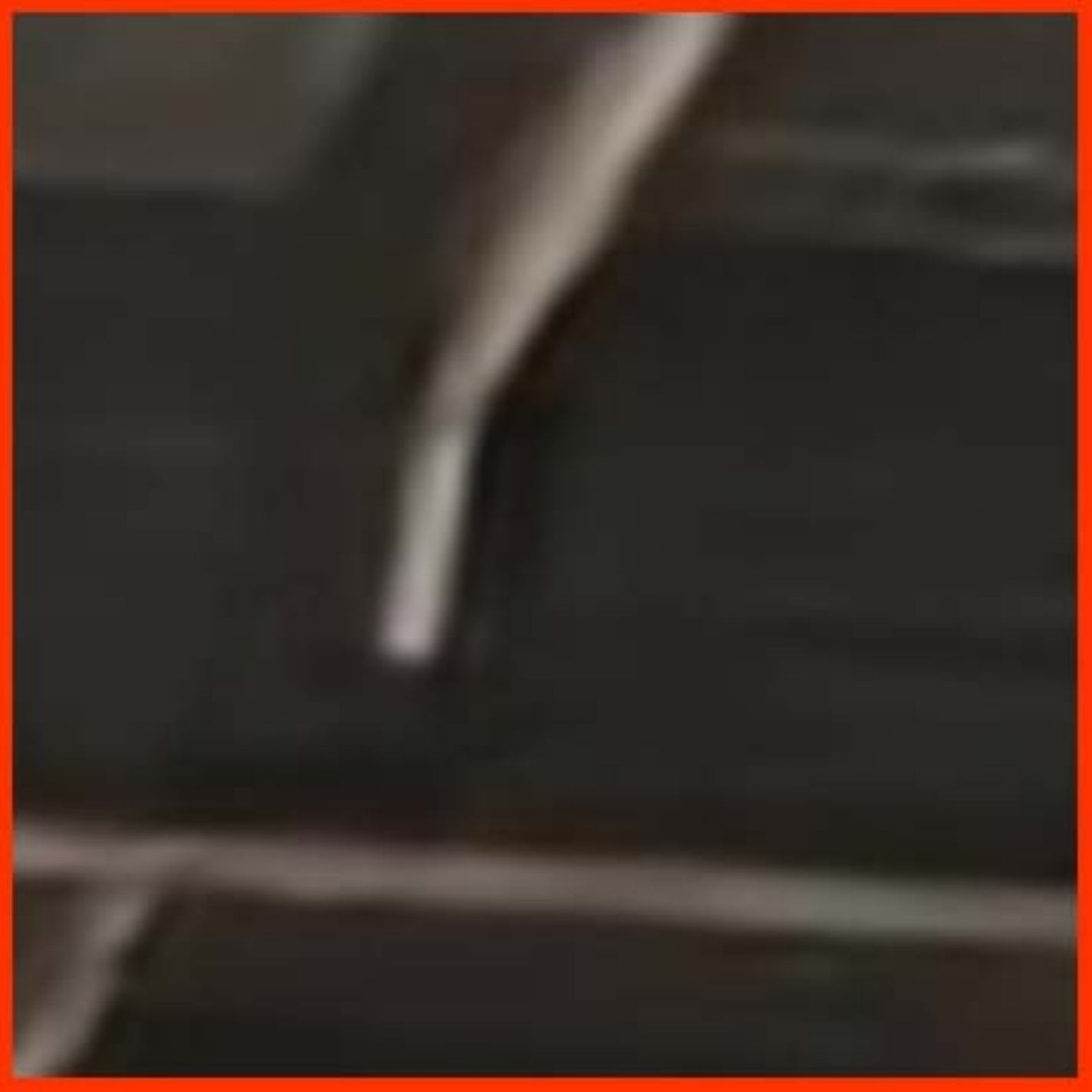} &
			\includegraphics[width=\swsix]{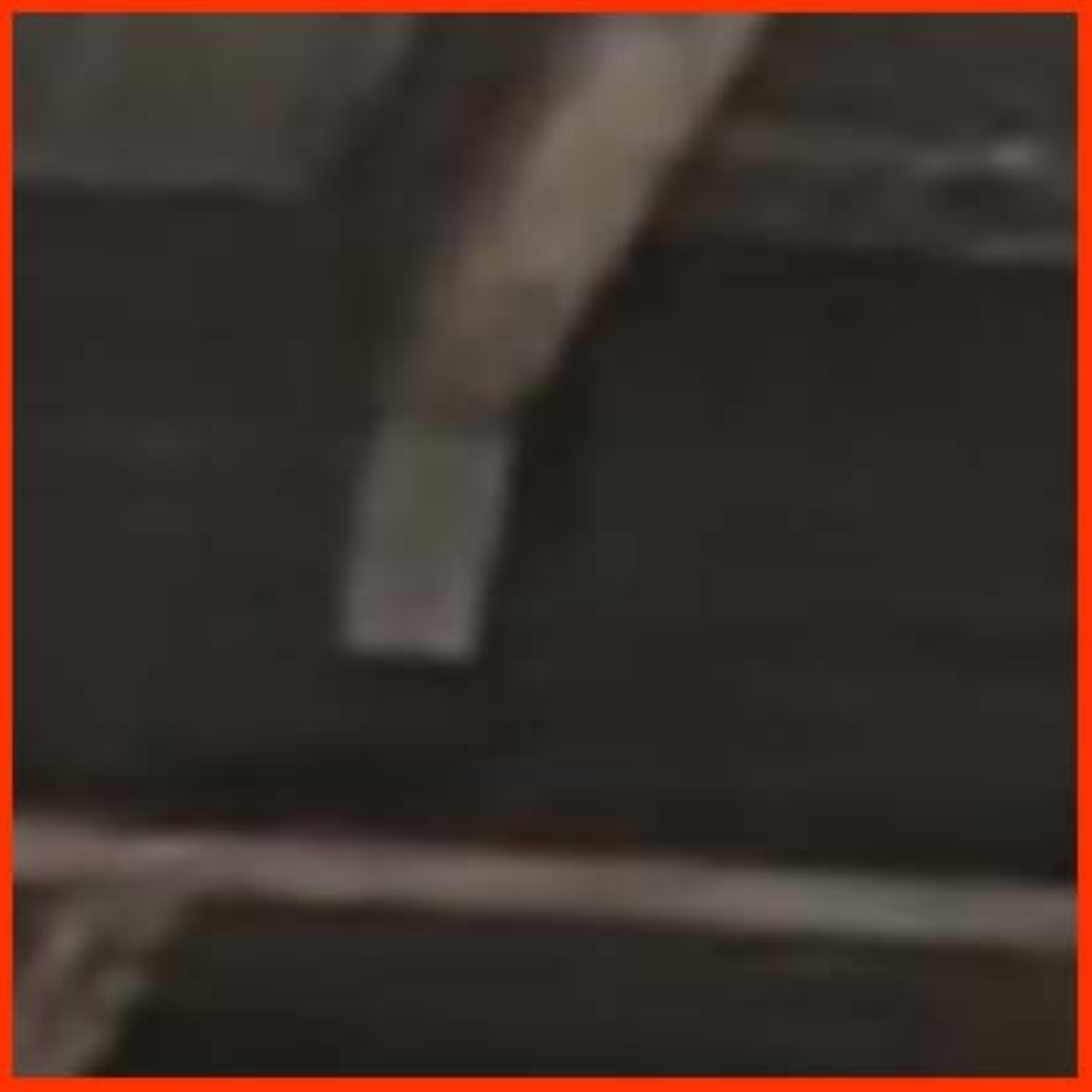} &
			\includegraphics[width=\swsix]{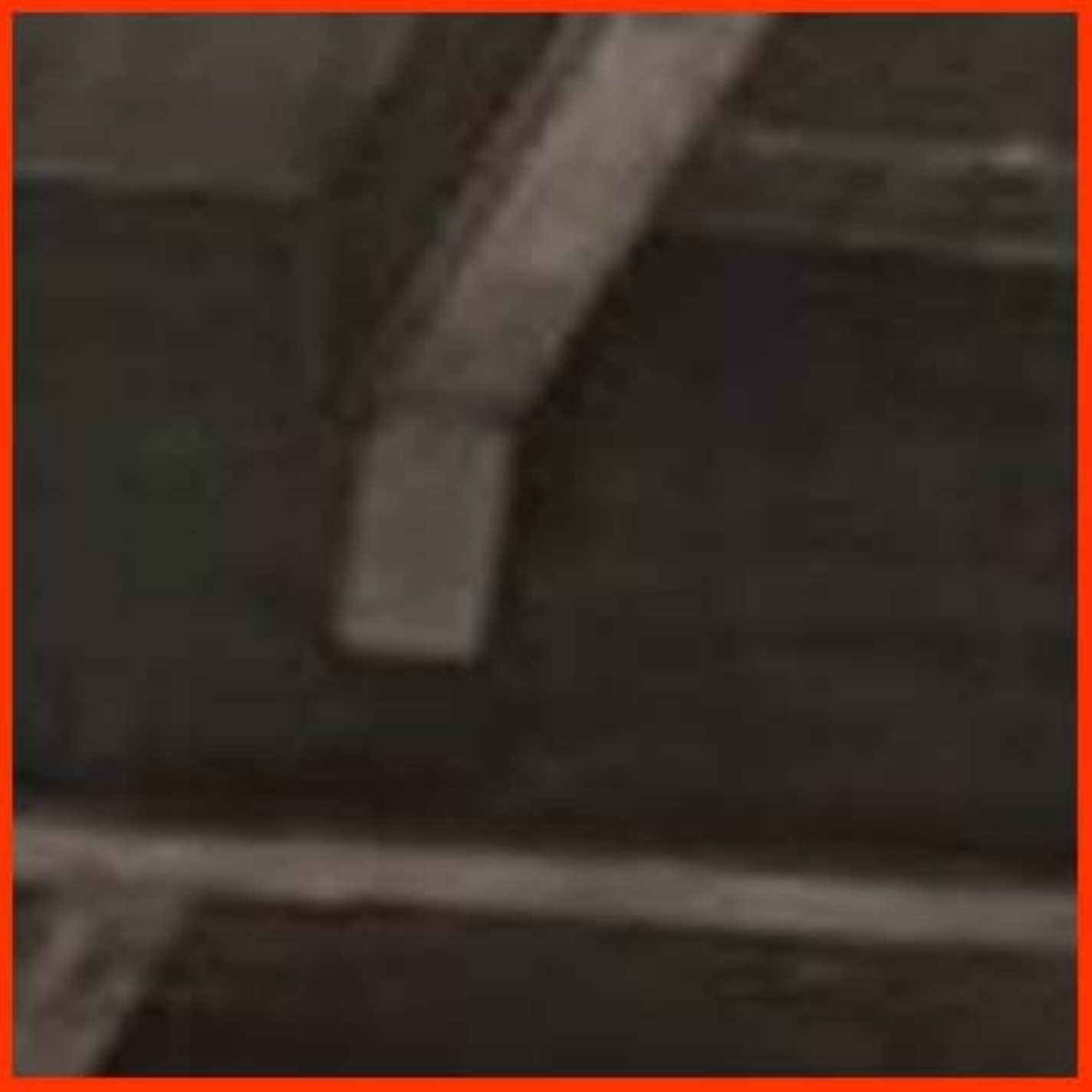} \\
			(e) blurry image & (f) Nah \cite{nah2017deep} & (g) Tao \cite{tao2018scale} & (h) Zhang \cite{zhang2018dynamic} & (i) ours & (j) clean image \\
		\end{tabular}
	\end{center}
	\vspace{-3mm}
	\caption{
		Effectiveness of the optical flow.
		(a) to (d) are the blurry image, our estimated horizontal optical flow, the restored image from our network and the sharp image.
		(e) and (j) are cropped patches from the blurry image and the sharp image.
		(f) to (i) are cropped patches from restored image of \cite{nah2017deep}, \cite{tao2018scale}, \cite{zhang2018dynamic} and our network respectively.
		\cite{nah2017deep}, \cite{tao2018scale} and \cite{zhang2018dynamic} treat the support beam of the bridge as a blurry region and generate artifact.
		According to (b), there is not strong motion in the support beam.
		With the information from the estimated optical flow, our network will not consider this region as a blurry region and generate a correctly-restored image.
	}
	\label{fig:flow1}
\vspace{-2mm}
\end{figure*}

\renewcommand{\tabcolsep}{1pt}
\begin{figure*}[t]\footnotesize
	\begin{center}
		\begin{tabular}{cccc}
			\includegraphics[width=\swfour]{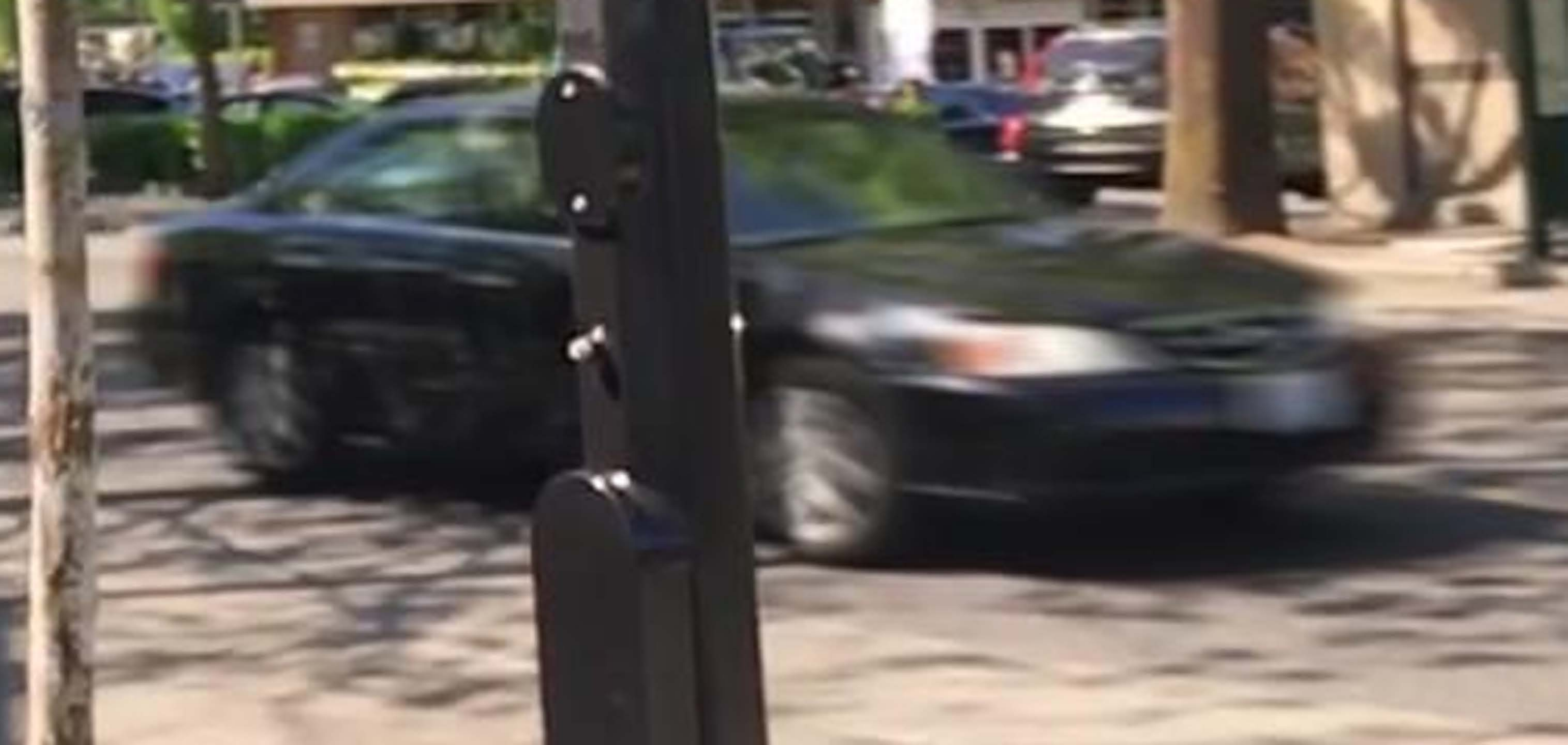} &
			\includegraphics[width=\swfour]{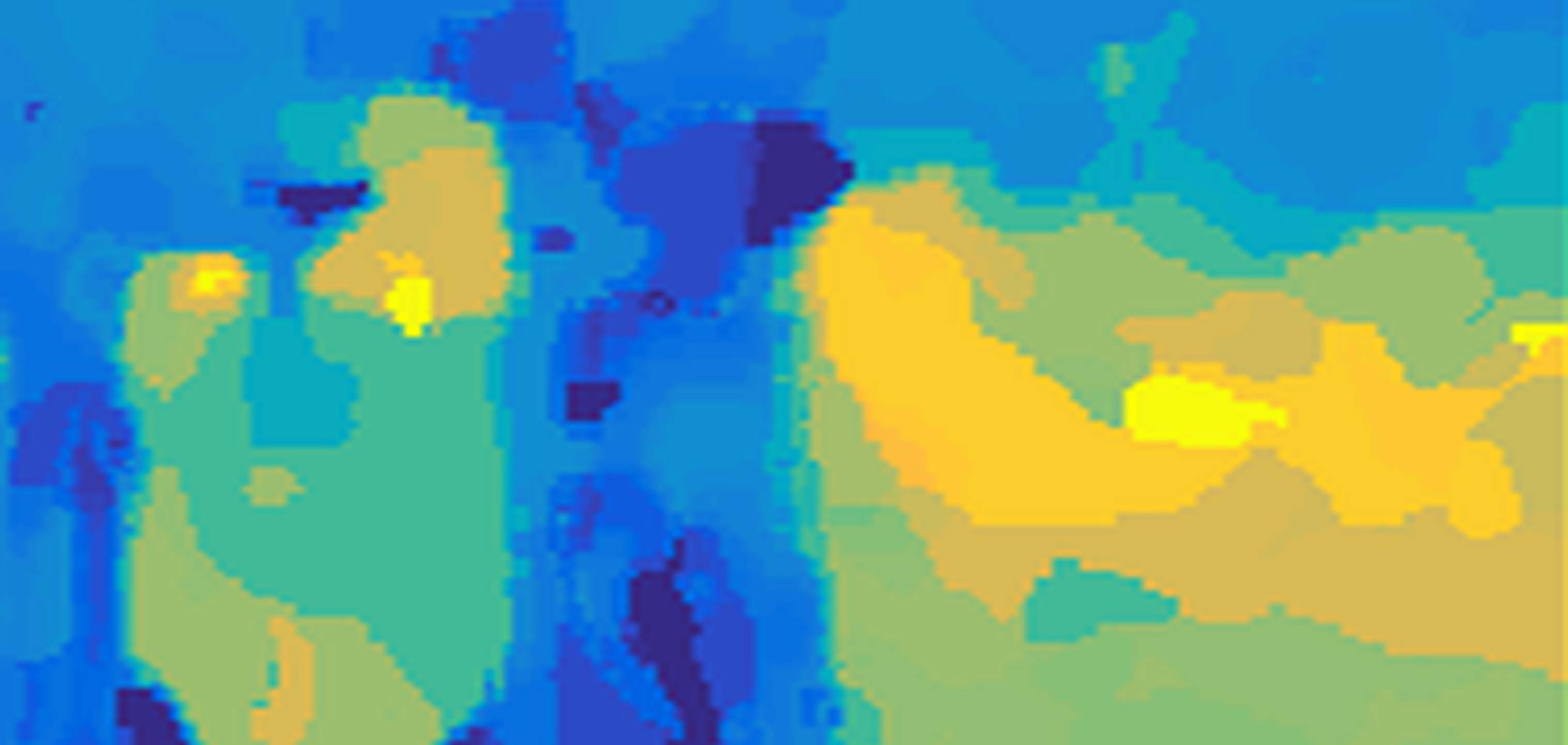} &
			\includegraphics[width=\swfour]{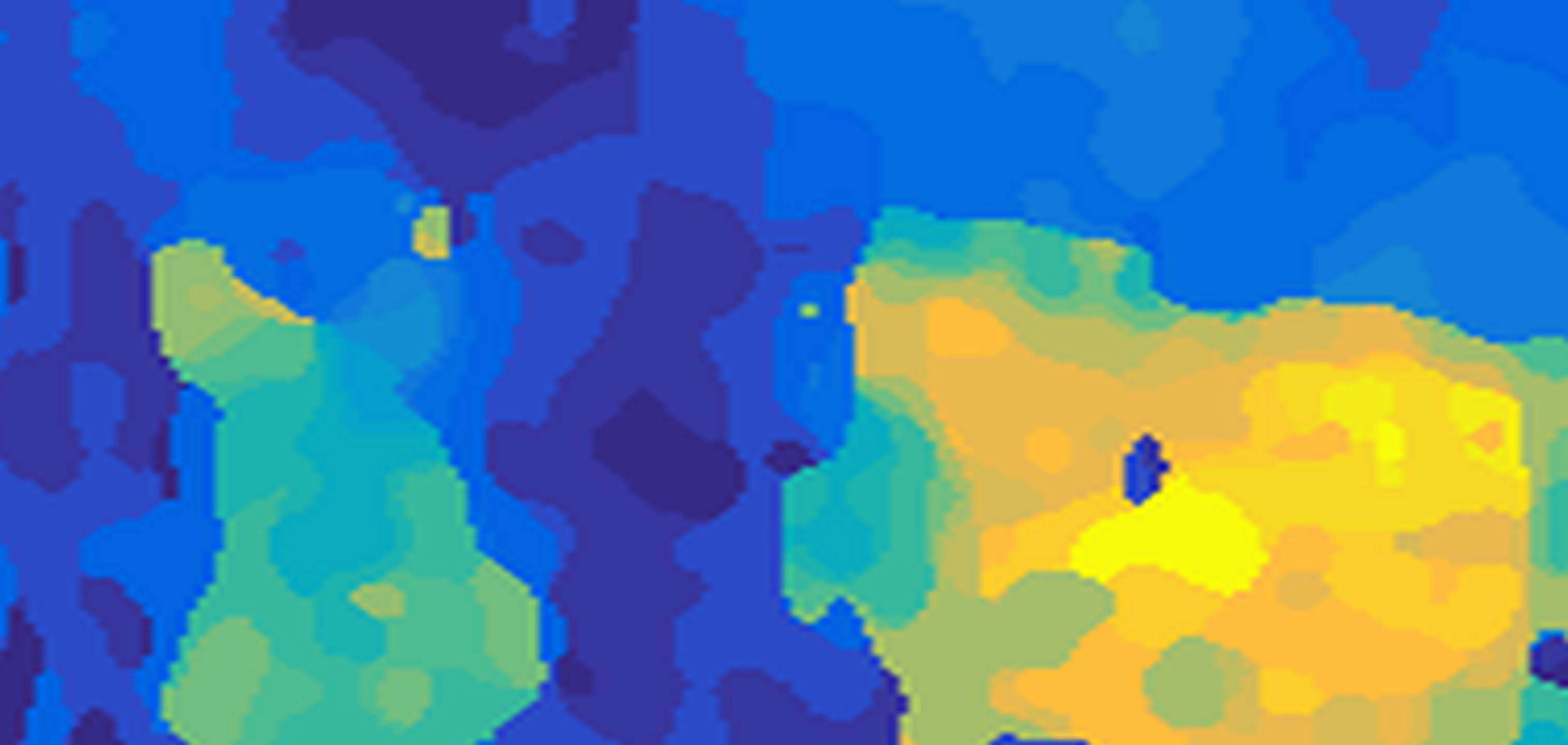} &
			\includegraphics[width=\swfour]{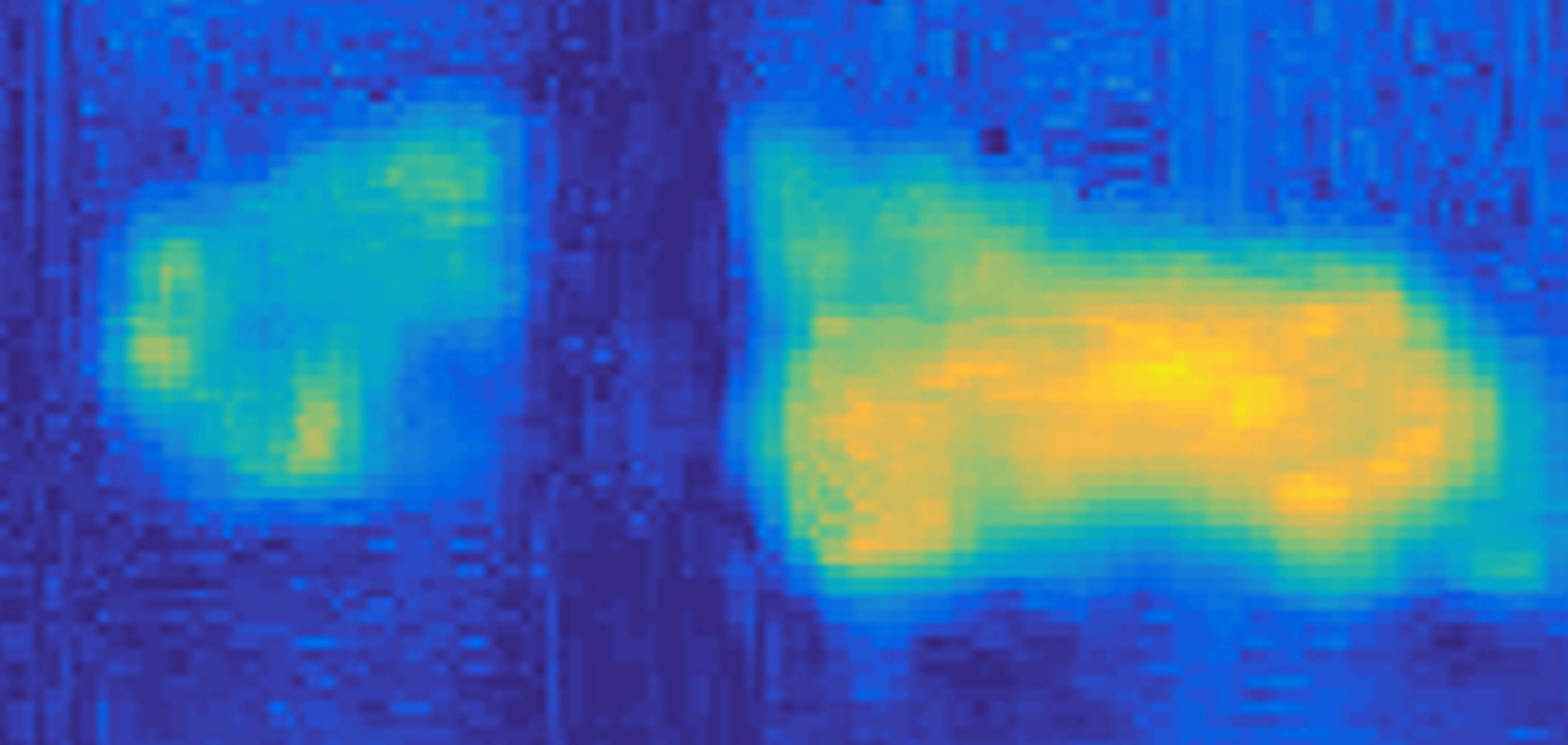} \\
			(a) blurry image & (b) horizontal flow from Sun \cite{sun2015learning} & (c) horizontal flow from Gong \cite{gong2017motion} & (d) horizontal flow from ours \\
			\includegraphics[width=\swfour]{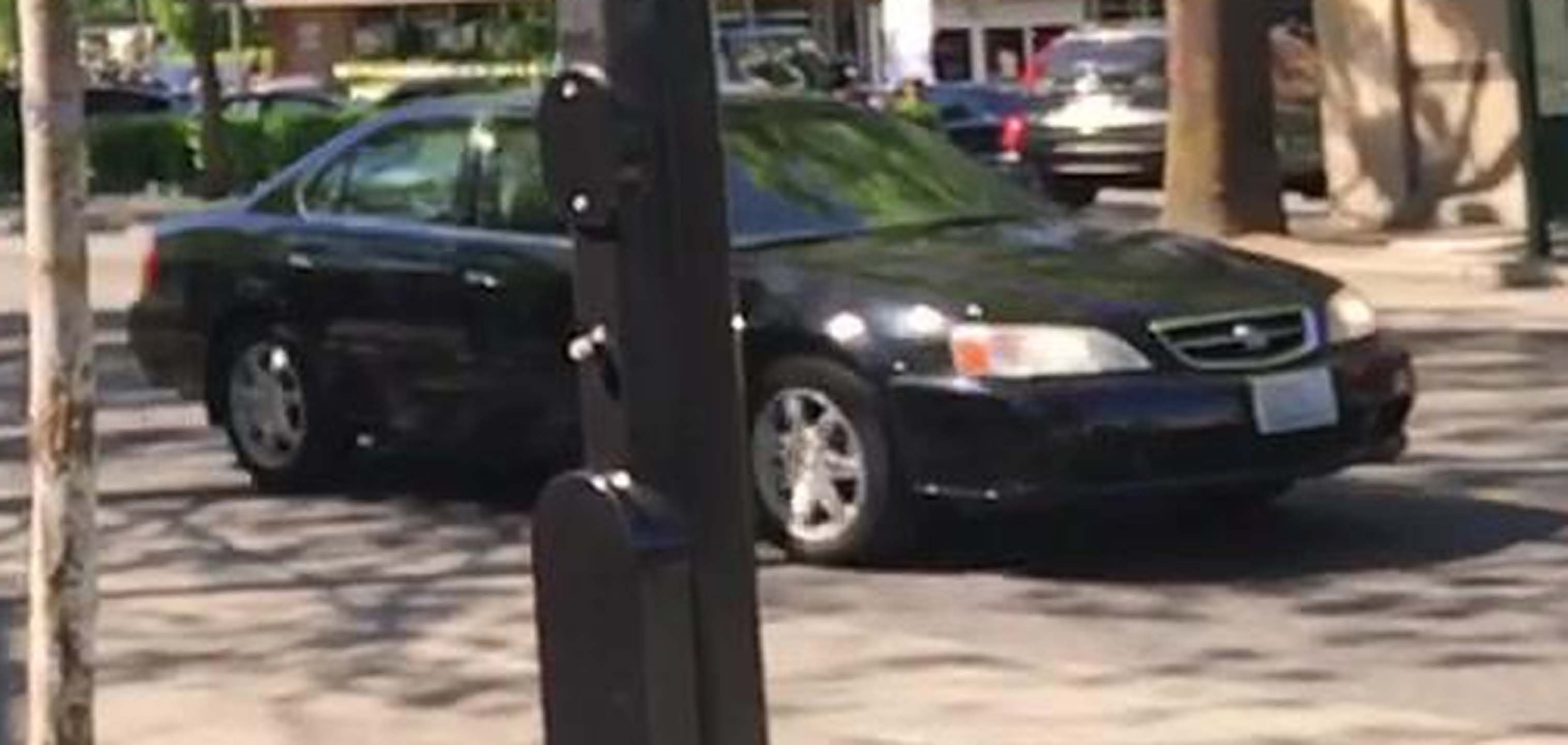} &
			\includegraphics[width=\swfour]{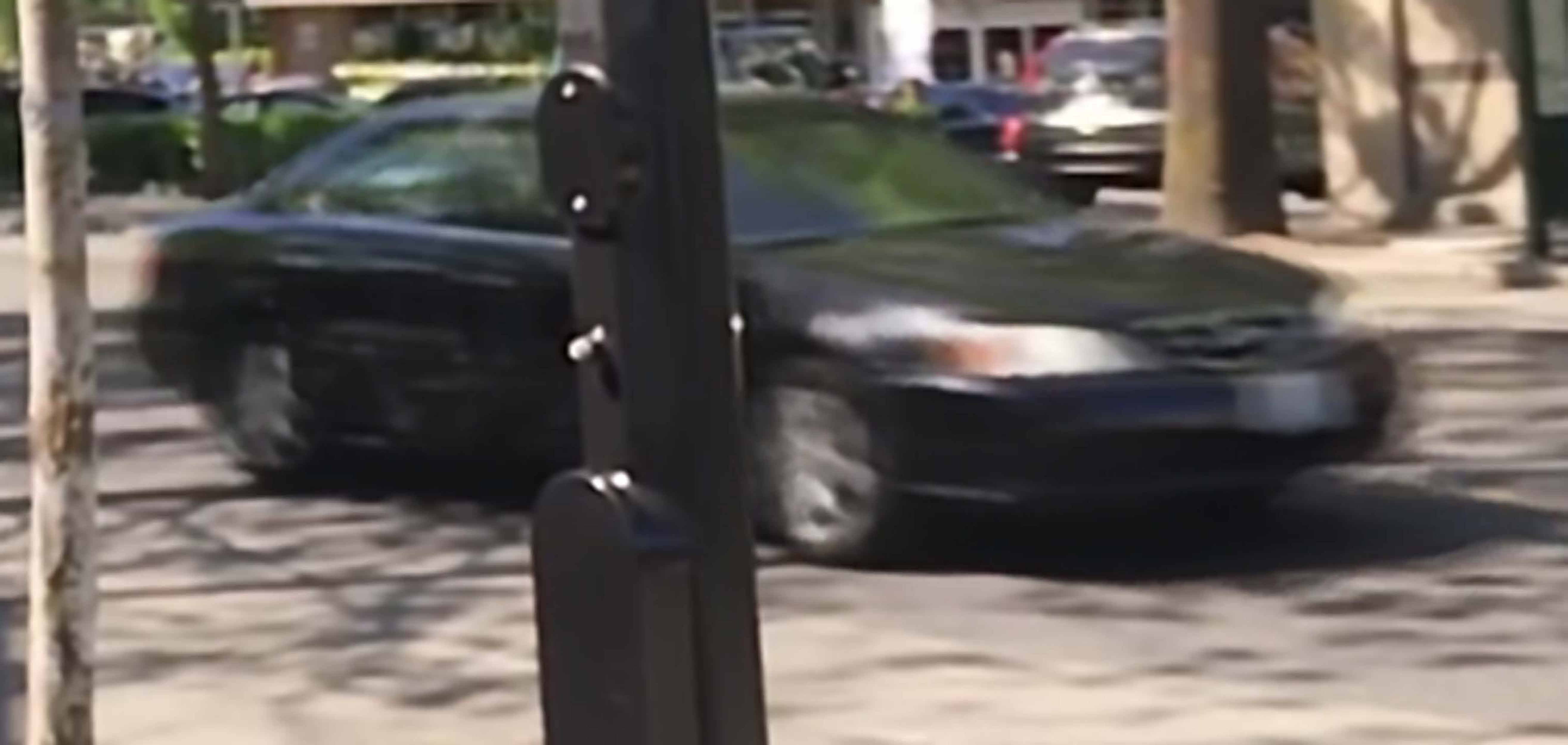} &
			\includegraphics[width=\swfour]{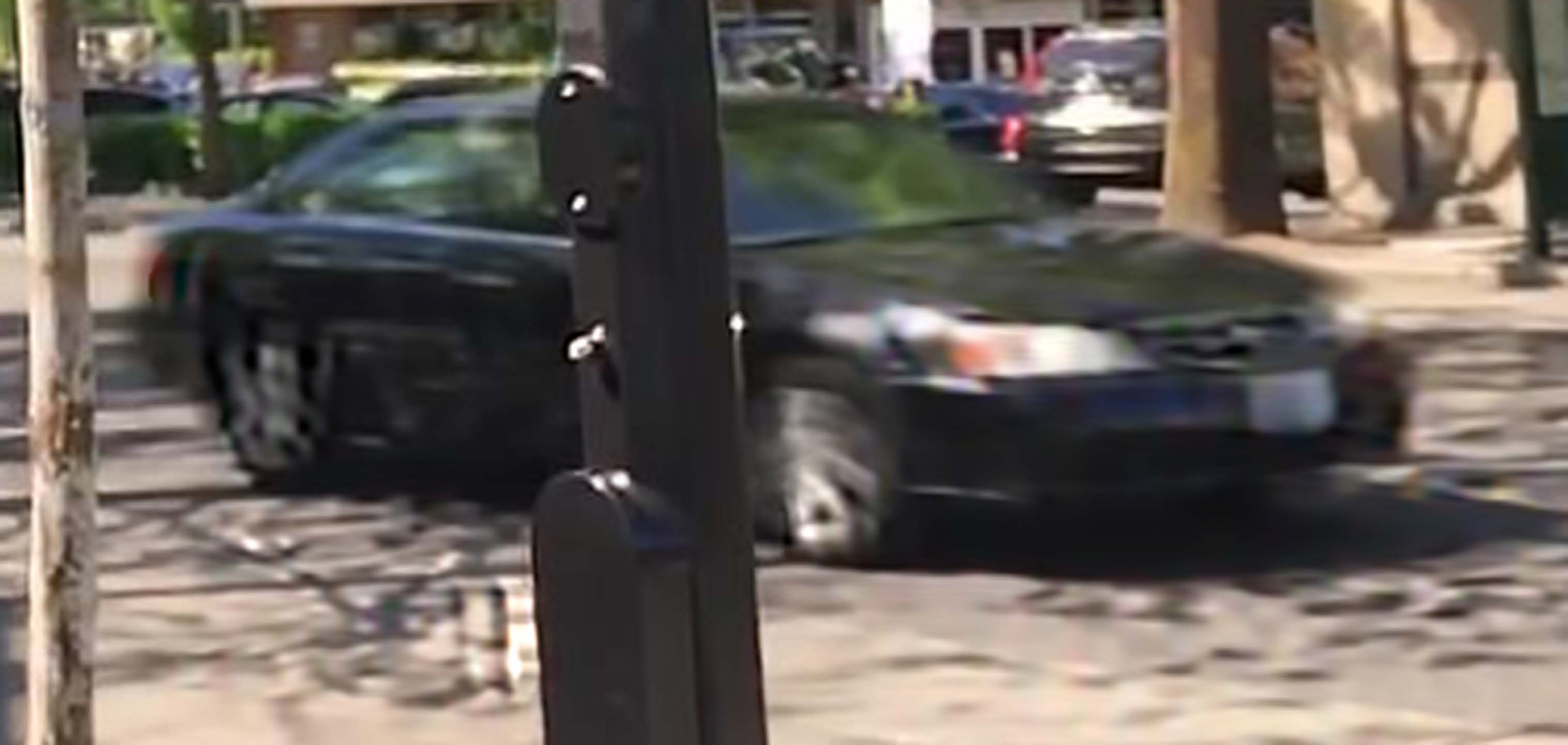} &
			\includegraphics[width=\swfour]{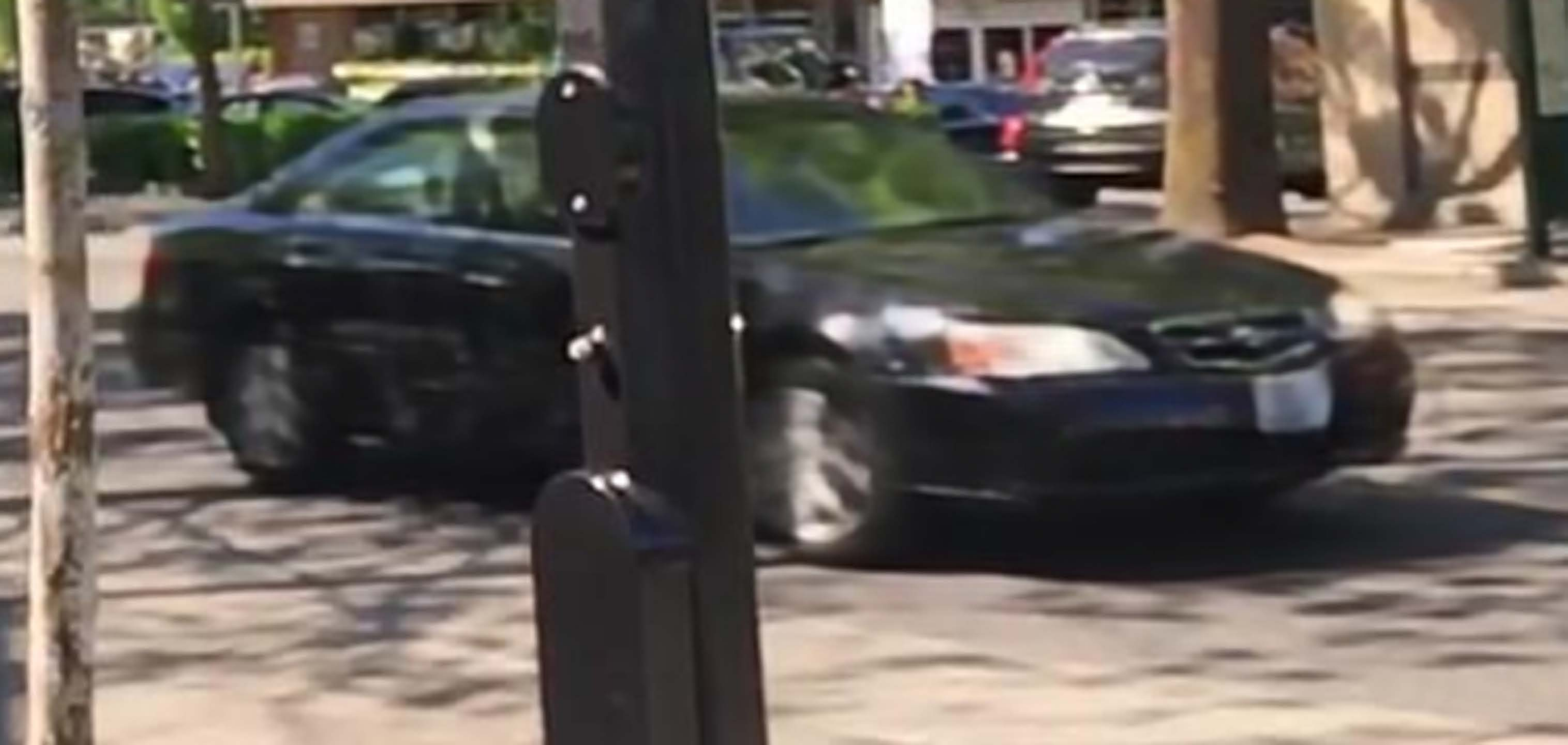} \\
			(e) clean image & (f) Sun \cite{sun2015learning} & (g) Gong \cite{gong2017motion} & (h) ours \\
		\end{tabular}
	\end{center}
	\vspace{-3mm}
	\caption{
		Effectiveness of optical flow.
		(a) and (e) are the blurry image and sharp image.
		(b) to (d) are estimated blur kernels or optical flow from \cite{sun2015learning}, \cite{gong2017motion} and the proposed network.
		(f) to (h) are the restored images from \cite{sun2015learning}, \cite{gong2017motion} and the proposed network.
		Since only one image is used to estimate the flow in \cite{sun2015learning} and \cite{gong2017motion}, they are inaccurate and generate blurry images with artifacts in (f) and (g).
	}
	\label{fig:flow0}
\vspace{-2mm}
\end{figure*}

\subsection{Running Time and Model Size}
We evaluate our algorithm and state-of-the-art dynamic scene image or video deblurring methods on the same server with an Intel(R) Xeon(R) CPU and an Nvidia Titan Xp GPU.
Table~\ref{table:runtime} shows that the time costs of traditional non-deep learning based methods~\cite{whyte2012non, kim2015generalized} are high since they need a complex optimization process.
Even though \cite{sun2015learning} and \cite{gong2017motion} propose to use CNNs to estimate spatially variant blur kernels, they still use a traditional non-blind deblurring algorithm~\cite{zoran2011learning} to restore the sharp images which is time-consuming.
Su~et al.~\cite{su2017Deep} use CNN to recover sharp images from blurry videos, but they need to utilize an optical flow algorithm \cite{liu2010high} to align the inputs of the network which is slow.
We note that \cite{nah2017deep, kupyn2017deblurgan, tao2018scale} develop an end-to-end trainable network.
However, these networks are relatively deep in order to enlarge the receptive field which is very important for deblurring.
Whereas, this will lead to a relatively large model size as well as long processing time.
To tackle the above drawbacks, both \cite{zhang2018dynamic} and the proposed network apply spatially variant RNNs.
In addition, the proposed network is more efficient than \cite{zhang2018dynamic} even with one additional input image to estimate optical flow.
By utilizing temporal information, the algorithm by~\cite{kim2017online} is more efficient and has a smaller model size.
Whereas, their performance is worse than ours.

\section{Analysis and Discussions}
We have shown that the proposed algorithm performs favorably against state-of-the-art methods.
In this section, we analyze the proposed method, compare it with the most related methods, and show how it removes blur.
For fair comparisons, we retrain the most related algorithm using the same training dataset and settings as the proposed method.

\subsection{Effectiveness of the RNNs}
In the proposed image deblurring network, the weights of the spatially variant RNNs are provided by the optical flow estimation network, so that the network can remove the blur under the guidance of the optical flow.
To validate the effectiveness of the spatially variant RNNs, we remove them as well as the optical flow estimation network for fair comparisons.
According to Table~\ref{table:ablation}, the method without using spatially variant RNNs and optical flow (`w/o RNNs' for short) cannot effectively remove blur.

To further validate the effect of the spatially variant RNNs, we replace the RNNs by concatenation layers which fuse the features (e.g., RNN weights) from the optical flow estimation network with those (e.g., RNN filtering features) from the deblurring network (`concat RNN weights' for short).
We note that even though it has the same model size relative to the proposed network, it just directly fuses the information from optical flow to features from the blurry images.
As concatenating features cannot be used as a deconvolution operation like spatial RNNs, it cannot generate favorable results compared to the proposed network.
%

\renewcommand{\tabcolsep}{10pt}
\begin{table*}[!t]\footnotesize
	\centering
	\caption{Ablation study w.r.t. the proposed algorithm on video deblurring dataset~\cite{su2017Deep}.
		`w/o RNNs' denotes that we remove all the RNNs as well as the flow estimation network;
		`concat RNN weights' denotes that we replace RNNs by concatenating the features from flow estimation network with the deblurring network;
		`w/o flow losses' denotes that we remove the optical flow constraint;
		`w/o flow' removes the optical flow constraint as well as the second blurry image as the input of the network.
		`two direction flows' denotes that we three blurry images as the input of the optical flow estimation network to estimate forward and backward optical flow;
		`RNNs in decoder' denotes that we set RNNs in the decoder of the deblurring network.
	}
	\label{table:ablation}
	\vspace{-2mm}
	\begin{tabular}{c|cc|ccc|c|c}
		\toprule
		& \multicolumn{2}{|c|}{For RNNs} & \multicolumn{3}{c|}{For optical flow} & For features & \\
		& w/o RNNs & Concat RNN Weights & w/o Flow & w/o Flow Losses & Bidirection Flows & RNNs in Decoder & Ours \\
		\midrule
		PSNR & 28.84 & 29.71 & 30.11 & 30.35 & 30.56 & 29.72 & 30.58 \\
		SSIM  & 0.901 & 0.914 & 0.922 & 0.924 & 0.927 & 0.914 & 0.928 \\
		\bottomrule
	\end{tabular}
\vspace{-3mm}
\end{table*}

\renewcommand{\tabcolsep}{1pt}
\begin{figure*}[t]\footnotesize
	\centering
	\begin{tabular}{cccc}
		\includegraphics[width=0.23\linewidth]{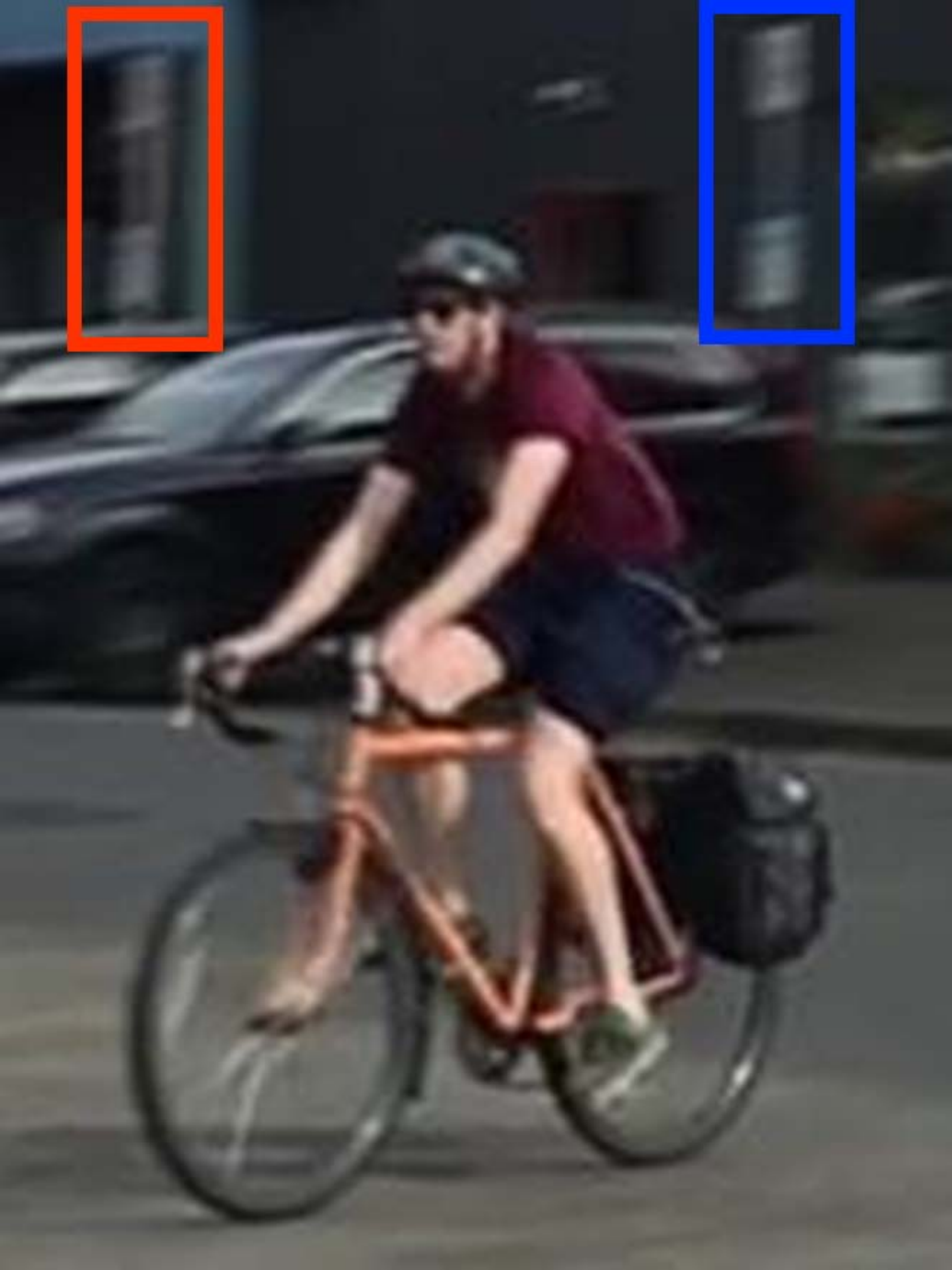} &
		\includegraphics[width=0.23\linewidth]{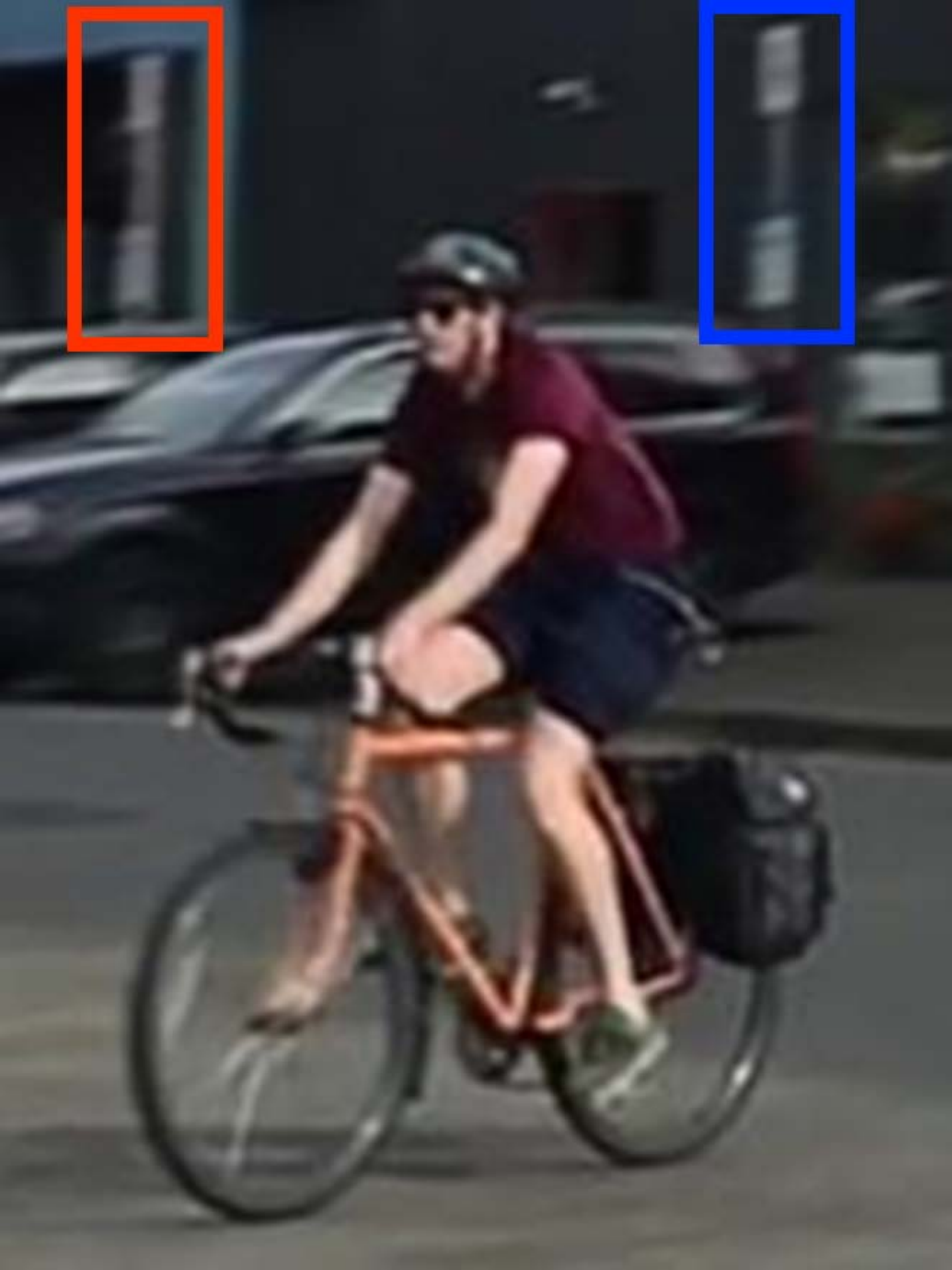} &
		\includegraphics[width=0.23\linewidth]{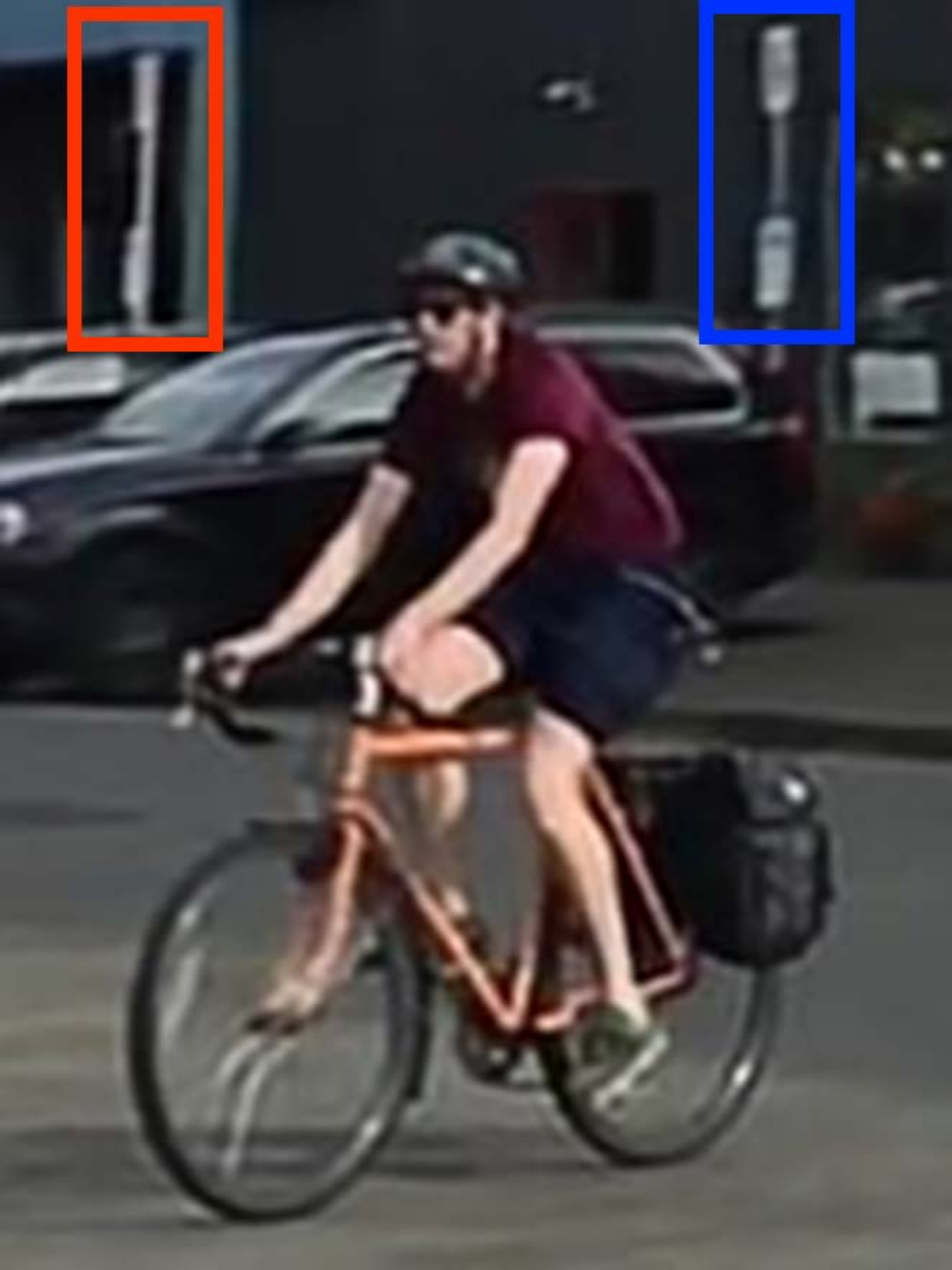} &
		\includegraphics[width=0.23\linewidth]{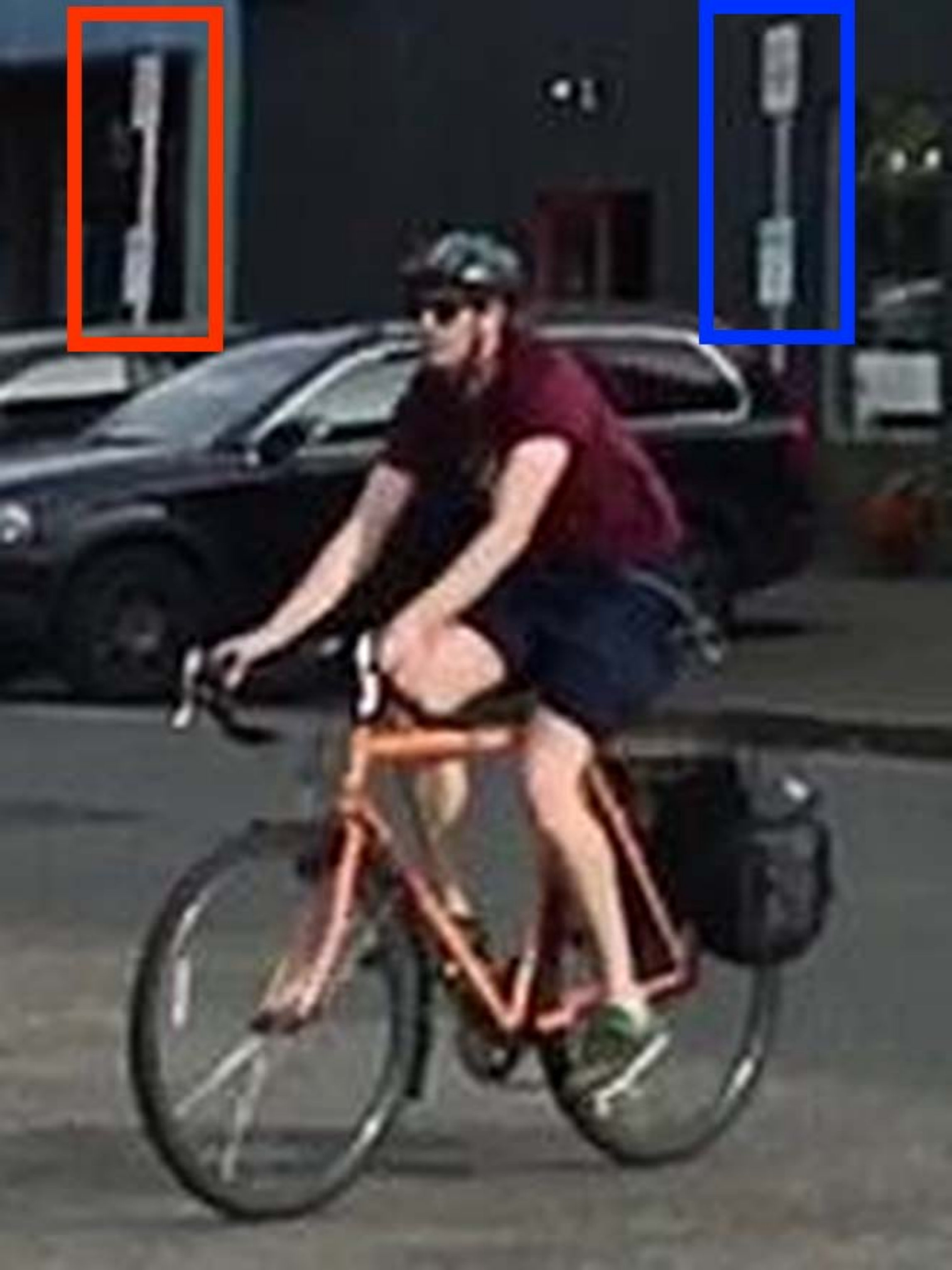} \\
		(a) blurry image & (b) without optical flow & (c) ours & (d) clean image \\
	\end{tabular}
	\caption{
		Effectiveness of the optical flow to remove the blur in the background region.
		To remove the influence of the optical flow, the inputs of the optical flow estimation network are two exactly the same blurry images.
		In this way, the network cannot remove blur which is shown in (b).
		With the help of the consecutive blurry image to estimate the optical flow, the proposed network can remove the blur properly in (c).
	}
	\label{fig:flow2}
\end{figure*}

\renewcommand{\tabcolsep}{1pt}
\begin{figure*}[!t]\footnotesize
	\begin{center}
		\begin{tabular}{ccc}
			\includegraphics[width=\swthree]{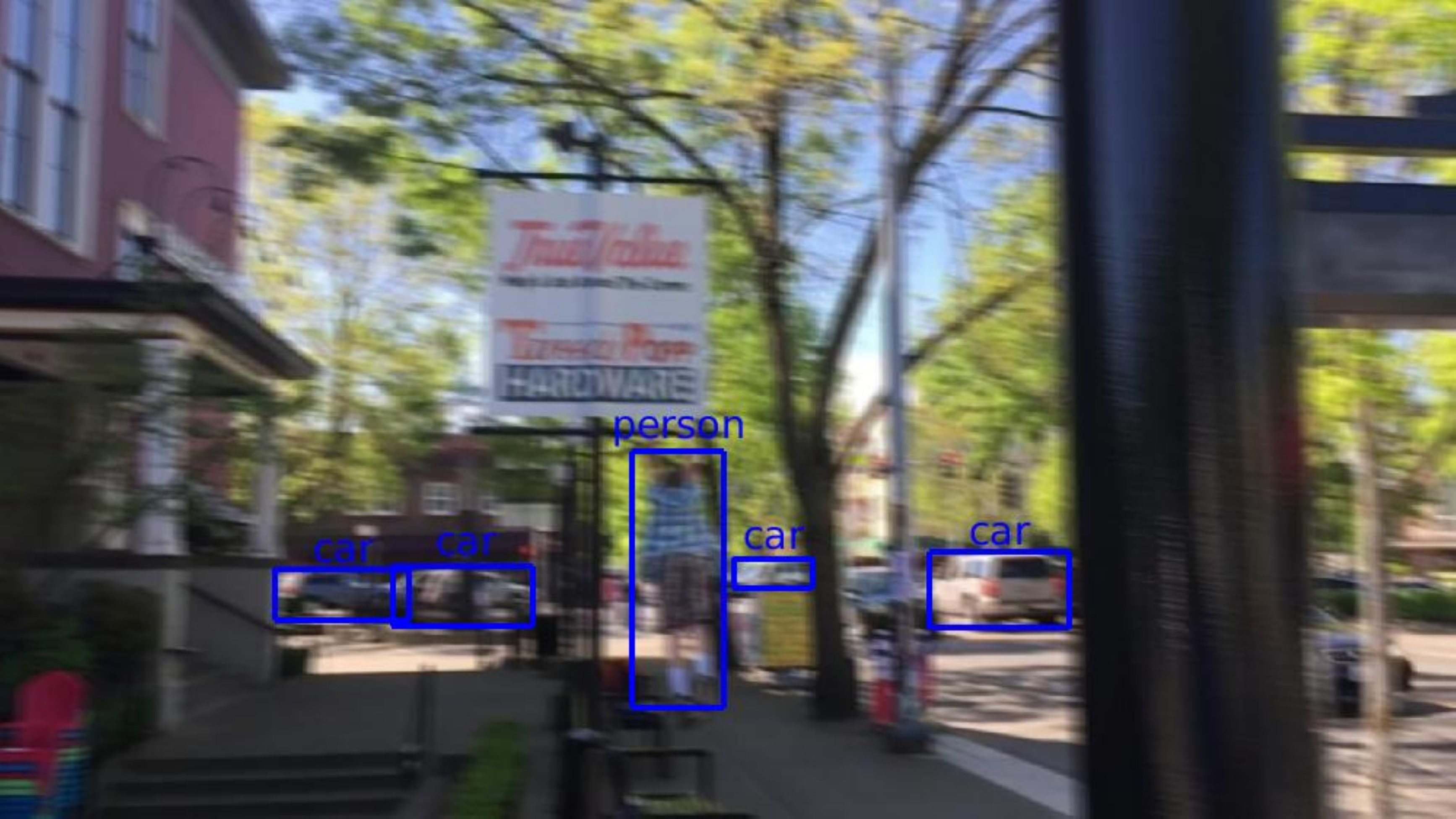} &
			\includegraphics[width=\swthree]{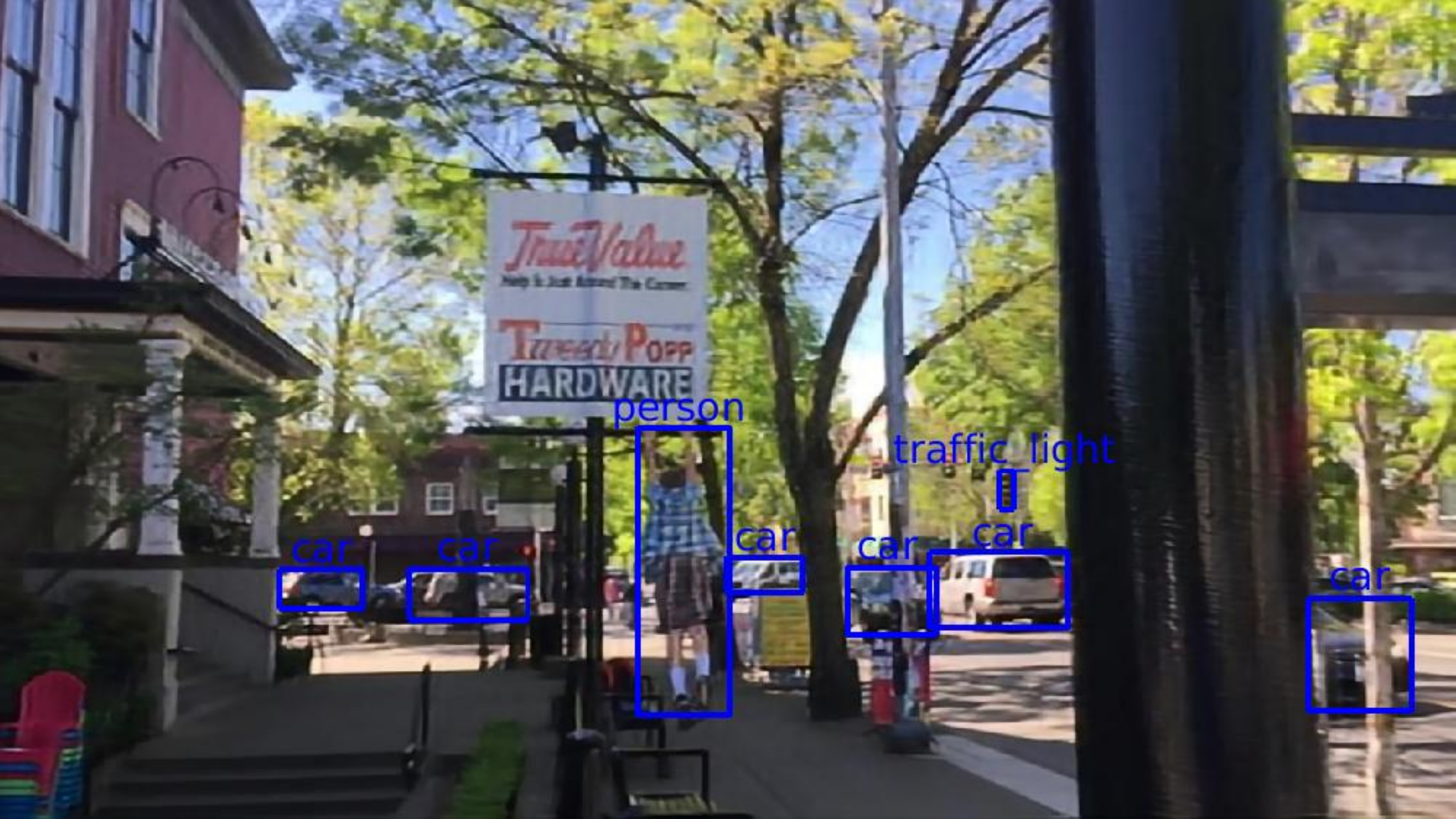} &
			\includegraphics[width=\swthree]{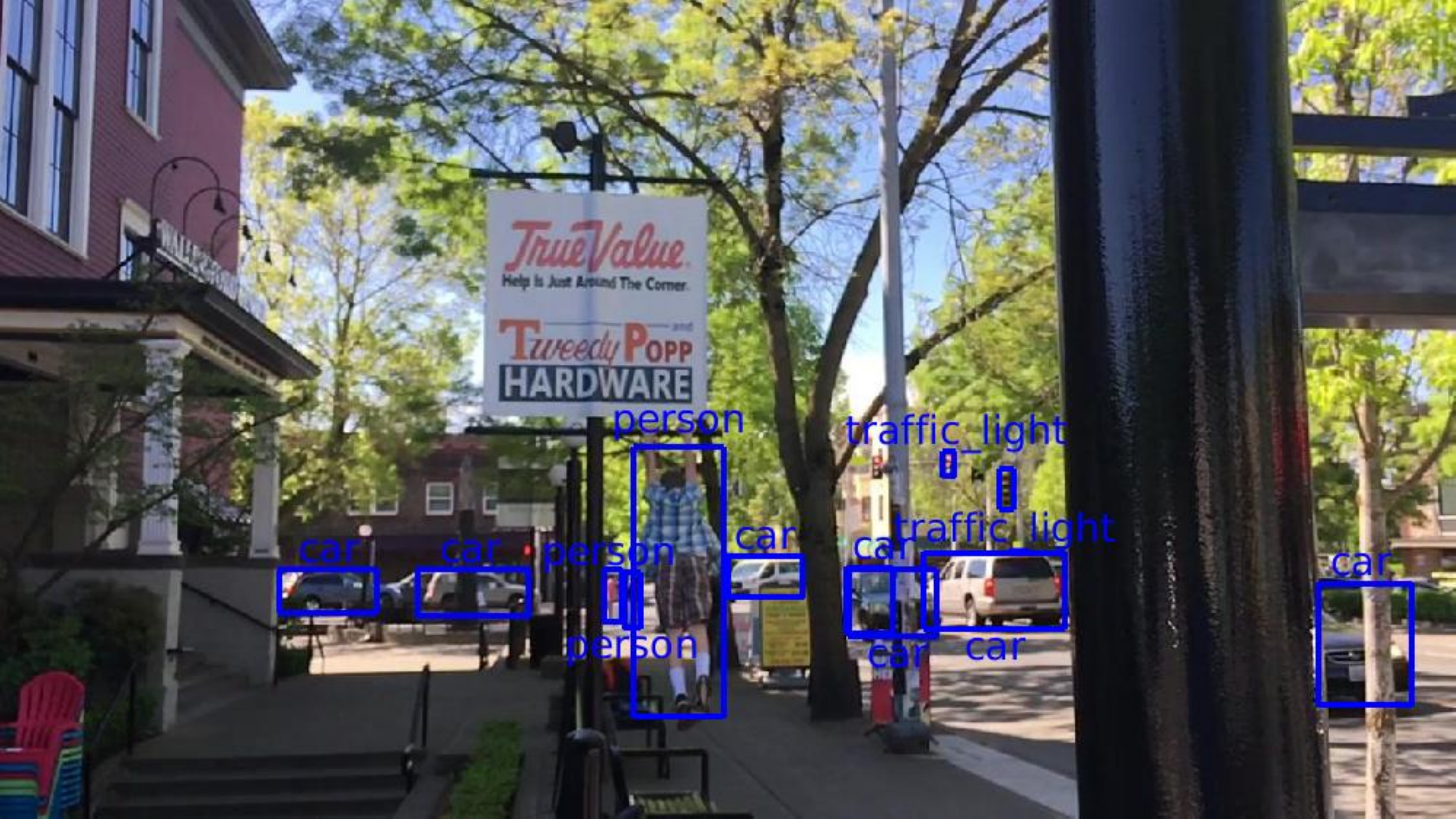} \\
			\includegraphics[width=\swthree]{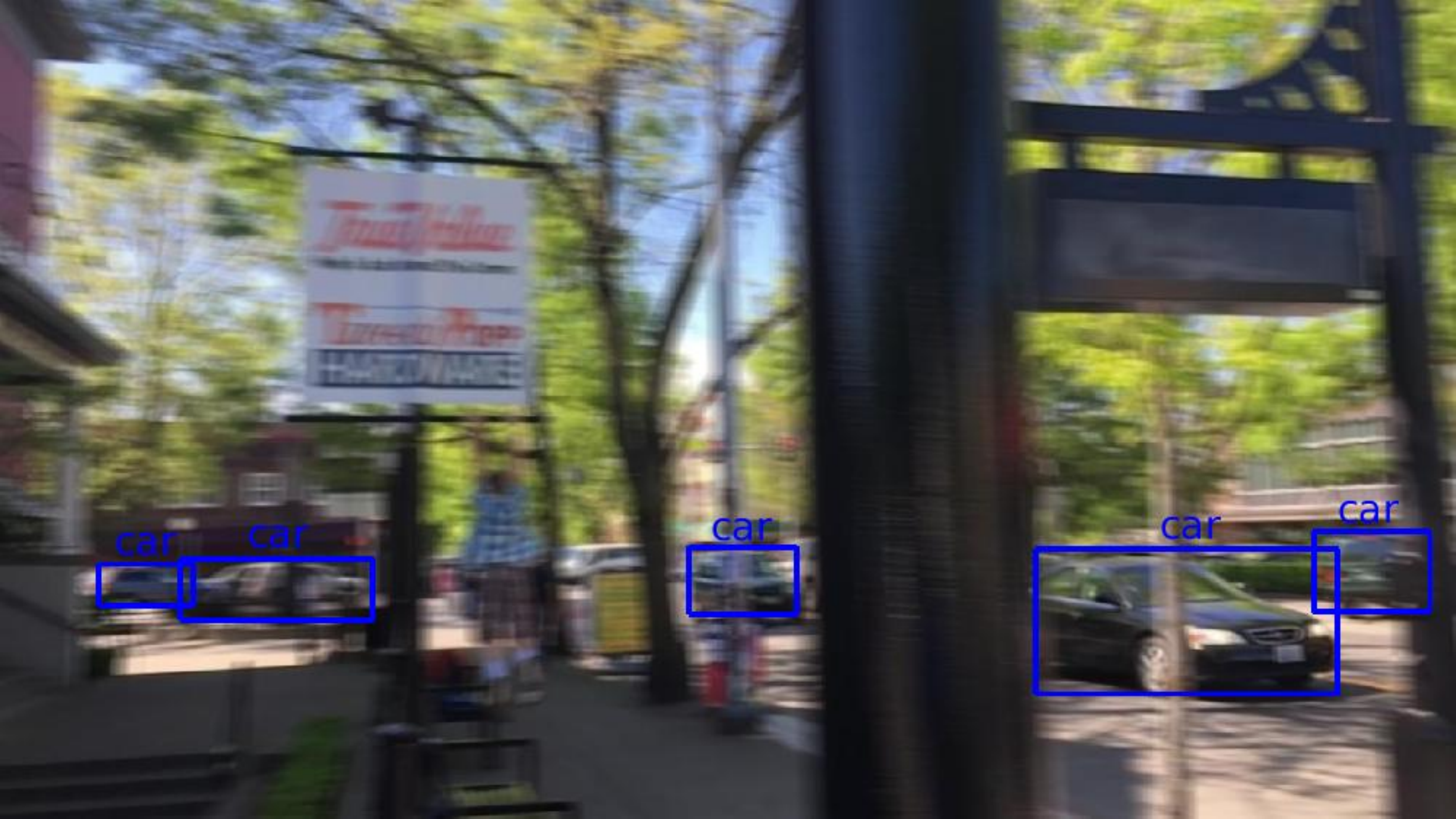} &
			\includegraphics[width=\swthree]{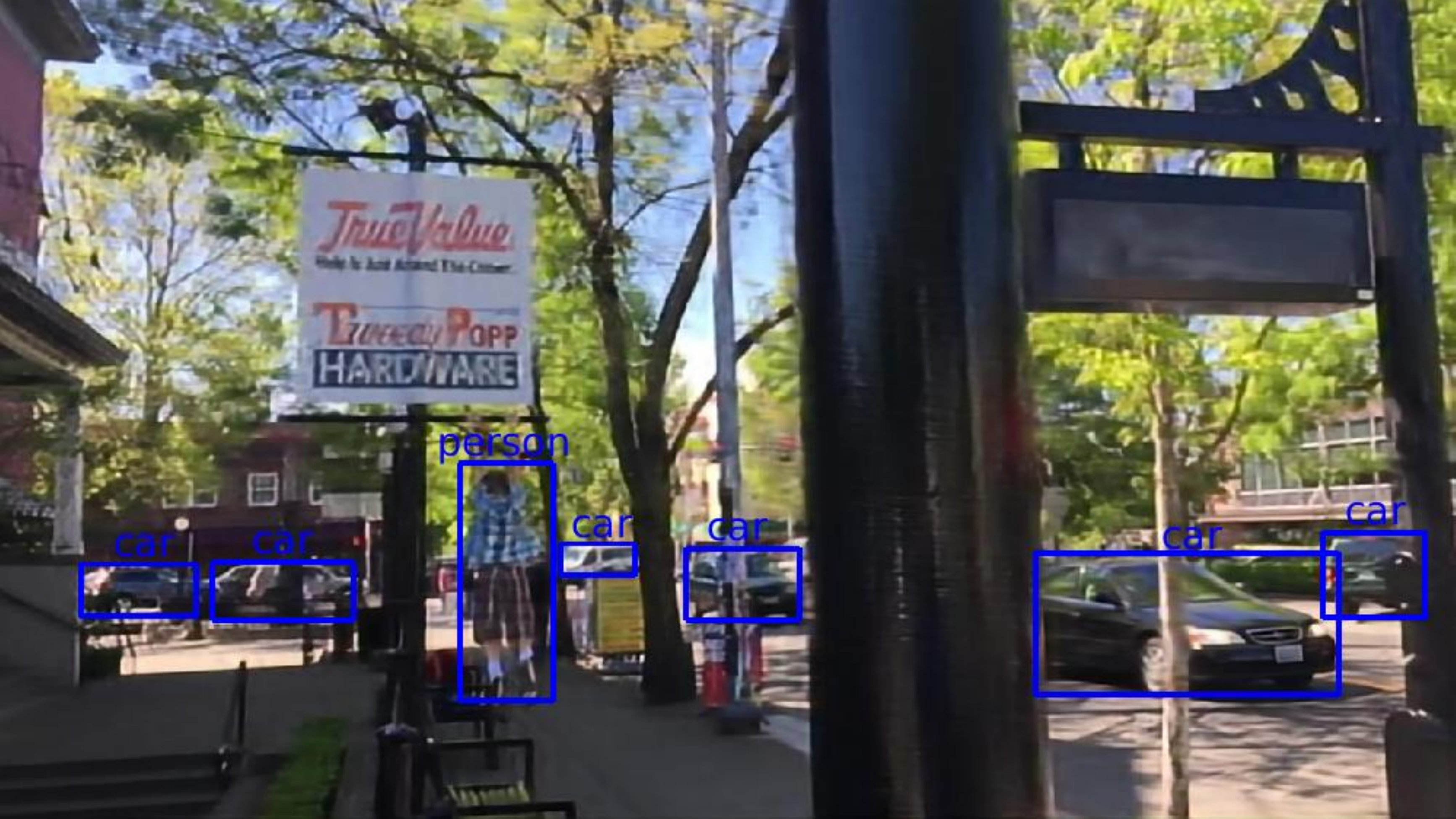} &
			\includegraphics[width=\swthree]{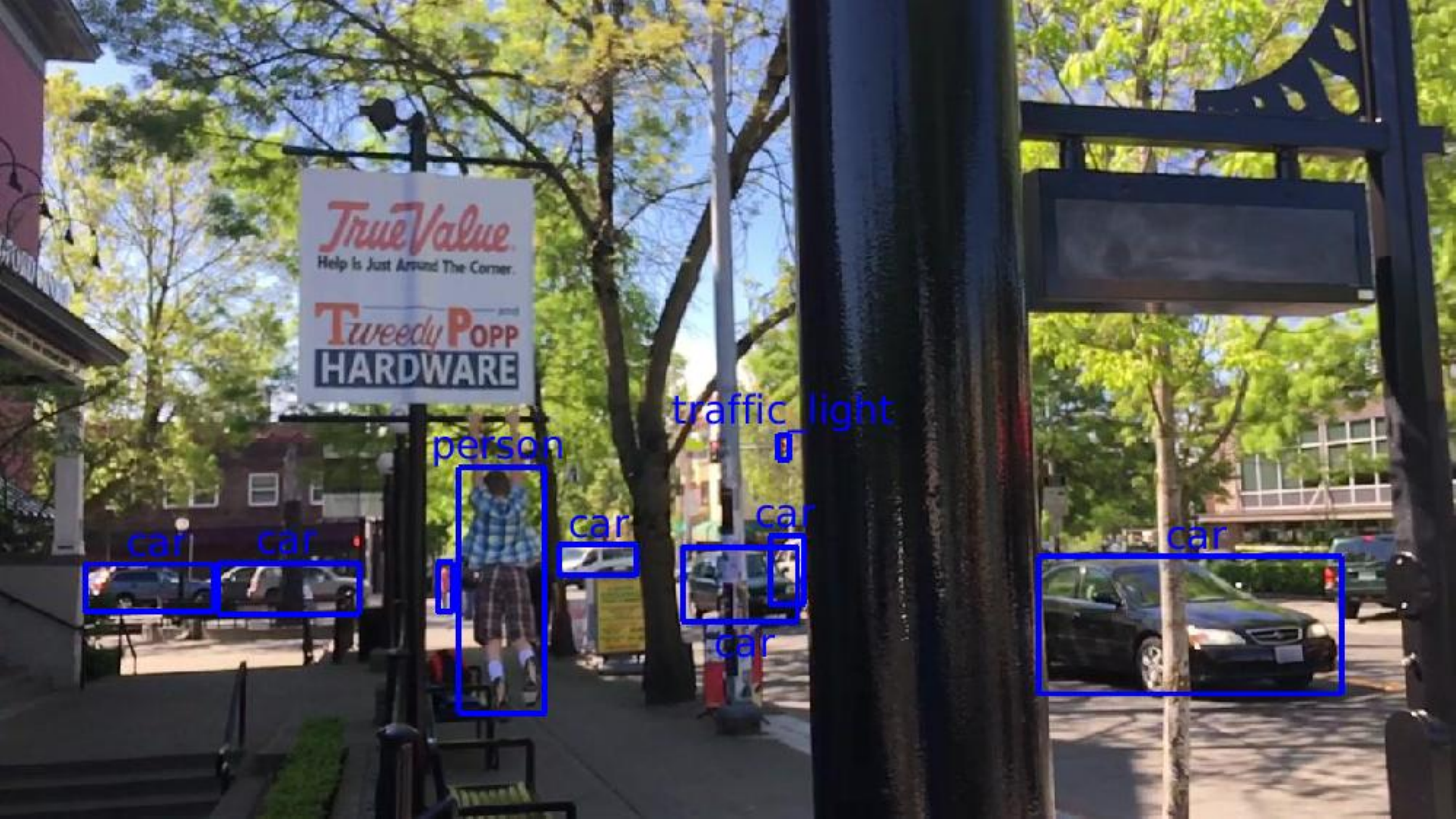} \\	
			blur & our deblur & sharp		
		\end{tabular}
	\end{center}
	\vspace{-3mm}
	\caption{
		Object detection comparison for blurry image, our deblurred image as well as sharp image.
		With the help of our proposed deblurring network, more objects can be detected especially for the smaller objects.
	}
	\label{fig:detect}
\vspace{-2mm}
\end{figure*}

\begin{figure*}[!t]\footnotesize
	\begin{center}
		\begin{tabular}{cccc}
			\includegraphics[width=\swfour]{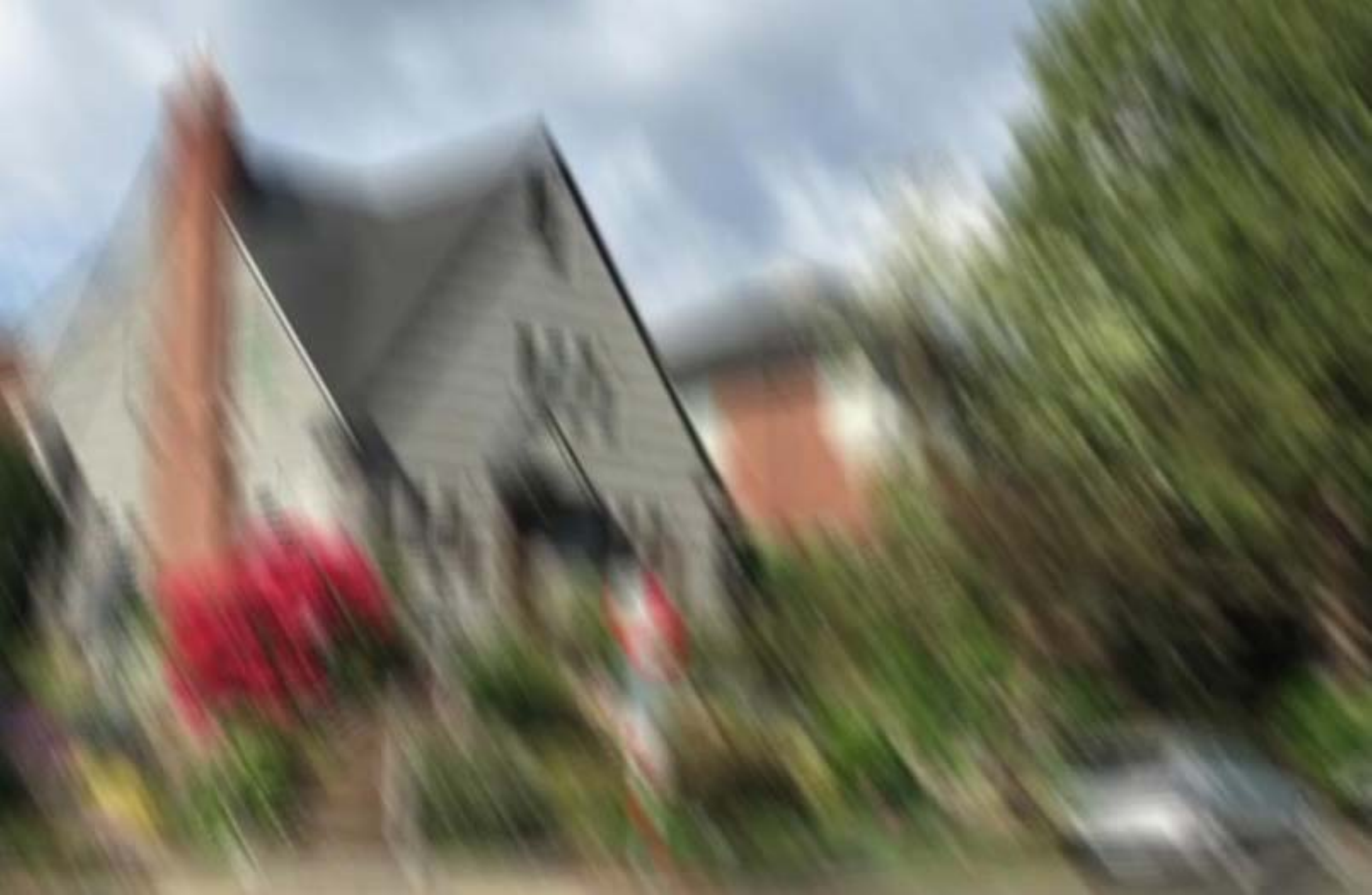} &
			\includegraphics[width=\swfour]{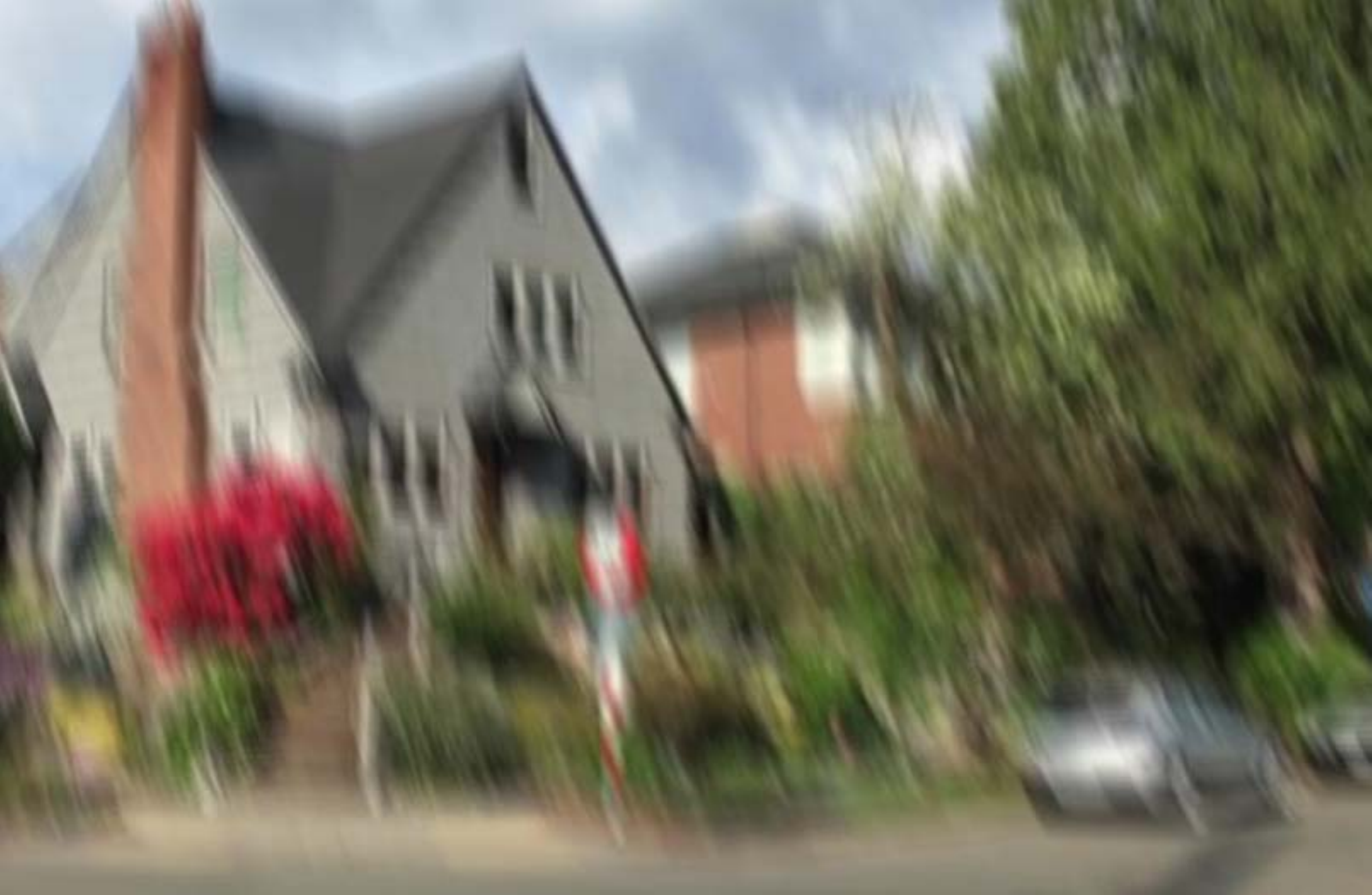} &
             \includegraphics[width=\swfour]{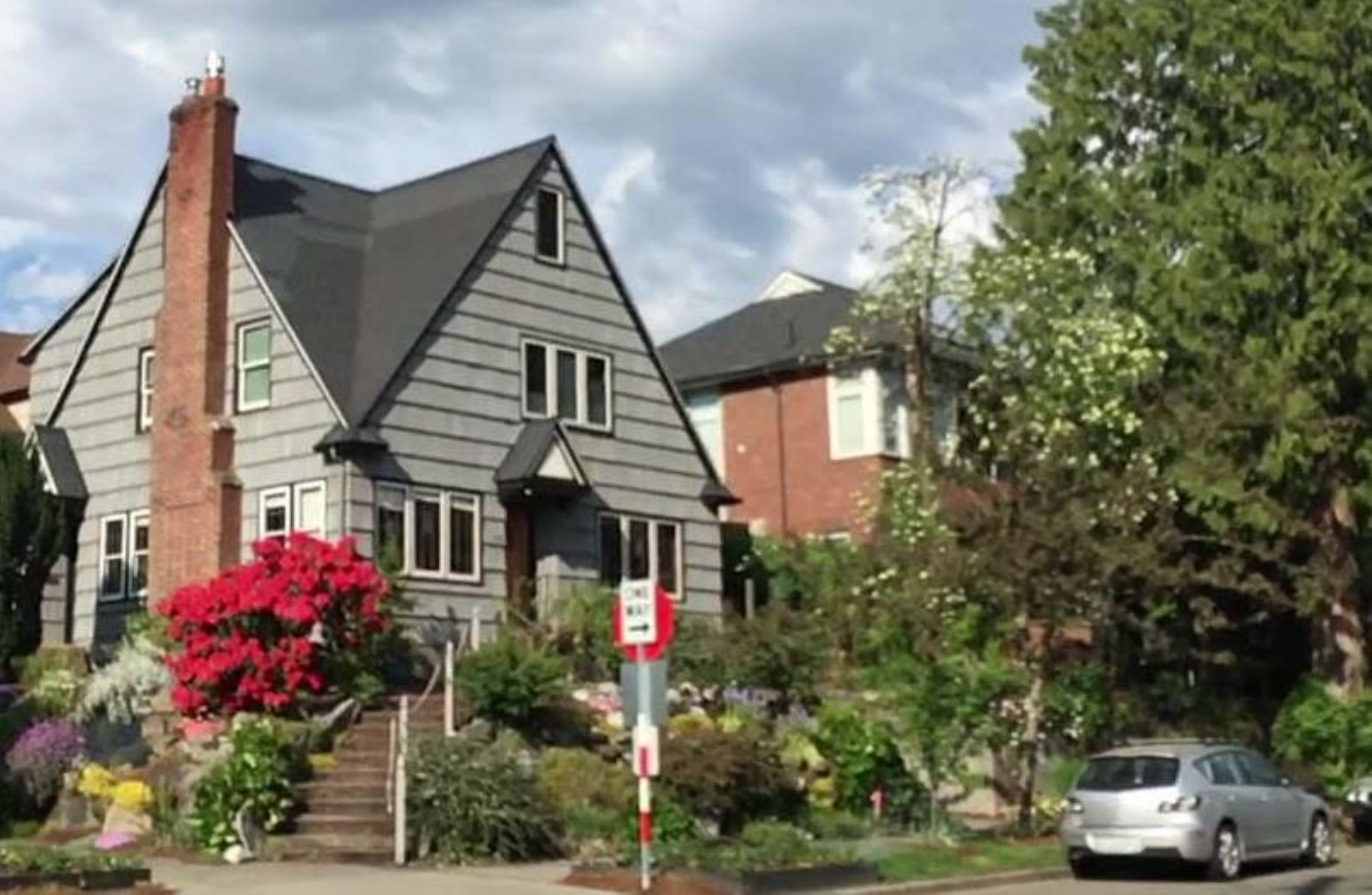} 	 &
             \includegraphics[width=\swfour]{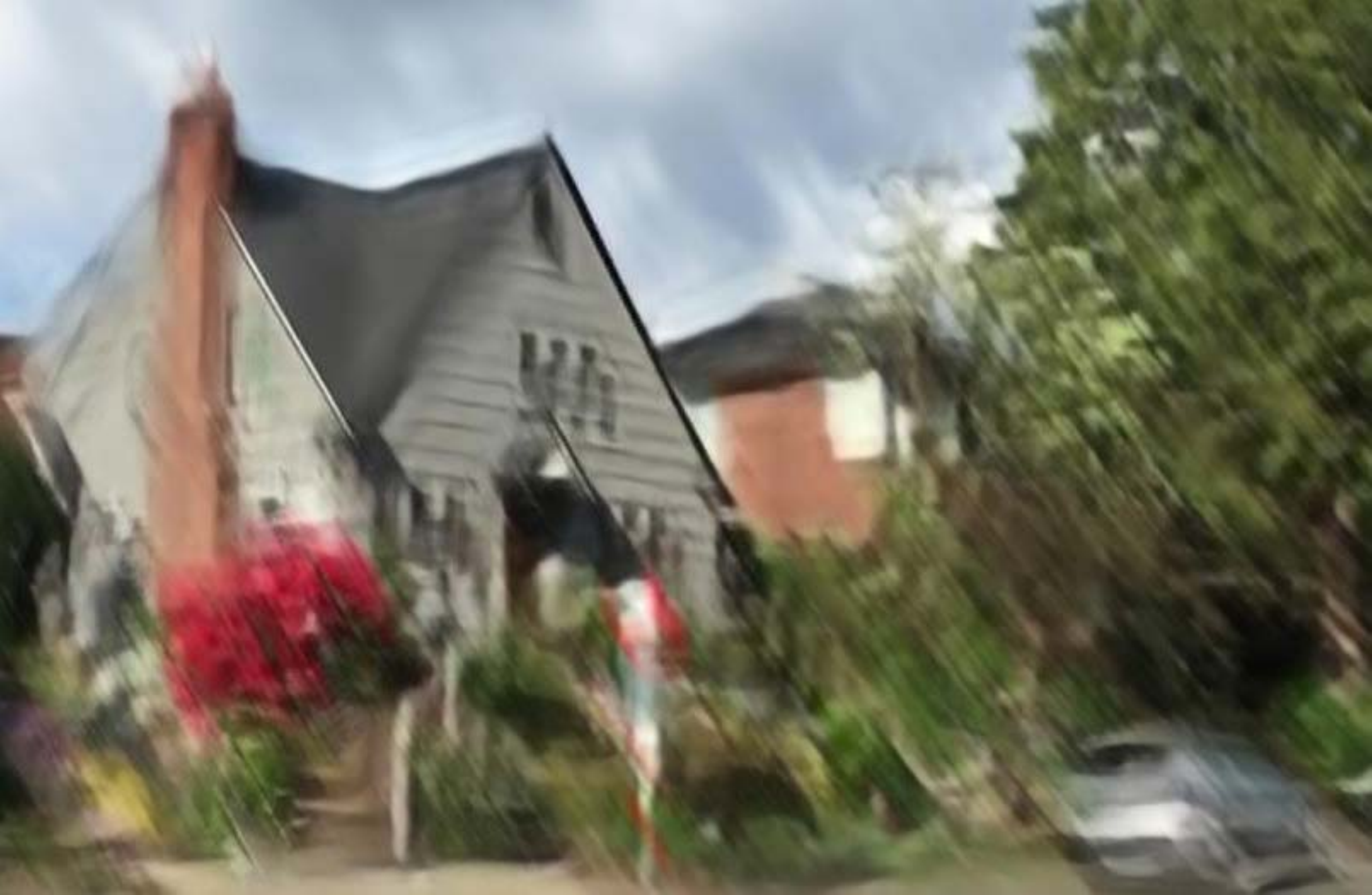}\\
			\multicolumn{2}{c} {(a) two consecutive blurry images} & (b) clean image & (c) Nah \cite{nah2017deep} \\
			\includegraphics[width=\swfour]{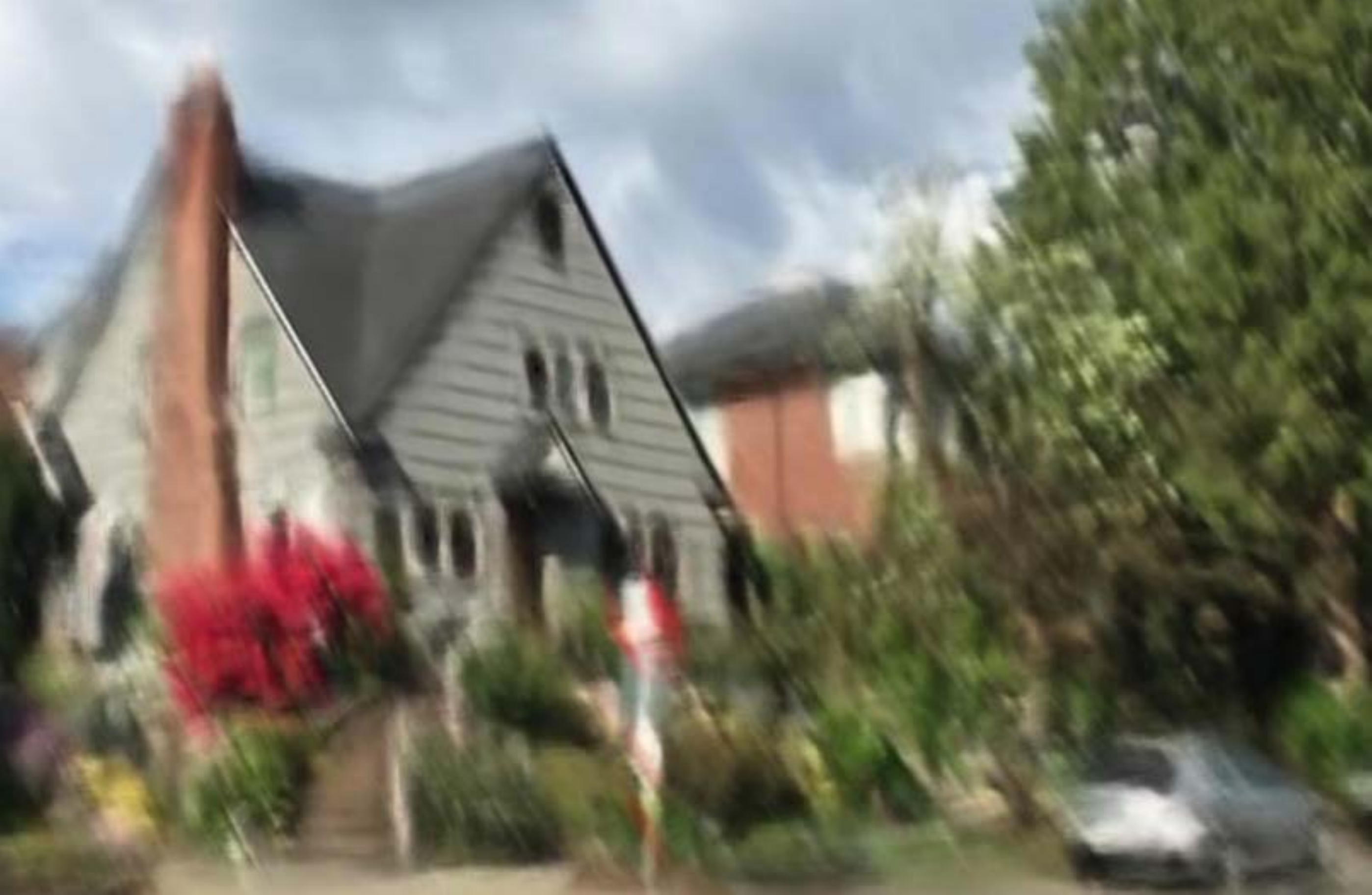} &
			\includegraphics[width=\swfour]{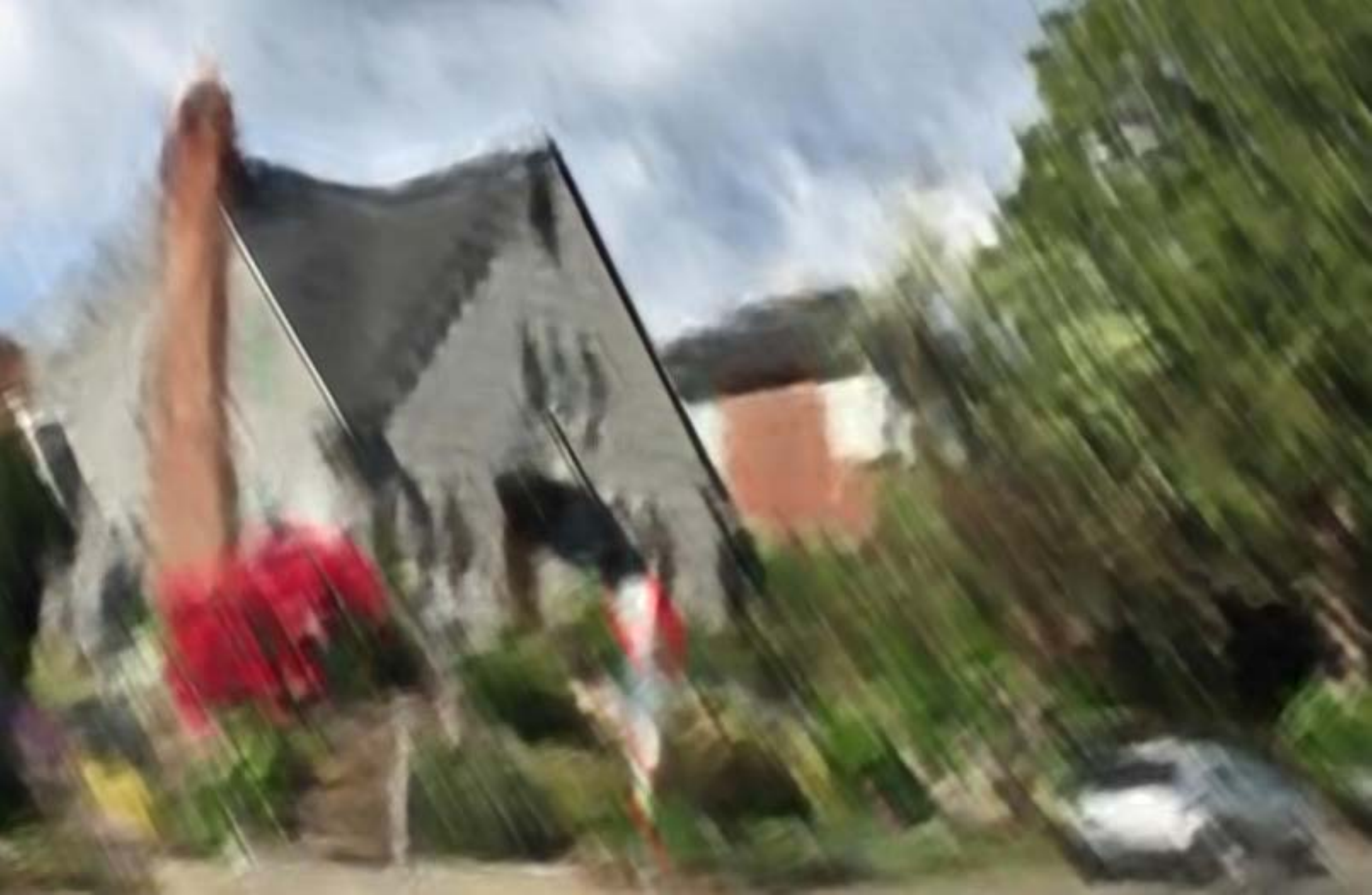} &
			\includegraphics[width=\swfour]{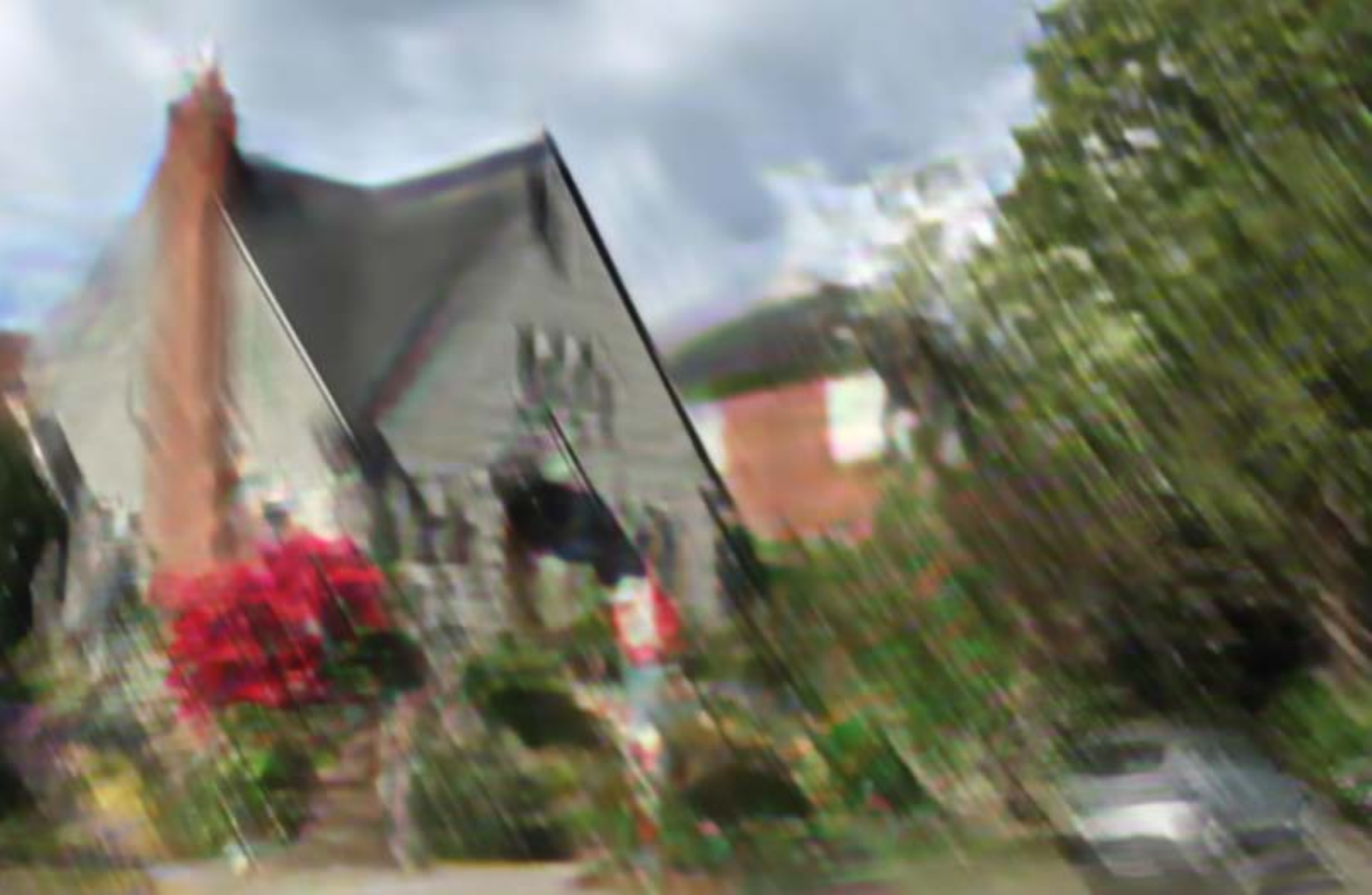} &
            \includegraphics[width=\swfour]{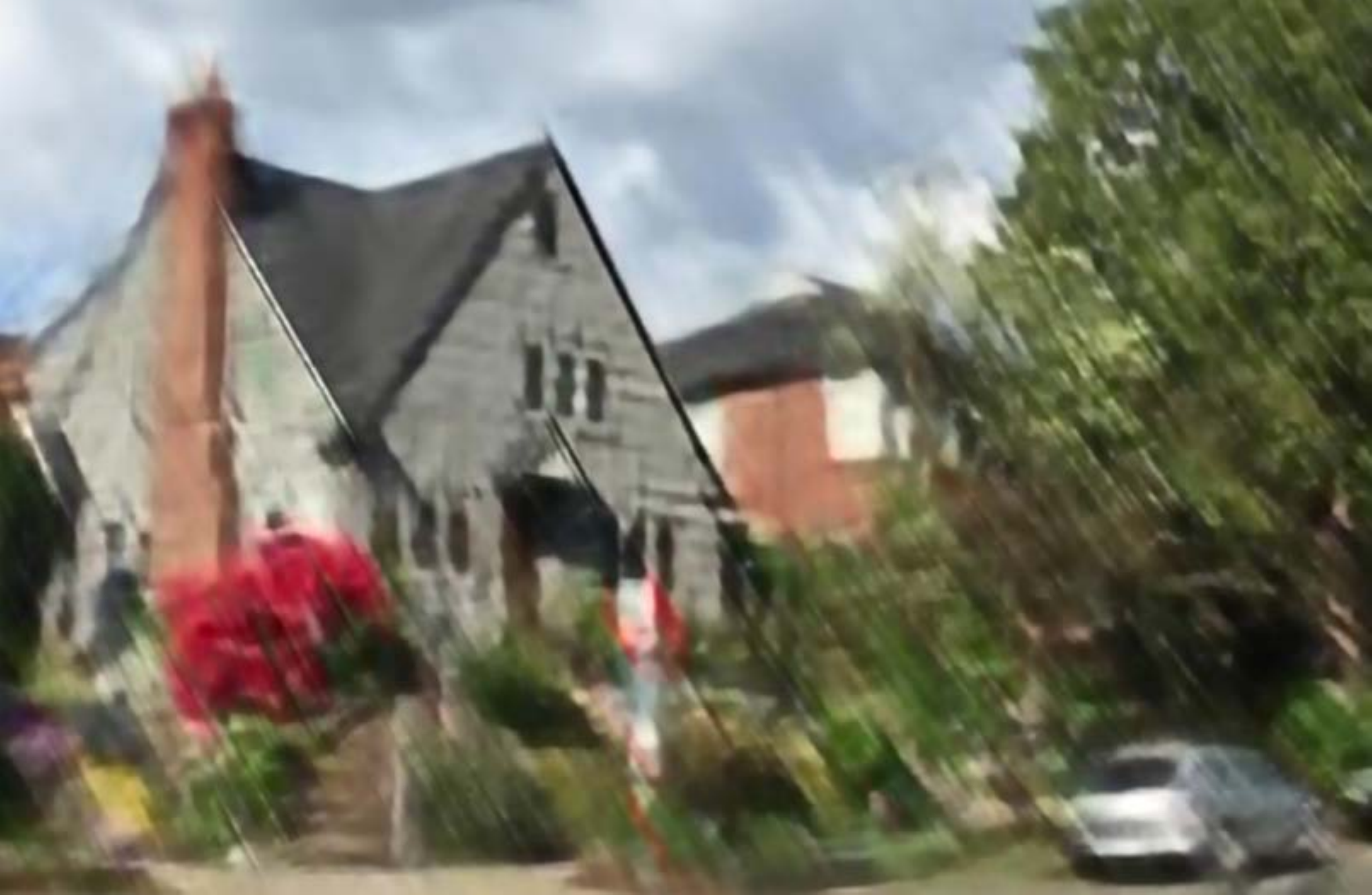}\\
			(d) Su \cite{su2017Deep} & (e) Kim  \cite{kim2017online} & (f) Tao \cite{tao2018scale} & (g) ours \\
		\end{tabular}
	\end{center}
	\vspace{-3mm}
	\caption{
		Limitations of the proposed method.
		Similar to other methods, our method cannot handle images with large blur well and there still exist blur and artifacts in the restored images.
		In addition, two consecutive blurry images (a)-(b) may have appearance differences when the blur is large.
		This scenario leads to an inaccurate optical flow estimation, and the proposed method cannot handle it well as shown in (g).
	}
	\label{fig:limit}
	\vspace{-3mm}
\end{figure*}

\subsection{Effectiveness of the Optical Flow}
As the spatially variant RNN is proposed by~\cite{zhang2018dynamic}, one may wonder that the deblurring effect is mainly due to the use of spatially variant RNN.
However, we show that optical flow also plays an indispensable role in image deblurring.

From the analysis of Section~\ref{subsec:motivation}, the spatially variant blur kernel has a connection with the optical flow which is estimated from two blurry images.
To demonstrate the effect of the optical flow, we compare our method with the most related method~\cite{zhang2018dynamic}.
In Figure~\ref{fig:flow1}, all state-of-the-art end-to-end learning methods~\cite{nah2017deep, tao2018scale, zhang2018dynamic} treat the support beam of the bridge as a blurry region and generate artifacts.
However, there is no strong motion in this region according to the optical flow estimation.
By utilizing the optical flow information, our network does not consider this region to be blurry and correctly restores a clear image.

We also compare the proposed network with two methods \cite{sun2015learning, gong2017motion} which use a single image to estimate spatially variant blur kernels.
It shows that they cannot accurately estimate the blur kernel according to Figure~\ref{fig:flow0}(b) and (c).
With inaccurate blur kernels, \cite{sun2015learning, gong2017motion} will generate blurry images with artifacts which can be seen from Figure~\ref{fig:flow0}(f) and (g).
As the proposed network manages to estimate the optical flow more accurately using two successive images, it can restore cleaner and sharper images in Figure~\ref{fig:flow0}(h).

To further validate the effectiveness of the optical flow, the inputs of the optical flow estimation network are two exactly the same blurry image and the estimated optical flow should be zero.
In this way, the network cannot remove the blur as in Figure~\ref{fig:flow2}(b).
With the optical flow estimated from the two consecutive blurry images, our network can remove the blur properly in Figure~\ref{fig:flow2}(c).
%

We also train a network without the optical flow constraint (the first term in Eq.~\ref{eq:loss}). We use `w/o flow losses' for short.
%
As there is no constraint for the optical flow, the network cannot effectively relate estimated RNN weights with the optical flow and the deblurring results are worse than the proposed method according to Table~\ref{table:ablation}.
Furthermore, we conduct another experiment with only one blurry image as the input of the network.
In this way, no optical flow information can be utilized in this experiment (`w/o flow' for short) and the network estimates RNN weights from a single image.
This method does not perform well compared to the proposed network.

Different from the proposed network that only estimates the optical flow of one direction, we additionally compare it with a network considering both the forward and backward optical flow simultaneously with three blurry input images (`bidirection flows' for short).
%
According to Table~\ref{table:ablation}, the result is similar to the proposed network.
This is because FlowNets~\cite{dosovitskiy2015flownet} is used here to take efficiency into consideration and it may not have enough ability to simultaneously estimate bidirection flow accurately.
Thus, we use one direction flow in the proposed network.

\vspace{-3mm}
\subsection{Analysis of the Features}
\vspace{-1mm}
According to the analysis from Section~\ref{subsec:network}, the spatially variant RNNs cannot efficiently propagate information in the sparse feature maps.
As the features in decoder is more likely sparse, the proposed network set RNNs in the encoder of the deblurring network.
To validate this, we compare the proposed network with the same number of RNNs in the decoder of the deblurring network (`RNNs in decoder' for short).
As can be seen from Table~\ref{table:ablation}, it cannot deblur as effectively as the proposed network.

\vspace{-3mm}
\subsection{Improvement of the object detection}
\vspace{-1mm}

Our method can facilitate the object detection task. To validate that the proposed deblurring method can help improve the object detection performance, we compare the objects detected from blurry image by \cite{liu2020holistically}, our deblurred image as well as sharp image in Figure~\ref{fig:detect}.
The detection results show that more objects can be detected from our deblurred image than the blurry one.\\

\vspace{-3mm}
\subsection{Limitations}
\vspace{-1mm}
Although our method is able to deblur dynamic scenes, it is less effective for the images with significant blur effect. Figure~\ref{fig:limit} shows one example where the adjacent images contain significant blur and there exists significant different blur effect on the two images. The structures of the results by the proposed method are not preserved well. In addition, there still exists blur residual.
The main reason is that large blur will lead to the two consecutive blurry images with significant appearance differences, which interferes the optical flow estimation.
Thus, inaccurate optical flow will increase the difficulty for the sharp image restoration.

\vspace{-2mm}
\section{Conclusions}
\vspace{-1mm}

We have proposed an effective image deblurring algorithm based on optical flow and spatially variant recurrent neural network.
We first develop an effective deep neural network to estimate the optical flow and then use it to guide the spatially variant recurrent neural network for image deblurring.
We analyze the relationship between blur kernel and optical flow and demonstrate that using optical flow is able to provide useful information for dynamic scene deblurring.
Our network is trained in an end-to-end manner.
Both quantitative and qualitative experimental results demonstrate the effectiveness of the proposed method in terms of accuracy, speed as well as model size.
We also believe that our deblurring method can facilitate object detection as well as tracking.

\ifCLASSOPTIONcaptionsoff
  \newpage
\fi
\bibliographystyle{IEEEtran}
\bibliography{egbib}

\begin{IEEEbiography}[{\includegraphics[width=1in,height=1.25in,clip,keepaspectratio]{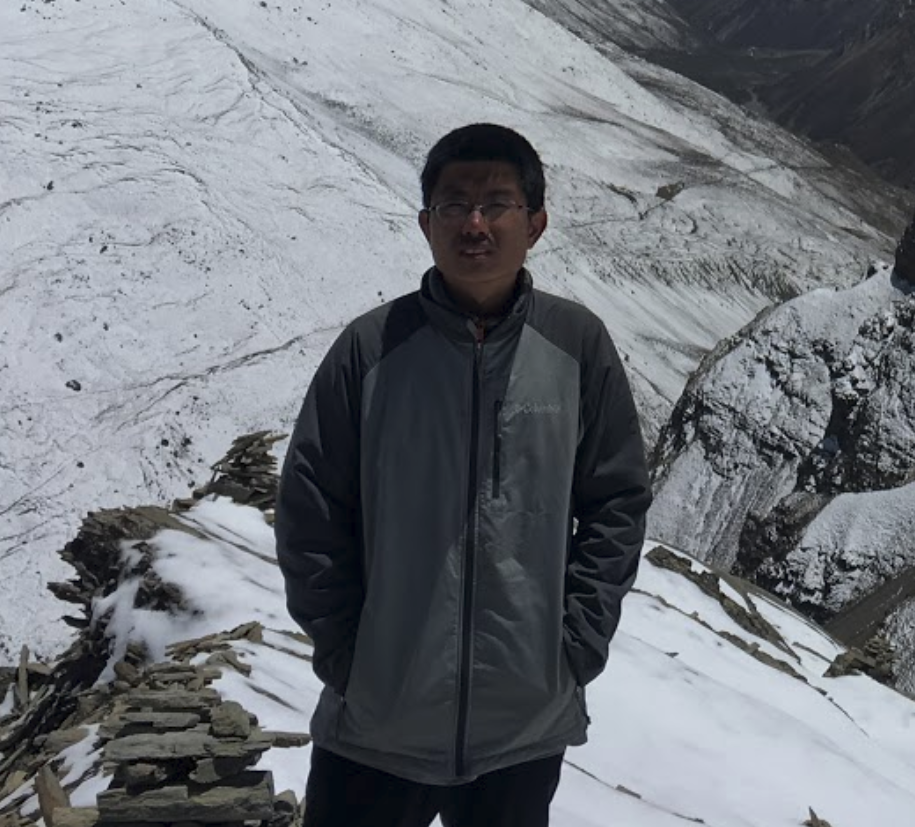}}]{Jiawei Zhang} currently works for SenseTime. He has obtained a PhD degree from City University of Hong Kong in 2018, a master degree from Institute of Acoustics, Chinese Academy of Sciences (IACAS) in 2014 and a bachelor degree of University of Science and Technology of China (USTC) in 2011. His research interest includes computer vision and computational photography.
\end{IEEEbiography}

\begin{IEEEbiography}[{\includegraphics[width=1in,height=1.25in,clip,keepaspectratio]{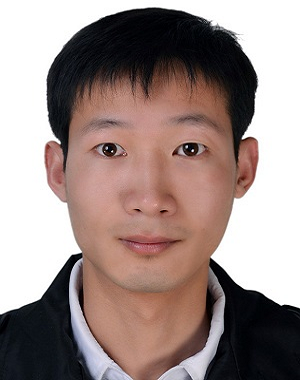}}]{Jinshan
    Pan} is a professor of School of Computer Science and Engineering, Nanjing University of Science and Technology.
    He received the Ph.D. degree in computational mathematics from the Dalian University of Technology, China, in 2017.
    His research interest includes image deblurring, image/video analysis and enhancement, and related vision problems.
\end{IEEEbiography}

\begin{IEEEbiography}[{\includegraphics[width=1in,height=1.25in,clip,keepaspectratio]{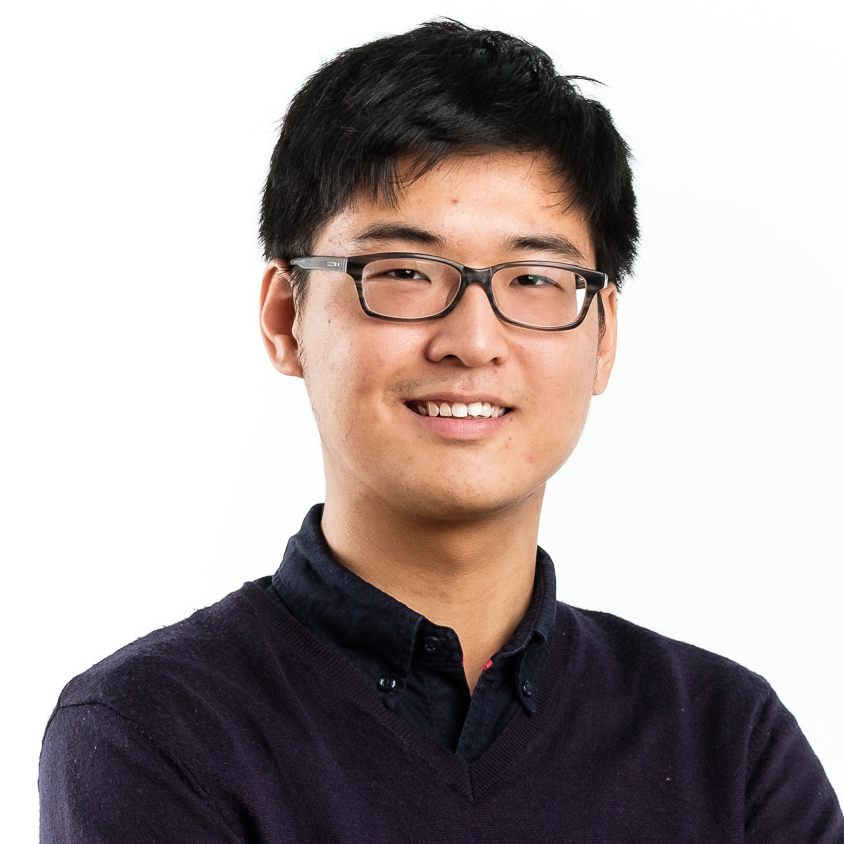}}]{Daoye Wang} is currently a master student at ETH Zurich. He received his bachelor degree from the University of Hong Kong. He was a research Intern at Sensetime Research for a year. His research interest includes computer vision and machine learning.
\end{IEEEbiography}

\begin{IEEEbiography}[{\includegraphics[width=1in,height=1.25in,clip,keepaspectratio]{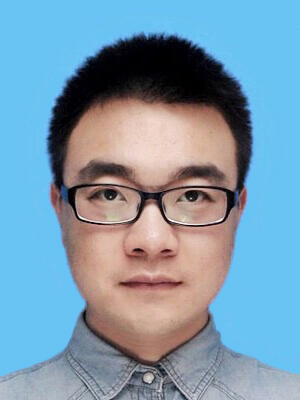}}]{Shangchen Zhou} is currently a Research Assistant at SenseTime. Before that, he received his B.Eng. and M.Eng. degrees from the University of Electronic Science and Technology of China and Harbin Institute of Technology in 2015 and 2018, respectively. His research interests include computer vision and image processing.
\end{IEEEbiography}

\begin{IEEEbiography}[{\includegraphics[width=1in,height=1.25in,clip,keepaspectratio]{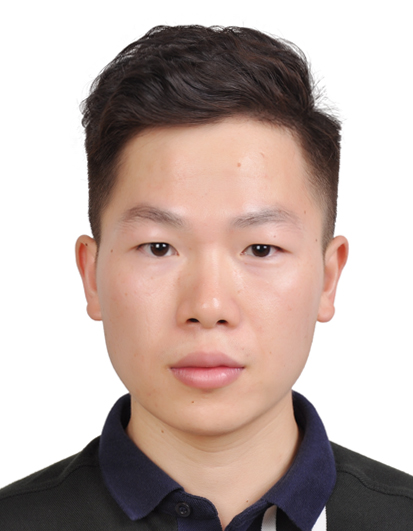}}]{Xing Wei} received the bachelor degree in automation and the Ph.D. degree in control science and engineering from Xi'an Jiaotong University, Xi'an, China, in 2013 and 2019, respectively. He is currently an Assistant Professor with the School of Software Engineering, Xi'an Jiaotong University. He was with the Department of Computer Science, City University of Hong Kong, for two years. His research interests include computer vision, machine learning, and artificial intelligence.
\end{IEEEbiography}

\begin{IEEEbiography}[{\includegraphics[width=1in,height=1.25in,clip,keepaspectratio]{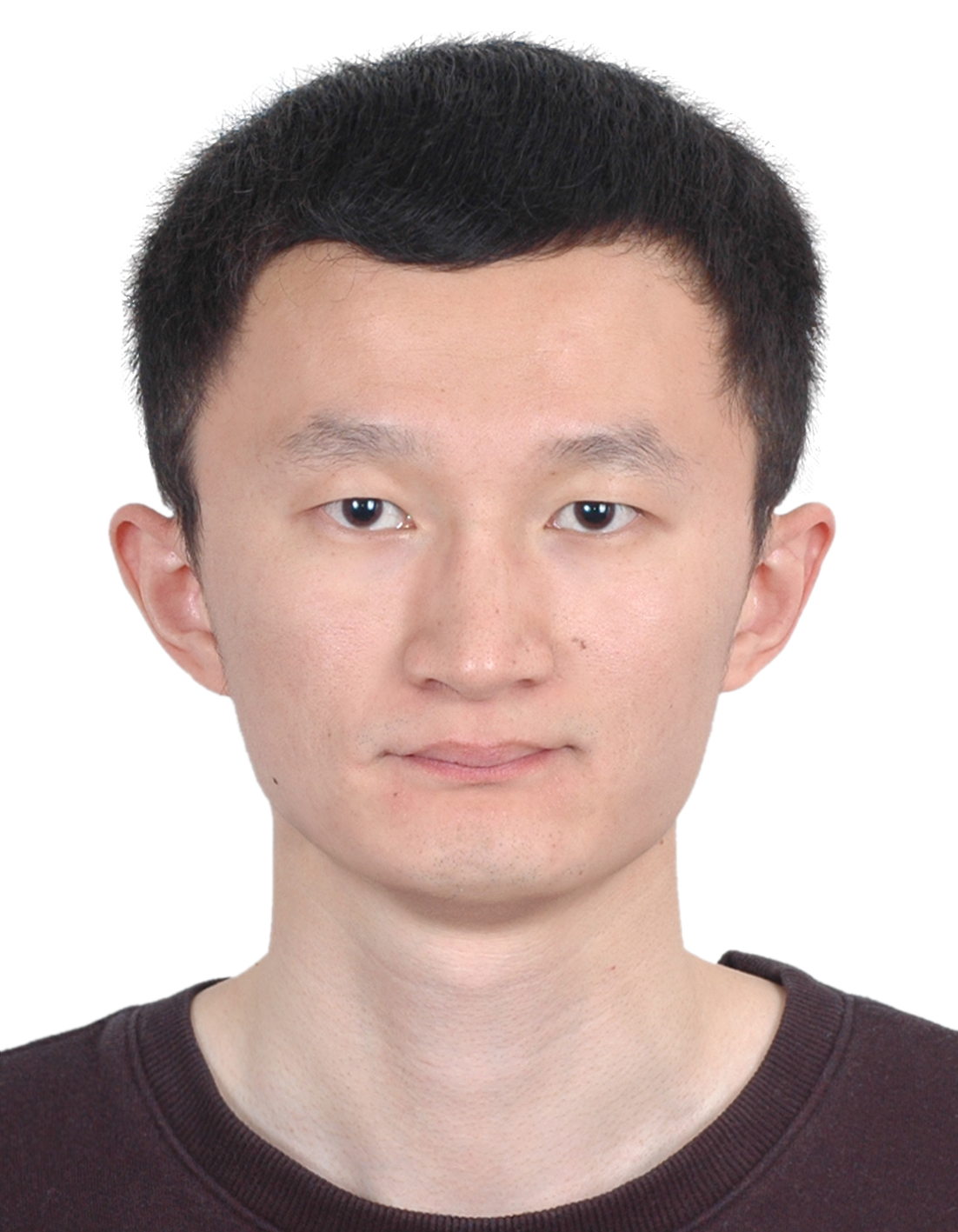}}]{Furong Zhao} is currently an Algorithm Engineer at Pinduoduo Inc. Before that, he was an Algorithm Engineer at Sensetime and received his B.Eng. and M.Eng. degrees from Shandong University and Huazhong University of Science and Technology in 2014 and 2017, respectively. His research interests include computer vision and image processing.
\end{IEEEbiography}

\begin{IEEEbiography}[{\includegraphics[width=1in,height=1.25in,clip,keepaspectratio]{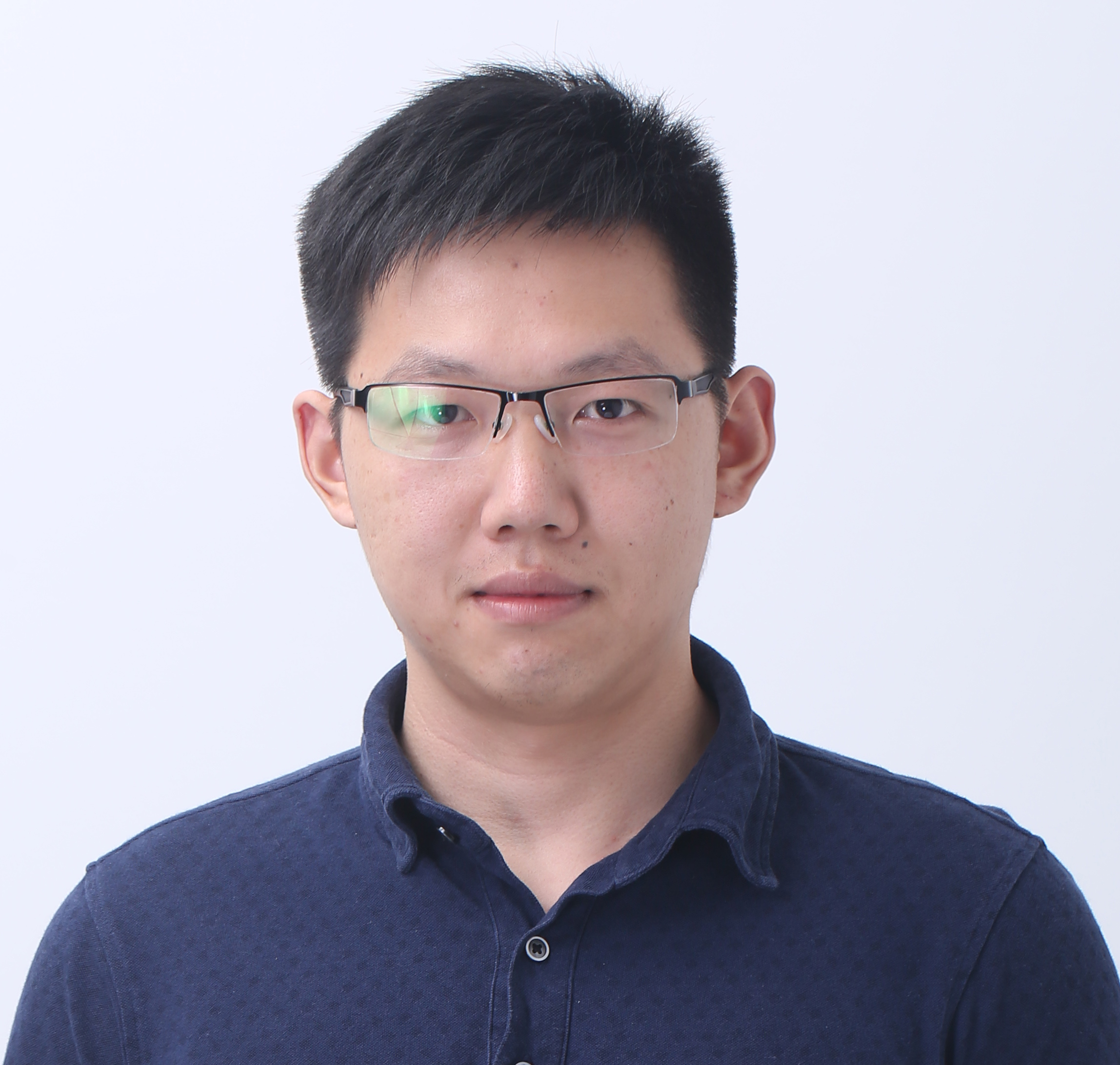}}]{Jianbo Liu} received his B.S. degree from Lanzhou University in 2012, and the M.S. degree from Shenzhen Institutes of Advanced Technology, Chinese Academy of Sciences in 2015. He is currently a Ph.D. student in the Department of Electronic Engineering at The Chinese University of Hong Kong. His research interests include computer vision, deep learning and scene understanding.
\end{IEEEbiography}

\begin{IEEEbiography}[{\includegraphics[width=1in,height=1.25in,clip,keepaspectratio]{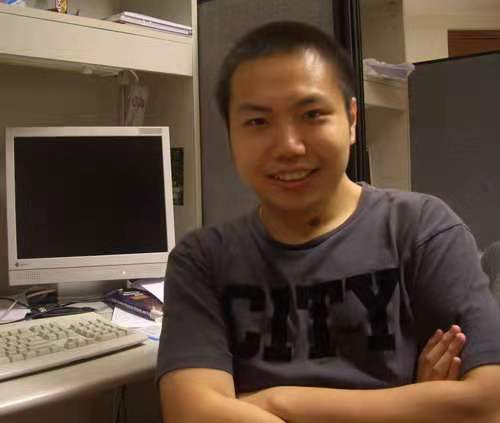}}]{Jimmy Ren} is a research director in SenseTime Research working on various topics in computational photography and computational imaging. He is also an adjunct faculty in Qing Yuan Research Institute, Shanghai Jiao Tong University. His research interests includes computer vision, image processing and machine learning.
\end{IEEEbiography}

\end{document}